\def\paperlanguage{} %% English
\documentclass[letterpaper, 10 pt, conference]{ieeeconf}  % Comment this line out if you need a4paper

\usepackage[dvips]{graphicx}
\usepackage{bm}
\usepackage{cite}
\usepackage{flushend}
\include{preamble}

\IEEEoverridecommandlockouts                              % This command is only needed if
\title{\LARGE \textbf
  {
    \switchlanguage%
    {%
      WiXus: A Wheeled-Legged Robot with Wire-Driven \\Environmental Utilizing to Integrate Mobility and Manipulation
    }%
    {%
      WiXus: A Wheeled-Legged Robot with Wire-Driven \\Environmental Utilizing to Integrate Mobility and Manipulation
    }%
  }
}

\author{Shintaro Inoue$^{1}$, Kento Kawaharazuka$^{1}$, Temma Suzuki$^{1}$, Sota Yuzaki$^{1}$, and Kei Okada$^{1}$% <-this % stops a space
  \thanks{$^{1}$ The authors are with the Department of Mechano-Informatics, Graduate School of Information Science and Technology, The University of Tokyo, 7-3-1 Hongo, Bunkyo-ku, Tokyo, 113-8656, Japan.
    {\texttt\small [s-inoue, kawaharazuka, t-suzuki, yuzaki, k-okada]@jsk.imi.i.u-tokyo.ac.jp}
  }
}
\begin{document}

\maketitle
\thispagestyle{empty}
\pagestyle{empty}

%%%%%%%%%%%%%%%%%%%%%%%%%%%%%%%%%%%%%%%%%%%%%%%%%%%%%%%%%%%%%%%%%%%%%%%%%%%%%%%%
\begin{abstract}
  \switchlanguage%
  {%
    Wheeled-legged robots, which have wheels at their feet and achieve high mobility by coordinating wheel drive and leg drive, 
    have been developed. 
    These robots have been developed purely as platforms specialized for locomotion. 
    Therefore, they do not have a means to repurpose their legs for roles other than locomotion, 
    such as object manipulation or tool utilization. 
    In this paper, we address the problem of how to draw out the potential task-execution capability of the legs 
    by freeing them from the roles of locomotion through external body support. 
    To this end, we propose and develop a new robot, WiXus, 
    which fuses a wheeled-legged mechanism with a wire-driven mechanism that utilizes the external environment. 
    The developed WiXus demonstrates not only planar locomotion with wheeled-legged drive, 
    but also three-dimensional mobility such as cliff climbing by coordinating wire-driven and wheeled-legged actuation. 
    Furthermore, by suspending the body with wire-driven actuation, 
    WiXus successfully repurpose its legs as arms to perform object manipulation, 
    (e.g., rescuing a dog (stuffed animal)), and tool utilization (e.g., harvesting an apple (mockup) with loppers). 
    This study demonstrates that the approach of utilizing the environment with wire-driven actuation 
    is a new design principle that extends the operational domain of wheeled-legged robots.  
  }%
  {%
    脚先にタイヤを有し、タイヤ駆動と脚駆動を連携させることで高い移動能力を持つ、脚車輪ロボットが開発されてきた。
    これらのロボットは、あくまで移動に特化したプラットフォームとして開発されてきた。
    そのため、その脚を支持や移動以外の役割、すなわち物体操作や道具利用へ転用する手段を持たない。
    本稿では、外部からの身体支持によって脚を移動と支持の役割から解放し、
    その潜在的なタスク実行能力をいかにして引き出すか、という課題に取り組む。
    そこで我々は、脚車輪機構と、ワイヤ駆動による外部環境利用機構とを融合させた新しいロボットWiXusを提案・開発する。
    開発したWiXusは、従来の脚車輪駆動を用いた平面移動に加えて、
    ワイヤ駆動と脚車輪駆動の協調による崖越えといった三次元移動能力を実証した。
    さらに、ワイヤ駆動で機体を浮遊させて脚を腕として利用し、犬（ぬいぐるみ）を救助する物体操作や、
    刈込鋏でリンゴ（模型）を収穫する道具利用にも成功した。
    本研究は、ワイヤ駆動による環境利用というアプローチが、脚車輪ロボットの活動領域を拡張する、
    新たな設計指針であることを示すものである。
  }%
\end{abstract}

\section{Introduction}\label{sec:introduction}
\switchlanguage%
{%
  Wheeled-legged robots, which combine legged and wheeled actuation, 
  have recently been studied and developed actively as locomotion platforms that achieve both high-speed locomotion 
  on flat terrain and strong traversability on uneven terrain. 
  In academic research, robots such as ANYmal on wheels \cite{bjelonic2021wholebody,joonho2024learning}, 
  SR600 \cite{zhang2019system,liu2019dynamic}, and SUSTech-Nezha \cite{yang2023design} have been developed. 
  Furthermore, practical implementations by industry are also rapidly advancing, 
  including Ascento \cite{klemm2019ascento}, TITA \cite{tita2024}, DIABOLO \cite{diabolo2024}, Go2-W \cite{go2w2025}, 
  and B2-W \cite{b2w2025}. 
  These developments indicate that wheeled-legged robots are already establishing their position 
  as one of the completed forms of ground locomotion robots.

  The high mobility of these robots comes from the ability to switch between the advantages of legs 
  and wheels depending on the situation. 
  On flat ground, they perform energy-efficient high-speed locomotion with wheels, 
  while when encountering obstacles such as steps or rubble, 
  they function as legs to traverse various terrains efficiently and stably. 
  Based on this excellent locomotion capability, wheeled-legged robots with manipulators have emerged, 
  such as Handle \cite{handle2019}, which performs logistics tasks, 
  and humanoid-type wheeled-legged robots with dual arms aiming for versatile operations \cite{li2018wlr,li2019wlr2}. 
  However, even these robots \cite{handle2019,li2018wlr,li2019wlr2} fundamentally depend on the stability and driving force 
  provided by having their legs or wheels in contact with the ground. 
  Therefore, high places unreachable by legs, cliffs, and vertical walls remain outside their operational domain.

  \begin{figure}[t]
    \begin{center}
      \includegraphics[width=1.0\columnwidth]{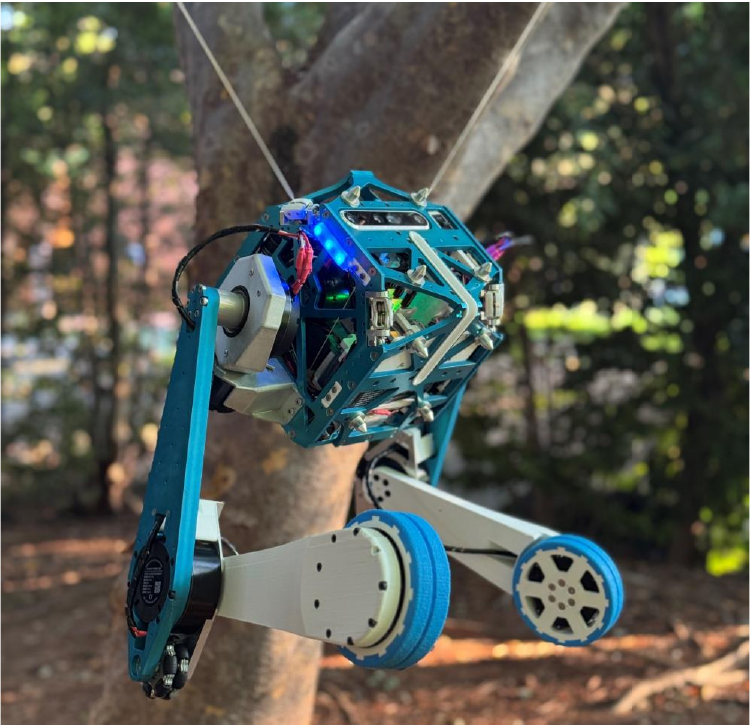}
      \vspace{-4ex}
      \caption{
        Overview of WiXus, a robot that fuses a wheeled-legged system with a wire-driven system that anchors to the environment.
      }
      \vspace{-5.0ex}
      \label{fig:fig1}
    \end{center}
  \end{figure}

  In contrast, to tackle the challenge of tasks in large three-dimensional spaces, an entirely different class of robots, 
  wire-driven robots, has been studied. 
  Cable-Driven Parallel Robots (CDPRs), which use wires deployed from fixed frames to actuate end-effectors, 
  can support heavy loads with high precision in vast workspaces, and have applications such as CoGiRo \cite{8967836}, 
  which mounts a manipulator, and SkyCam \cite{cone1985skycam}, which moves cameras in stadiums. 
  More recently, this approach has evolved further, with research that utilizes the external environment itself, 
  such as poles and trees, as the support base instead of relying on fixed frames. 
  For example, there are studies where tethered drones serve as anchors to improve the traversability 
  of ground vehicles \cite{8794265}, as well as CubiX \cite{inoue2024cubix,inoue2024overcoming}, 
  a portable wire-driven robot that autonomously anchors wires to surrounding structures.

  In this study, we propose a new approach that fuses wheeled-legged system and wire-driven system. 
  Specifically, while using the high ground mobility of a wheeled-legged robot as the foundation, 
  we augment it with the ability to reach areas that were previously outside the operational domain, 
  such as high places, cliffs, and vertical walls, by utilizing the environment through wire anchoring.
  Furthermore, by suspending the body with wire-driven actuation, the legs are freed from locomotion, 
  and can be repurposed as arms for object manipulation and tool utilization. 
  To embody this concept, we propose WiXus, shown in \figref{fig:fig1}, and demonstrate its design and capabilities. 
  WiXus represents a novel attempt to integrate environmental utilization by wire-driven actuation into a wheeled-legged robot, 
  achieving tasks that were previously difficult, such as cliff climbing, manipulation while floating, and tool utilization. 
  % While the demonstrations involve partial operator input, 
  % they represent an initial step to validate the feasibility of the proposed concept, 
  % which can be extended to autonomous task execution by integrating environmental perception and motion planning. 
  % Moreover, 
  The proposed approach of utilizing the environment has the potential to extend to real-world applications 
  such as outdoor operations, disaster response, and agriculture, 
  and thus has practical significance beyond laboratory demonstrations.
}%
{%
  % 脚車輪の紹介
  脚駆動と車輪駆動を融合させた脚車輪ロボットは、
  平坦地での高速移動と不整地での高い踏破性を両立する移動プラットフォームとして、
  近年活発な研究開発が行われている。
  学術研究ではこれまでに、ANYmal on wheels\cite{bjelonic2021wholebody,joonho2024learning}や
  SR600\cite{zhang2019system,liu2019dynamic}、SUSTech-Nezha\cite{yang2023design}などが開発されてきた。
  さらに、 Ascento\cite{klemm2019ascento}やTITA\cite{tita2024}、
  DIABOLO\cite{diabolo2024}、Go2-W\cite{go2w2025}、B2-W\cite{b2w2025}などをはじめとする
  企業による実用化も急速に進んでいる。
  これらは、脚車輪ロボットがすでに地上を移動するロボットの1つの完成形として、
  その地位を確立しつつあることを示している。

  \begin{figure}[t]
    \begin{center}
      \includegraphics[width=1.0\columnwidth]{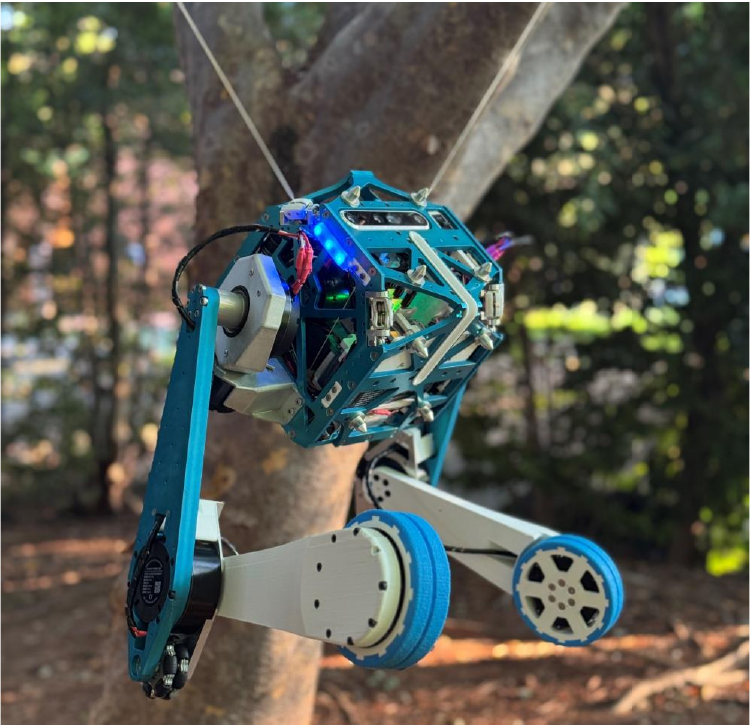}
      \vspace{-4.0ex}
      \caption{
        Overview of WiXus, a robot that fuses a wheeled-legged system with a wire-driven system that anchors to the environment.
      }
      \vspace{-4.0ex}
      \label{fig:fig1}
    \end{center}
  \end{figure}

  % 脚車輪の利点と限界
  これらのロボットの高い移動能力は、脚と車輪が持つ利点を状況に応じて使い分けられることに起因する。
  平坦地では車輪によるエネルギー効率の良い高速移動を行い、
  段差や瓦礫といった障害物に遭遇した際には脚として機能させることで、
  様々な環境を効率的かつ安定的に走破してきた。
  近年では、この優れた移動能力を基盤とし、
  マニピュレータを搭載して物流作業を行うHandle\cite{handle2019}や、
  双腕を持つことで汎用的な作業を目指すヒューマノイド型の脚車輪ロボット\cite{li2018wlr,li2019wlr2}も登場している。
  しかし、これらのロボット\cite{handle2019,li2018wlr,li2019wlr2}でさえ、 
  その安定性と駆動力は脚や車輪が地面に接地しているという点に根本的に依存している。
  そのため、脚では到達不可能な高所や、崖、壁面といった非水平な三次元空間は、依然として活動領域外であった。

  % CDPR、環境接続ワイヤ駆動ロボットの紹介
  これに対し、三次元的に広がる大きな空間での作業という課題に取り組む、
  全く異なる系統のロボットとして、ワイヤ駆動ロボットが存在する。
  固定フレームから展張したワイヤを用いるCDPR (Cable-Driven Parallel Robot) は、
  広大な作業空間において重量物を高精度に支持でき、
  マニピュレータを搭載したCoGiRo\cite{8967836}や、スタジアムのカメラを移動させるSkyCam\cite{cone1985skycam}といった応用例がある。
  近年、このアプローチはさらに発展し、固定フレームに頼らず柱や木などの外部環境そのものを支持基盤として利用する研究が登場した。
  例えば、ワイヤ接続されたドローンがアンカー役となり地上車両の走破性を向上させる研究\cite{8794265}や、
  ロボット自身が周囲の構造物にワイヤを接続するポータブルな
  ワイヤ駆動ロボットであるCubiX\cite{inoue2024cubix,inoue2024overcoming}が提案されている。

  % WiXus
  そこで本研究では、脚車輪ロボットとワイヤ駆動ロボットの二つを融合させるという、新しいアプローチを提案する。
  すなわち、脚車輪ロボットの持つ高い地上移動能力を基盤としながら、ワイヤによる環境利用によって、
  これまで活動領域外であった高所、崖や壁面といった場所へ到達する能力を付与する。
  さらに、ワイヤ駆動によって浮遊することで、これまで移動と支持に束縛されていた脚を解放し、
  道具利用や物体操作をができる腕として転用する。
  このコンセプトを具現化したロボットとして、本稿では\figref{fig:fig1}に示すWiXusを提案し、その設計と能力を実証する。
  % 位置目的の明確化
  WiXusは、ワイヤ駆動による環境利用を脚車輪ロボットに統合した新たな試みであり、
  崖上りや浮遊中の物体操作、道具利用といった従来困難であったタスクを実現する。
  これらの実験は操縦者の入力を含む形で実施されているが、提案コンセプトの有効性を実証するための第一歩であり、
  環境認識や運動計画を統合することで自律的なタスク遂行へと発展可能である。
  さらに、環境を利用するという本アプローチは、屋外や災害現場、農業分野など
  実環境での応用に拡張できる潜在性を有しており、単なる実験室デモにとどまらない実用的意義を持つ。
}%

\section{Design of WiXus} \label{sec:hardware}
\subsection{Overall Design}
\switchlanguage%
{%
  \begin{figure}[t]
    \begin{center}
      \includegraphics[width=1.0\columnwidth]{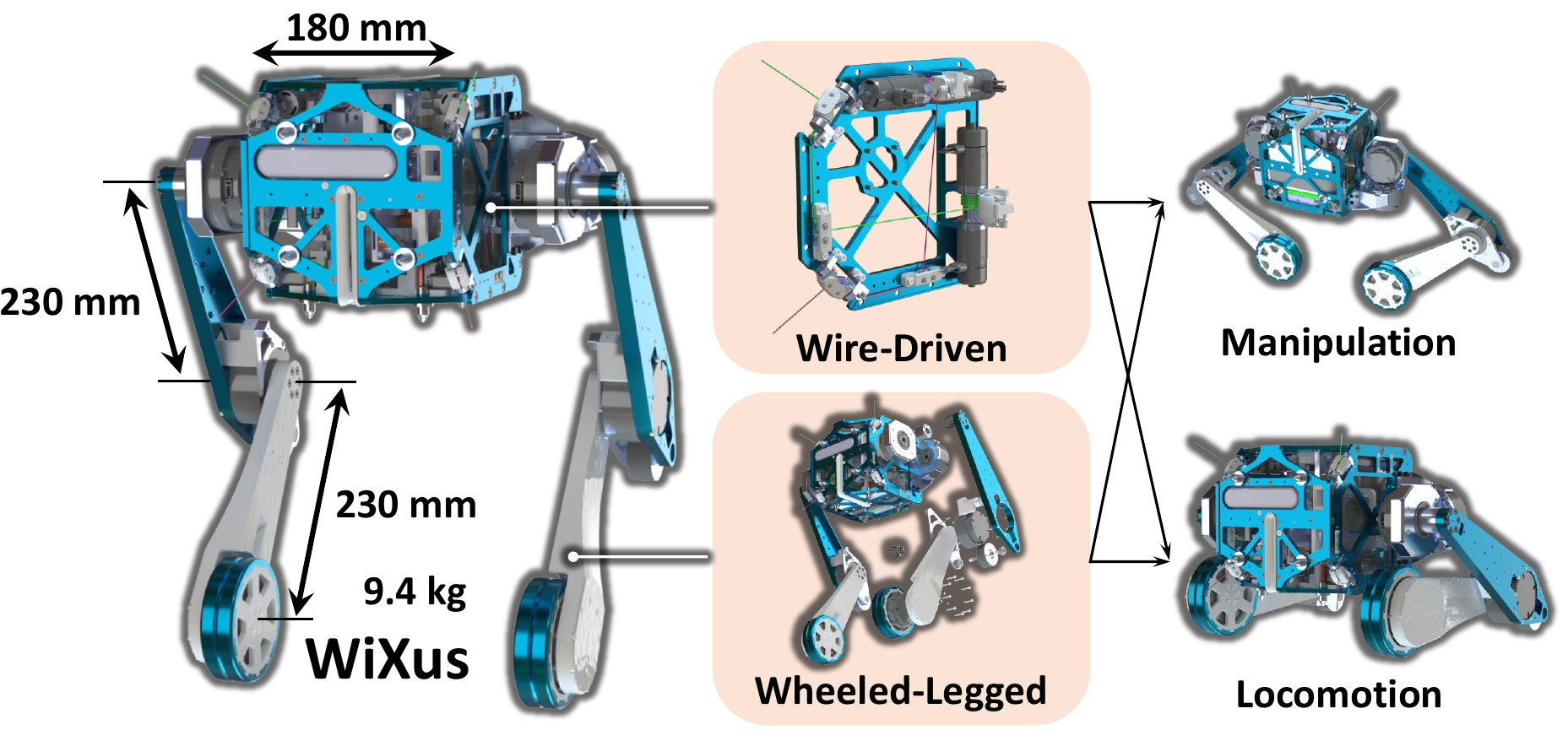}
      \vspace{-4.0ex}
      \caption{
        Hardware overview of WiXus. 
        It has a 180 mm cubic main body with modules for its wire-driven system and two attached wheeled-legs. 
        These components work together to achieve both manipulation and locomotion.
      }
      \vspace{-4.0ex}
      \label{fig:hardware_overview}
    \end{center}
  \end{figure}

  An overview of the design of WiXus is shown in \figref{fig:hardware_overview}.
  WiXus consists of a main body equipped with wire winding modules, 
  and two wheeled-legs each with a 3-DOF (Roll-Pitch-Pitch) leg and an attached wheel.
  By integrating wire-driven actuation with the wheeled-legs, 
  WiXus enhances both its object manipulation capability and locomotion capability.

  The main body is a cube with dimensions of 180 mm on each side, and four wires can be deployed from its front vertices.
  By anchoring these wires to surrounding structures in the environment and winding them, the robot can drive itself.
  In addition, one more wire can be deployed from the center of the rear side, which enables the robot to attach tools to its body.

  The two wheeled-legs are attached to the left and right sides of the main body.
  In addition to performing locomotion with legs and wheels, 
  they also function as manipulators when the robot is suspended by the wire-driven system.
  During manipulation, the main body is rotated by $-90$ degrees in the pitch direction, 
  which gives the legs 3 DOFs (Yaw-Pitch-Pitch) as arms.
}%
{%
  \begin{figure}[t]
    \begin{center}
      \includegraphics[width=1.0\columnwidth]{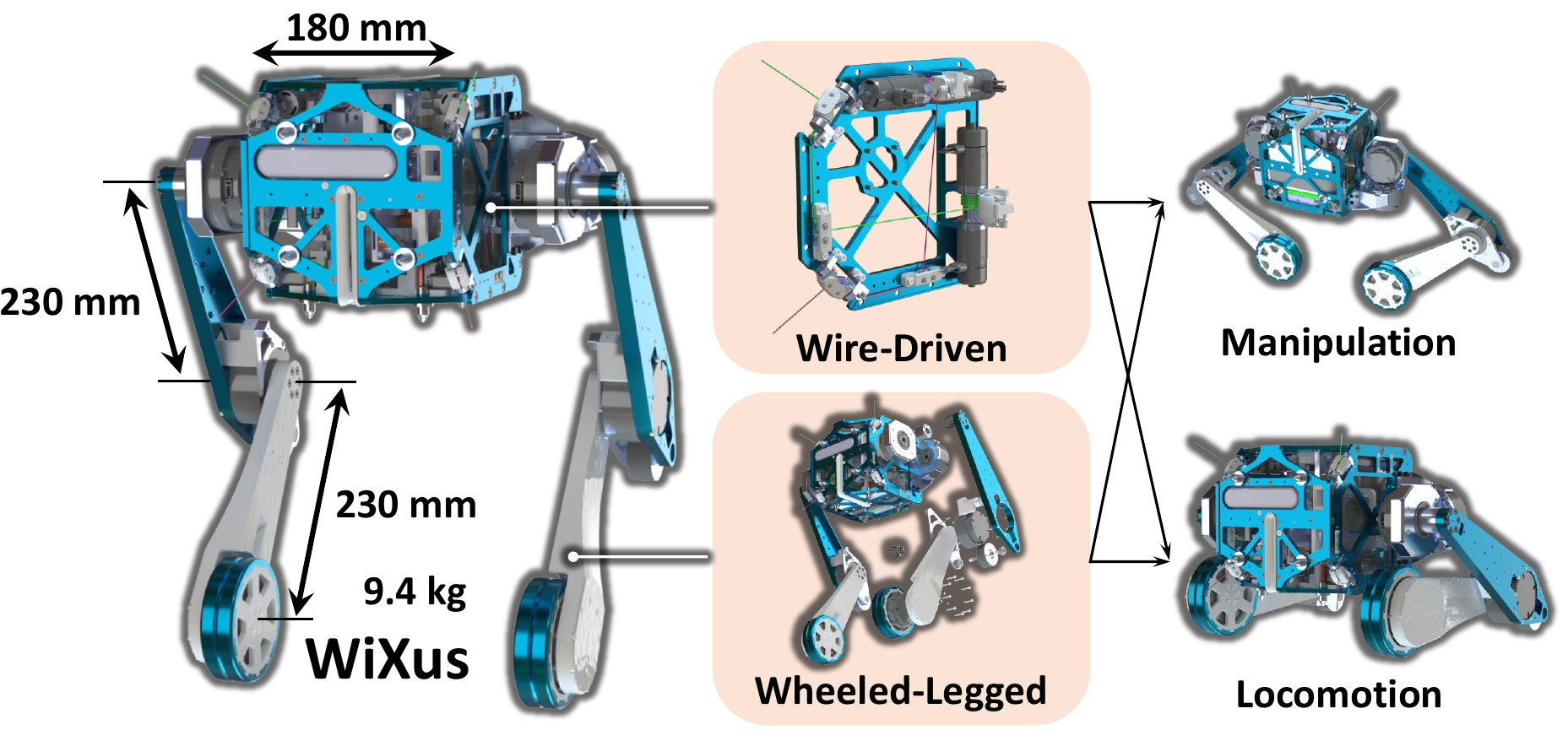}
      \vspace{-4.0ex}
      \caption{
        Hardware overview of WiXus. 
        It has a 180 mm cubic main body with modules for its wire-driven system and two attached wheeled-legs. 
        These components work together to achieve both manipulation and locomotion.
      }
      \vspace{-4.0ex}
      \label{fig:hardware_overview}
    \end{center}
  \end{figure}

  WiXusの設計の全体像を\figref{fig:hardware_overview}に示す。
  WiXusは、ワイア巻取りモジュールを内蔵したbase linkに、
  3自由度 (Roll-Pitch-Pitch) の脚に車輪を組み付けた2つの脚車輪からなる。
  ワイヤ駆動と脚車輪を融合させることで、
  ロボットの物体操作能力と移動能力をそれぞれ向上させる。

  base linkは180 mm立方の立方体であり、前面の各頂点から4本のワイヤが引き出せる。
  これらのワイヤをロボット外部の周辺環境に結びつけ、巻き取ることによって自らを駆動できる。
  また、後面の中心からも1本のワイヤを引き出すことができ、これで道具を身体に合体させることが可能である。

  2つの脚車輪はbase linkの両側面に取り付けられており、
  脚と車輪による移動に加えて、ワイヤ駆動で浮遊した後のマニピュレータの機能も果たす。
  マニピュレータ時は、ベースリンクがピッチ方向に-90度回転するため、腕自由度は3自由度 (Yaw-Pitch-Pitch) となる。
}%

\subsection{Wheeled-Leg Design}
\switchlanguage%
{%
  \begin{figure}[t]
    \begin{center}
      \includegraphics[width=1.0\columnwidth]{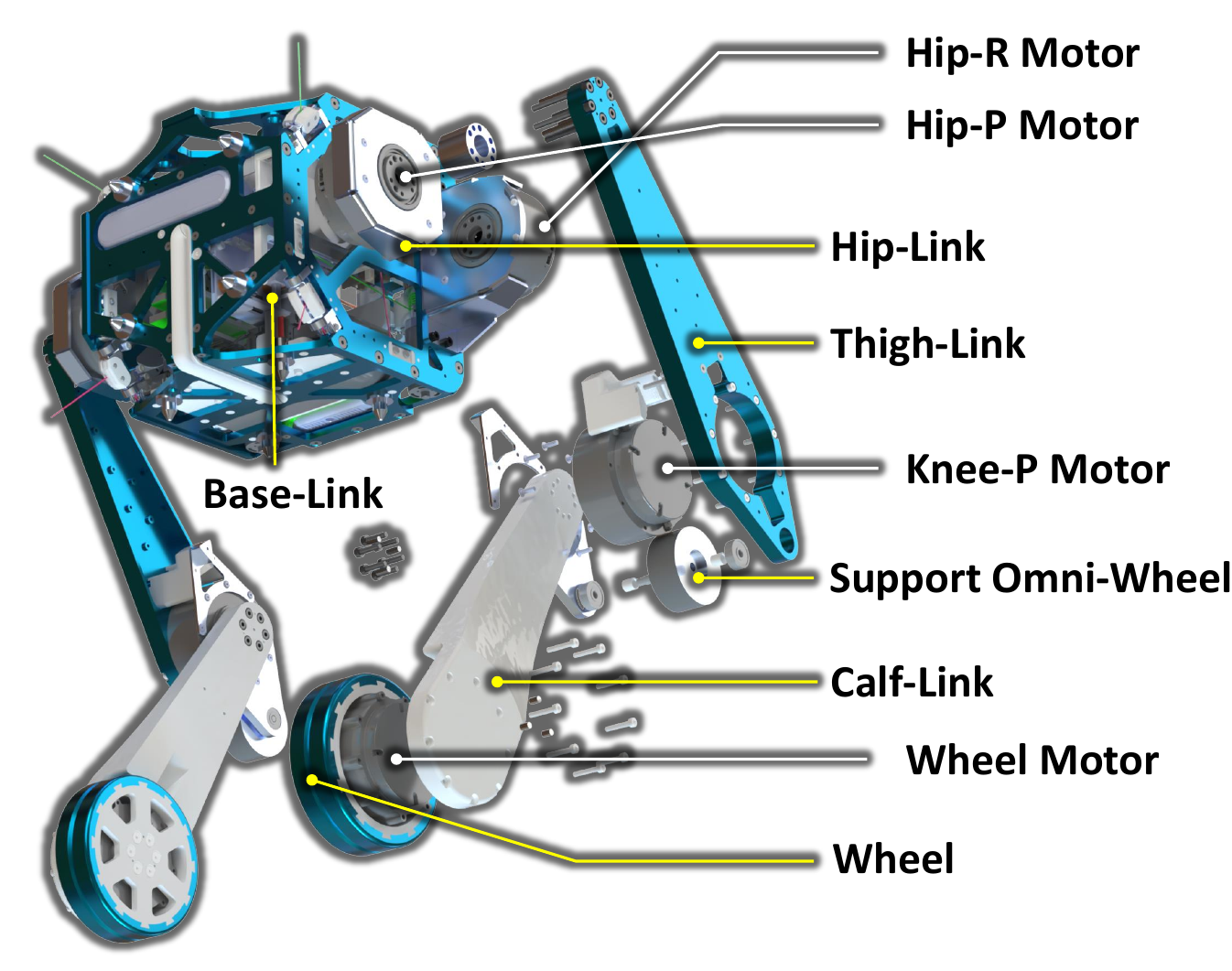}
      \vspace{-4.0ex}
      \caption{
        Overview of the wheeled-leg design. 
        It consists of a 3-DOF leg with a Roll-Pitch-Pitch joint configuration, a wheel mounted at its tip,
        and a support omni-wheel at the knee.
      }
      \vspace{-4.0ex}
      \label{fig:wheeled-leg}
    \end{center}
  \end{figure}

  The structure of the designed wheeled-leg is shown in \figref{fig:wheeled-leg}.
  Each wheeled-leg consists of a Hip-Link, a Thigh-Link, a Calf-Link, and a wheel.
  The joint configuration from the Base-Link to the Calf-Link follows the Roll-Pitch-Pitch arrangement, 
  which is the same as in typical wheeled-legged robots \cite{klemm2019ascento,tita2024,diabolo2024,go2w2025,b2w2025}.
  While these robots \cite{klemm2019ascento,tita2024,diabolo2024,go2w2025,b2w2025} use linkage mechanisms to 
  transmit the elbow joint rotation in order to reduce the weight of the leg tip, 
  WiXus directly mounts motors on the elbow joints to maximize the leg workspace.
  Robstride02 actuators were adopted for the joints 
  because of their high power density, and smaller CyberGear actuators of the same family were used for the wheels.

  The lengths of the Thigh-Link and Calf-Link are 230 mm each.
  This was determined as 80\% of the self-supportable leg length calculated 
  from the maximum continuous torque of the actuators and the mass of the robot.

  An omni-wheel is attached to the Thigh-Link as a support wheel, 
  enabling stable two-wheeled driving with the elbows in contact with the ground 
  (as shown in the lower right of \figref{fig:hardware_overview}).

  The Calf-Links and wheels were manufactured using a 3D printer, 
  while the other links were made from machined aluminum parts.
  For the filament of the 3D-printed parts, POTICON (Potassium Titanate Compound) NTL34M, a high-strength plastic, was used.
  For the wheels, 3D-printed TPU (Thermoplastic Polyurethane) parts with a fuzzy skin covering their outer surface 
  were fitted to increase the friction coefficient against the ground and manipulated objects.
}%
{%
  \begin{figure}[t]
    \begin{center}
      \includegraphics[width=1.0\columnwidth]{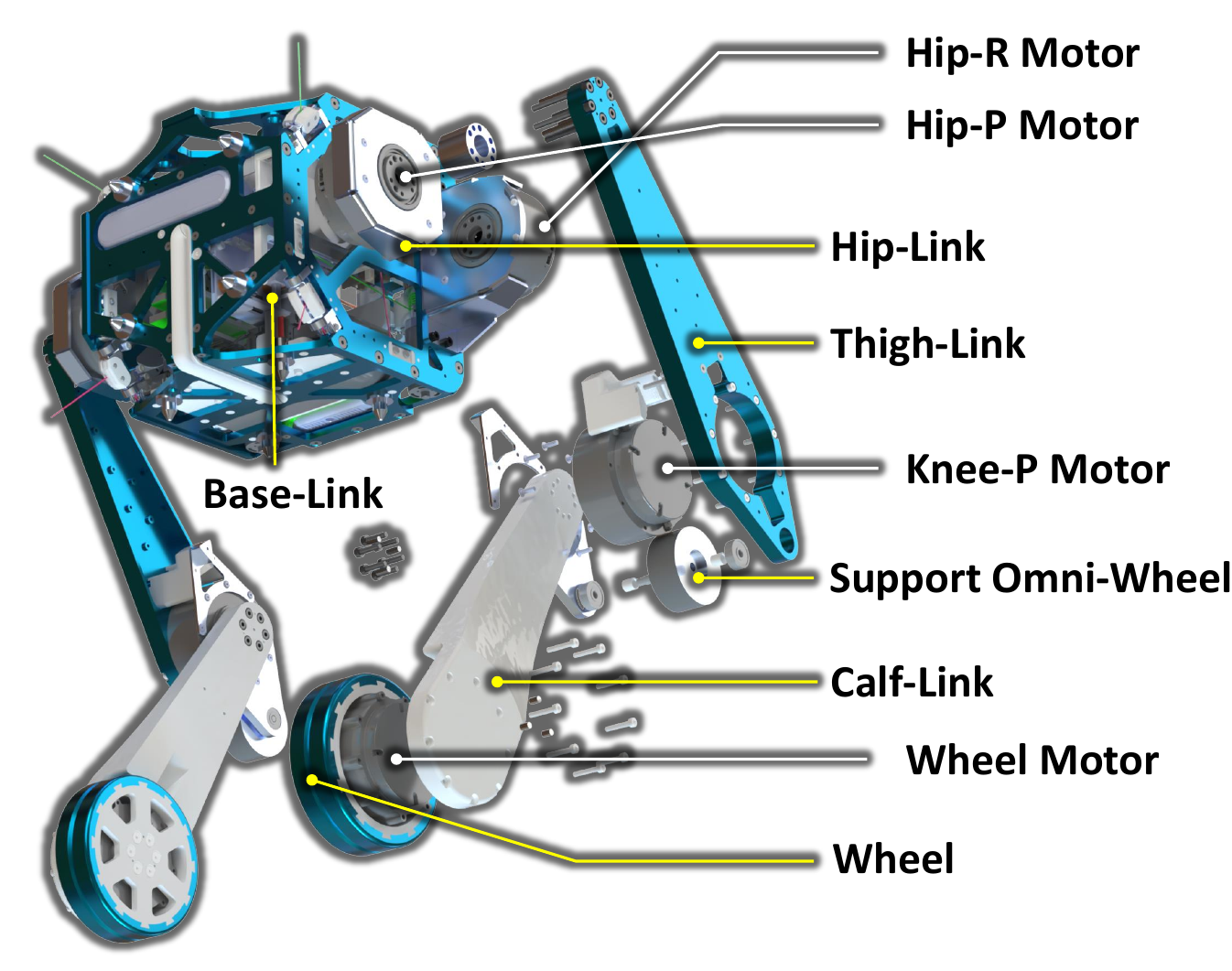}
      \vspace{-4.0ex}
      \caption{
        Overview of the wheeled-leg design. 
        It consists of a 3-DOF leg with a Roll-Pitch-Pitch joint configuration, a wheel mounted at its tip,
        and a support omni-wheel at the knee.
      }
      \vspace{-4.0ex}
      \label{fig:wheeled-leg}
    \end{center}
  \end{figure}
  設計した脚車輪の構造を\figref{fig:wheeled-leg}に示す。
  脚車輪は、hip link、thigh link、calf link、車輪の4部位からなる。
  base linkからcalf linkまでの接続軸の方向はRoll-Pitch-Pitchであり、
  一般的な脚車輪ロボットの関節配置\cite{klemm2019ascento,tita2024,diabolo2024,go2w2025,b2w2025}と同様である。
  これらのロボット\cite{klemm2019ascento,tita2024,diabolo2024,go2w2025,b2w2025}は
  脚先を軽量化するために肘関節の回転をリンク機構で行うが、
  WiXusでは脚の可動域を最大化するためにモータを肘関節に直接取り付けている。
  関節のモータには出力密度の高さからRobstride02を採用し、車輪にはそれと同系統で小型のCyberGearを採用した。

  thigh linkとcalf linkの長さはそれぞれ230 mmである。
  これは、搭載したモータの最大トルクとロボットの質量から計算される自立可能な脚長さを8割に縮めたものである。

  肘部分には補助輪としてオムニホイールを取り付けることで、
  肘を地面に付いた安定した2輪走行 (\figref{fig:hardware_overview}右下に示される姿勢) を可能としている。

  calf linkと車輪は3Dプリンタで造形しており、その他のリンクはアルミ加工部品である。
  3Dプリント部品のフィラメントには、高強度なプラスチックであるPOTICON (Potassium Titanate Compound) NTL34Mを使用した。
  車輪には外周を覆うようにファジースキンで3DプリントしたTPU (Thermoplastic Polyurethane) 部品をはめあわせることで、
  地面や物体操作対象との摩擦係数を上昇させている。
}%

\subsection{Wire Winding Module Design}
\switchlanguage%
{%
  \begin{figure}[t]
    \begin{center}
      \includegraphics[width=1.0\columnwidth]{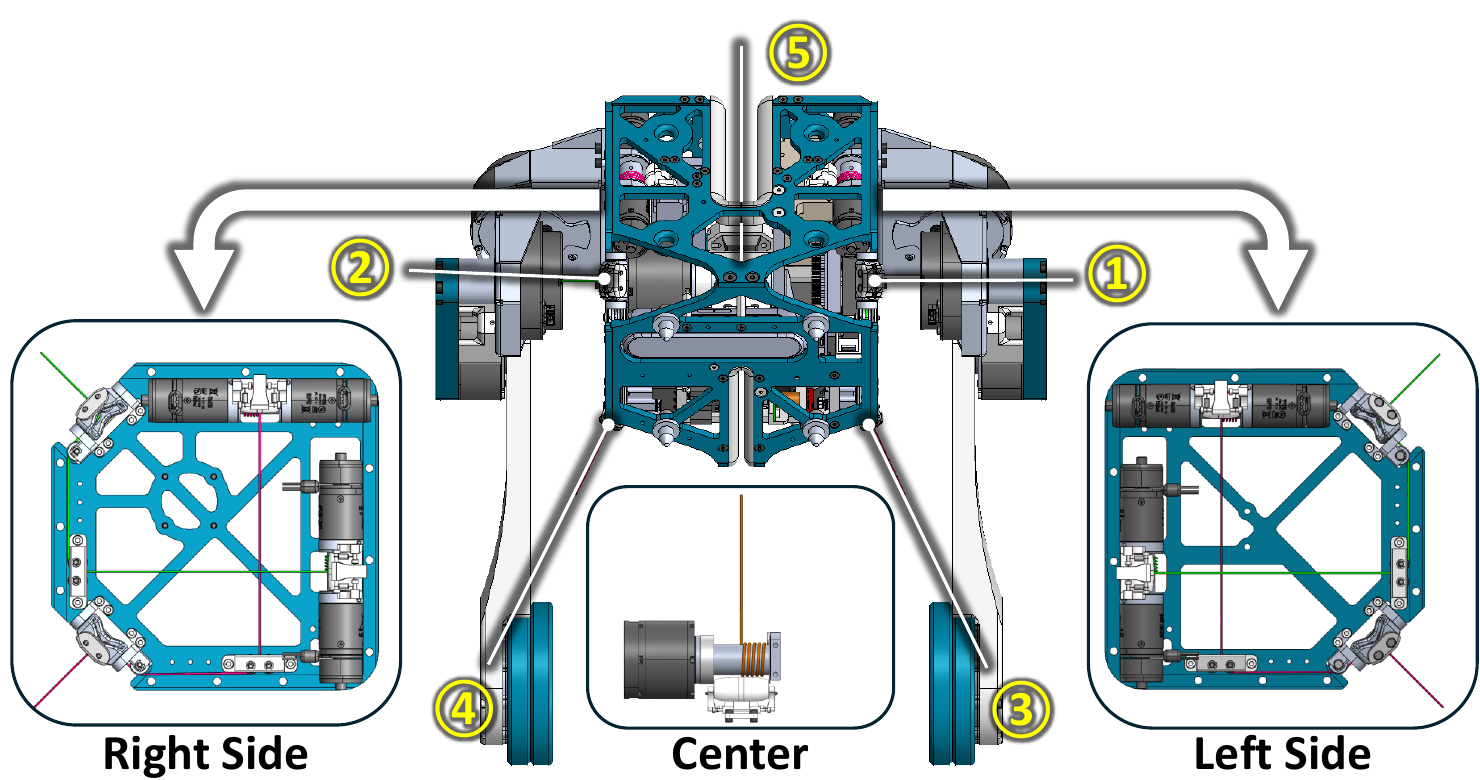}
      \vspace{-4.0ex}
      \caption{
        Arrangement of the wire winding modules on the main body.
        WiXus is equipped with a total of five modules: 
        four modules for environmental anchoring are mounted on the sides, 
        while a single module for tool attachment is located at the center.
      }
      \vspace{-2.0ex}
      \label{fig:wire_overview}
    \end{center}
  \end{figure}

  \begin{figure}[t]
    \begin{center}
      \includegraphics[width=1.0\columnwidth]{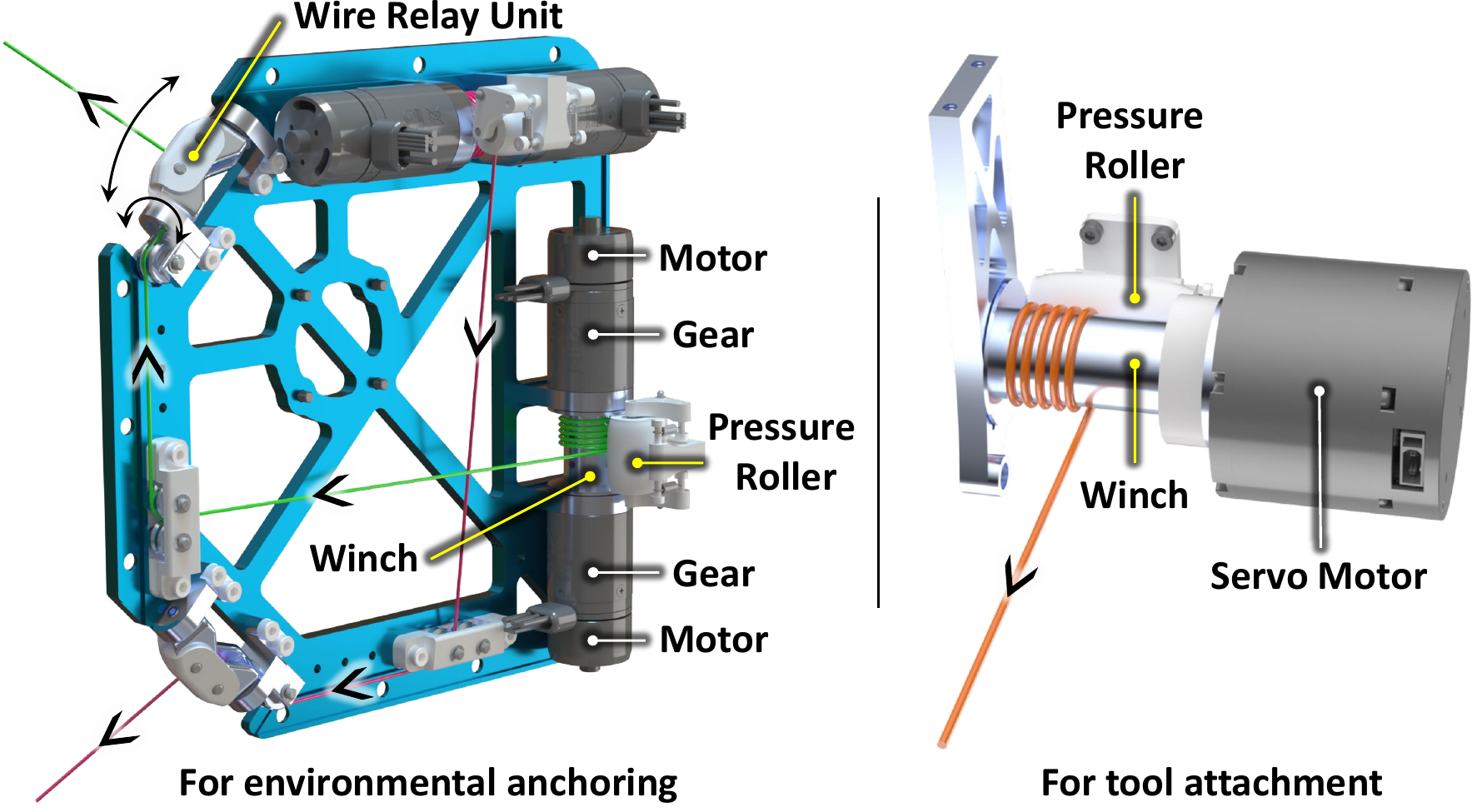}
      \vspace{-4.0ex}
      \caption{
        Overall structure of the two types of wire winding modules: 
        one for environmental anchoring and one for tool attachment. 
        A motor drives a winch to wind the wire.
      }
      \vspace{-5.0ex}
      \label{fig:wire_module}
    \end{center}
  \end{figure}

  The arrangement of the wires and wire winding modules of WiXus is shown in \figref{fig:wire_overview}.
  WiXus can wind a total of five wires: four wires for anchoring to the environment and one wire for attaching tools.

  The wires for environmental anchoring originate from each front vertex (\figref{fig:wire_overview}\ctext{1}--\ctext{4}).
  Each wire is wound by one of the wire winding modules installed in pairs on the left and right sides.
  The wires at \figref{fig:wire_overview}\ctext{1} and \ctext{3} correspond to the left-side modules, 
  and those at \ctext{2} and \ctext{4} correspond to the right-side modules.
  \figref{fig:wire_module} (left) shows the detailed structure of the side modules.
  Each wire winding module consists of a winch that winds the wire, 
  two motors that rotate the winch, and a pressing roller that holds the wire in place.
  The winch has a diameter of 15 mm and a width of 21.5 mm, and it is rotated by two motors mounted on both sides.
  M2006 motors, which are cylindrical and controllable via CAN communication, were adopted.
  The pressing roller keeps the wound wire aligned on the winch by pressing it with spring tension.
  Each wire is guided along the edge of the side surface through low-friction pulleys 
  to avoid interference with other wires and components, and exits the robot through a wire relay unit.
  The wire relay unit allows the wire to be routed in any direction around its origin.
  This structure is symmetrically installed on the left and right sides.
  The maximum continuous tension is about 120 N, and the maximum winding length is about 6 m.
  The wires are made of Vectran\textregistered, 
  a high-performance rope of about 1.0 mm diameter made from high-strength polyarylate fiber.

  The wire for tool attachment originates from the groove at the center of the rear side (\figref{fig:wire_overview}\ctext{5}).
  It is wound by a wire winding module mounted at the center, whose details are shown in \figref{fig:wire_module} (right).
  This wire winding module consists of a servo motor, a winch, and a pressing roller.
  The winch has a diameter of 22 mm and a width of 40 mm, and a small Robstride00 actuator of the same family 
  as the joint actuators was adopted as the servo motor.
  The maximum continuous tension is about 420 N, and the maximum winding length is about 6 m.
  The wire is made of Dyneema\textregistered, 
  which has a higher strength than the environmental anchoring wires and a diameter of about 2.0 mm.
}%
{%
  \begin{figure}[t]
    \begin{center}
      \includegraphics[width=1.0\columnwidth]{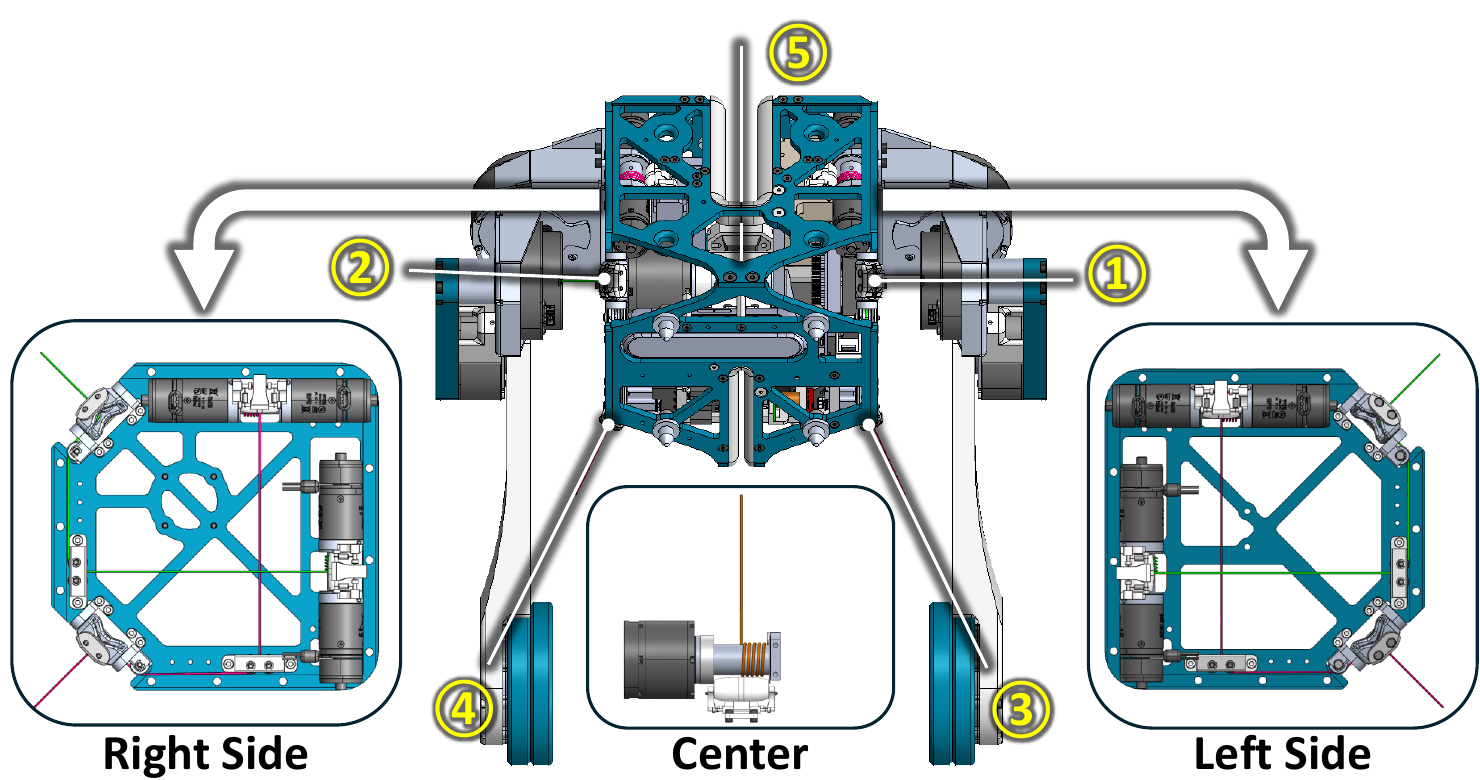}
      \vspace{-4.0ex}
      \caption{
        Arrangement of the wire winding modules on the main body.
        WiXus is equipped with a total of five modules: 
        four modules for environmental anchoring are mounted on the sides, 
        while a single module for tool attachment is located at the center.
      }
      \vspace{-2.0ex}
      \label{fig:wire_overview}
    \end{center}
  \end{figure}

  \begin{figure}[t]
    \begin{center}
      \includegraphics[width=1.0\columnwidth]{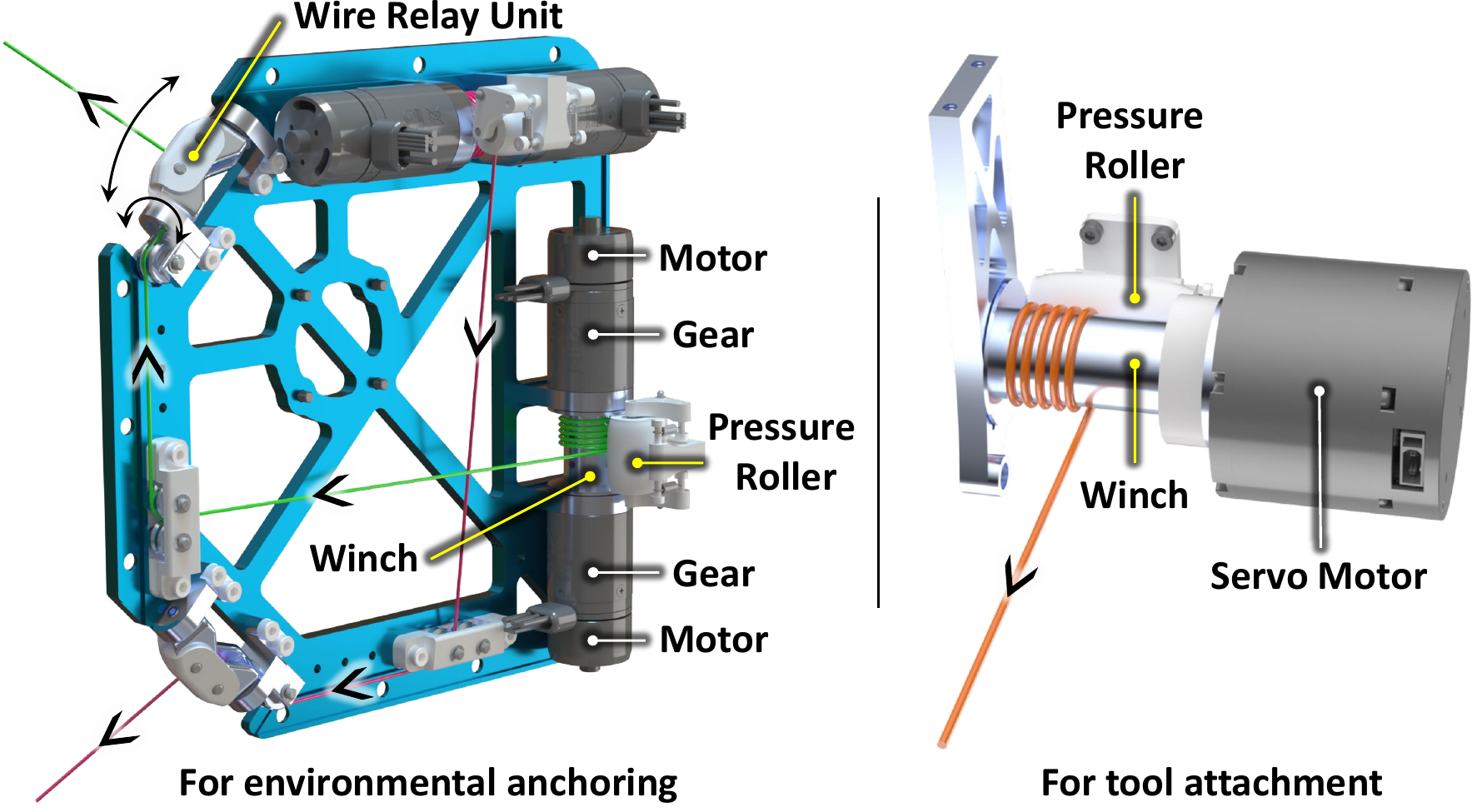}
      \vspace{-4.0ex}
      \caption{
        Overall structure of the two types of wire winding modules: 
        one for environmental anchoring and one for tool attachment. 
        A motor drives a winch to wind the wire.
      }
      \vspace{-4.0ex}
      \label{fig:wire_module}
    \end{center}
  \end{figure}
  
  WiXusにおけるワイヤとワイヤ巻取りモジュールの位置関係を\figref{fig:wire_overview}に示す。
  環境に接続するためのワイヤが4本、道具を合体させるためのワイヤが1本の
  計5本のワイヤをWiXusは巻き取ることができる。

  環境接続用のワイヤは、前面の各頂点 (\figref{fig:wire_overview}\ctext{1}--\ctext{4}) が出発点である。
  各ワイヤは両側面に2つずつ取り付けられたワイヤ巻取りモジュールから巻き取られる。
  \figref{fig:wire_overview}\ctext{1}, \ctext{3}のワイヤが左側面、
  \figref{fig:wire_overview}\ctext{2}, \ctext{4}のワイヤが右側面のワイヤ巻取りモジュールに対応している。

  \figref{fig:wire_module}左はその側面の詳細図である。
  ワイヤ巻取りモジュールは、ワイヤを巻き取るウィンチと、それを回転させるモータ2つ、ワイヤを押さえるローラからなる。
  ウィンチは$\phi$15 mm、幅21.5 mmであり、その両側それぞれに取り付けた2つのモータで回転する。
  モータには筒型でありCAN通信で制御可能なM2006を採用した。
  押さえローラは巻き取ったワイヤをウィンチに整列した状態で止めておく役割を持ち、バネ張力でワイヤをウィンチへ押し付けている。
  ワイヤは他のワイヤや部品に干渉しないようにプーリで低摩擦にガイドされながら側面の辺を通り、
  ワイヤ経由ユニットを介してロボット外へ出される。
  ワイヤ経由ユニットには、ワイヤを出発点を中心に任意の向きへ出すという役割がある。
  これらの構造が左側面と右側面で対称に搭載されている。
  最大連続張力は約120 N、最大巻取り長さは約6 mである。
  なお、巻き取るワイヤには高強力ポリアレート繊維を使用した線径が約1.0 mmの高性能ロープであるVectran\textregistered
  を採用した。

  道具合体用のワイヤは
  後面の中央の溝 (\figref{fig:wire_overview}\ctext{5}) が出発点である。
  ワイヤは中央に取り付けられたワイヤ巻取りモジュールから巻き取られ、
  \figref{fig:wire_module}右がその詳細図である。
  ワイヤ巻取りモジュールは、モータとウィンチ、押さえローラからなる。
  ウィンチは$\phi$22 mm、幅40 mmであり、
  モータには関節駆動に採用したものと同系統で小型であるRobstride00を採用した。
  最大連続張力は約420 N、最大巻取り長さは約6 mである。
  なお、巻き取るワイヤには環境接続用のワイヤよりも強度が高い、線径が約2.0 mmのDyneema\textregistered
  を採用した。
}%

\subsection{Onboard Circuits and Architecture}
\switchlanguage%
{%
  \begin{figure}[t]
    \begin{center}
      \includegraphics[width=1.0\columnwidth]{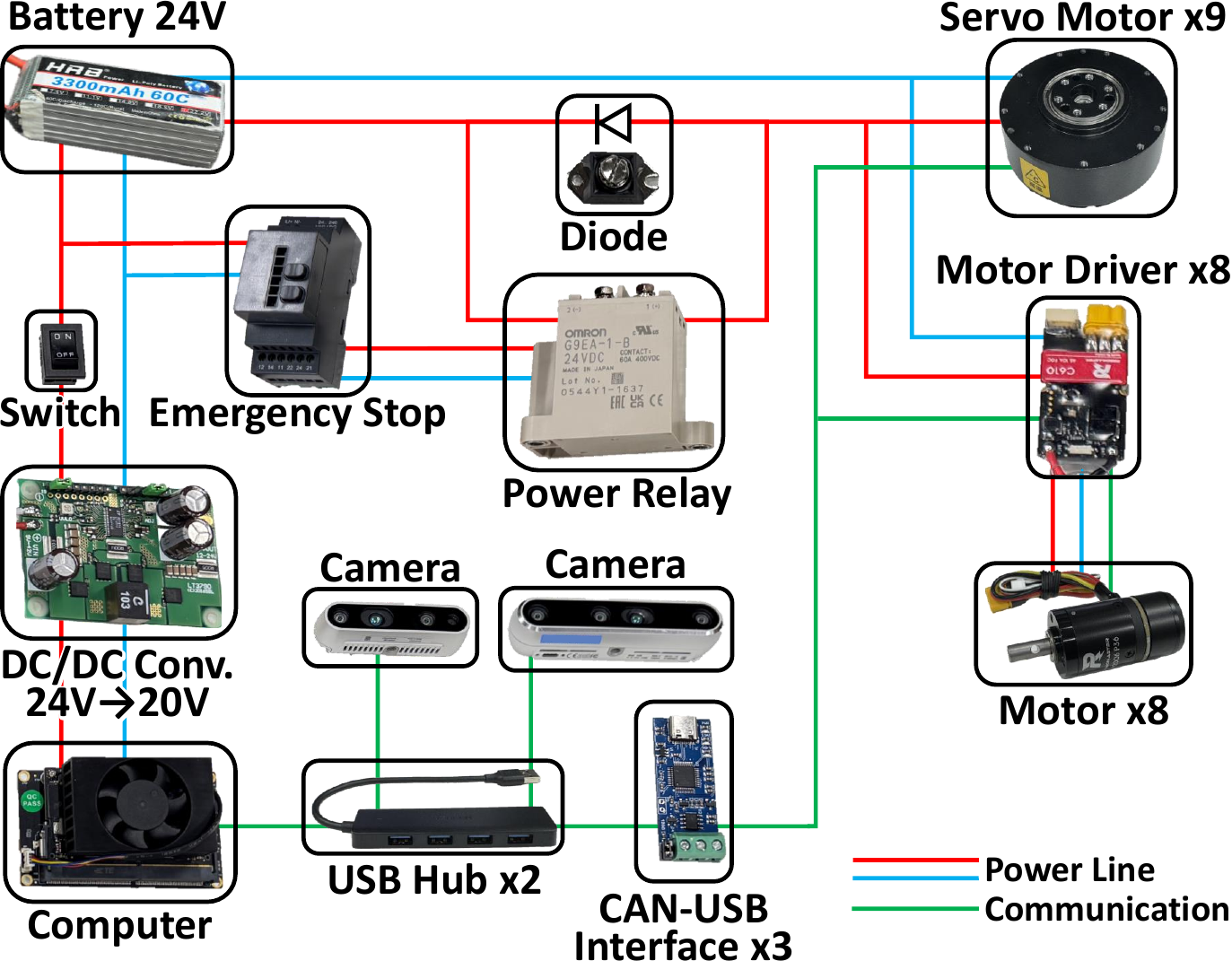}
      \vspace{-4.0ex}
      \caption{
        Circuit configuration of WiXus. 
        A total of 17 motors (eight for the wheeled-legs and nine for the wire-driven system) 
        are controlled by a computer via three CAN-USB interfaces. 
        The system also includes two RGB-D cameras with onboard IMUs.
      }
      \vspace{-3.0ex}
      \label{fig:circuit_configuration}
    \end{center}
  \end{figure}

  \begin{figure}[t]
    \begin{center}
      \includegraphics[width=1.0\columnwidth]{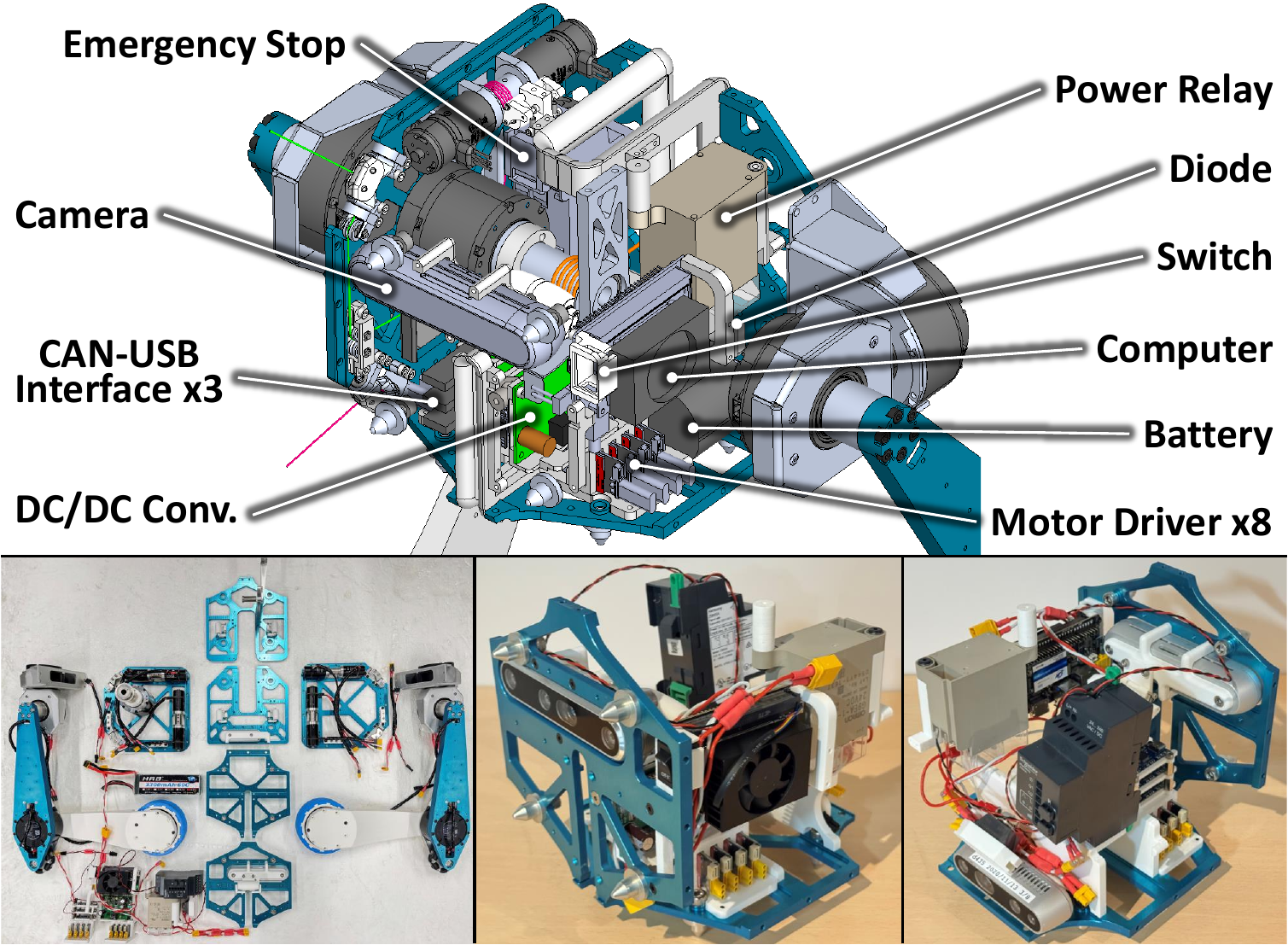}
      \vspace{-4.0ex}
      \caption{
        Layout of the electronic components.
        The components are housed within the main body, 
        arranged to prevent any interference with the wires.
      }
      \vspace{-5.0ex}
      \label{fig:actual_configuration}
    \end{center}
  \end{figure}

  The circuit configuration of WiXus is shown in \figref{fig:circuit_configuration}.
  A 24 V battery is connected to the computer (Jetson Orin Nano) through a power switch and a DC–DC converter.
  Two cameras (Intel RealSense D455 and D435i) and three CAN-USB interfaces are connected to the computer via two USB hubs.
  The D435i is not used in this study.
  The computer communicates with each servo motor and motor driver through the CAN-USB interfaces.
  The same battery also powers the motors through a power relay.
  The power relay is controlled by a wireless emergency stop system, 
  which allows the robot to be remotely stopped in case of abnormal behavior during experiments.
  In addition, a current path with a diode is implemented to allow 
  regenerative current from the motors to flow back to the battery.
  Note that the motors for the joints, wheels, and tool attachment are servo motors with integrated motor drivers.

  As shown in the upper part of \figref{fig:actual_configuration}, 
  these circuit components are placed inside the robot without interfering with the wires.
  This circuit configuration is implemented in the actual robot as shown in the lower part of \figref{fig:actual_configuration}.
  \figref{fig:actual_configuration} (lower left) shows the fully disassembled robot laid out, 
  and \figref{fig:actual_configuration} (lower right) shows only the assembled portion around the circuit components.
}%
{%
  \begin{figure}[t]
    \begin{center}
      \includegraphics[width=1.0\columnwidth]{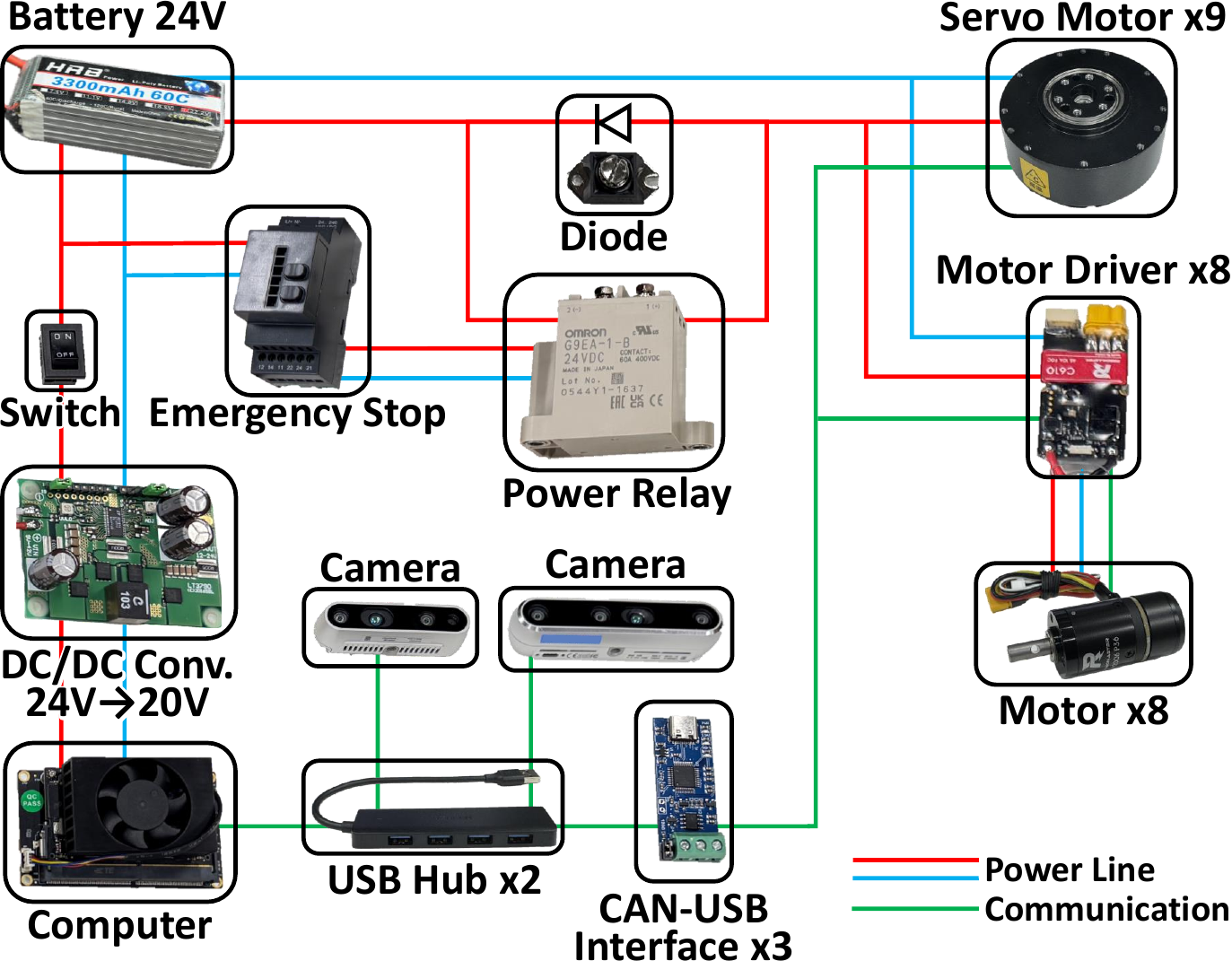}
      \vspace{-4.0ex}
      \caption{
        Circuit configuration of WiXus. 
        A total of 17 motors (eight for the wheeled-legs and nine for the wire-driven system) 
        are controlled by a computer via three CAN-USB interfaces. 
        The system also includes two RGB-D cameras with onboard IMUs.
      }
      \vspace{-3.0ex}
      \label{fig:circuit_configuration}
    \end{center}
  \end{figure}

  \begin{figure}[t]
    \begin{center}
      \includegraphics[width=1.0\columnwidth]{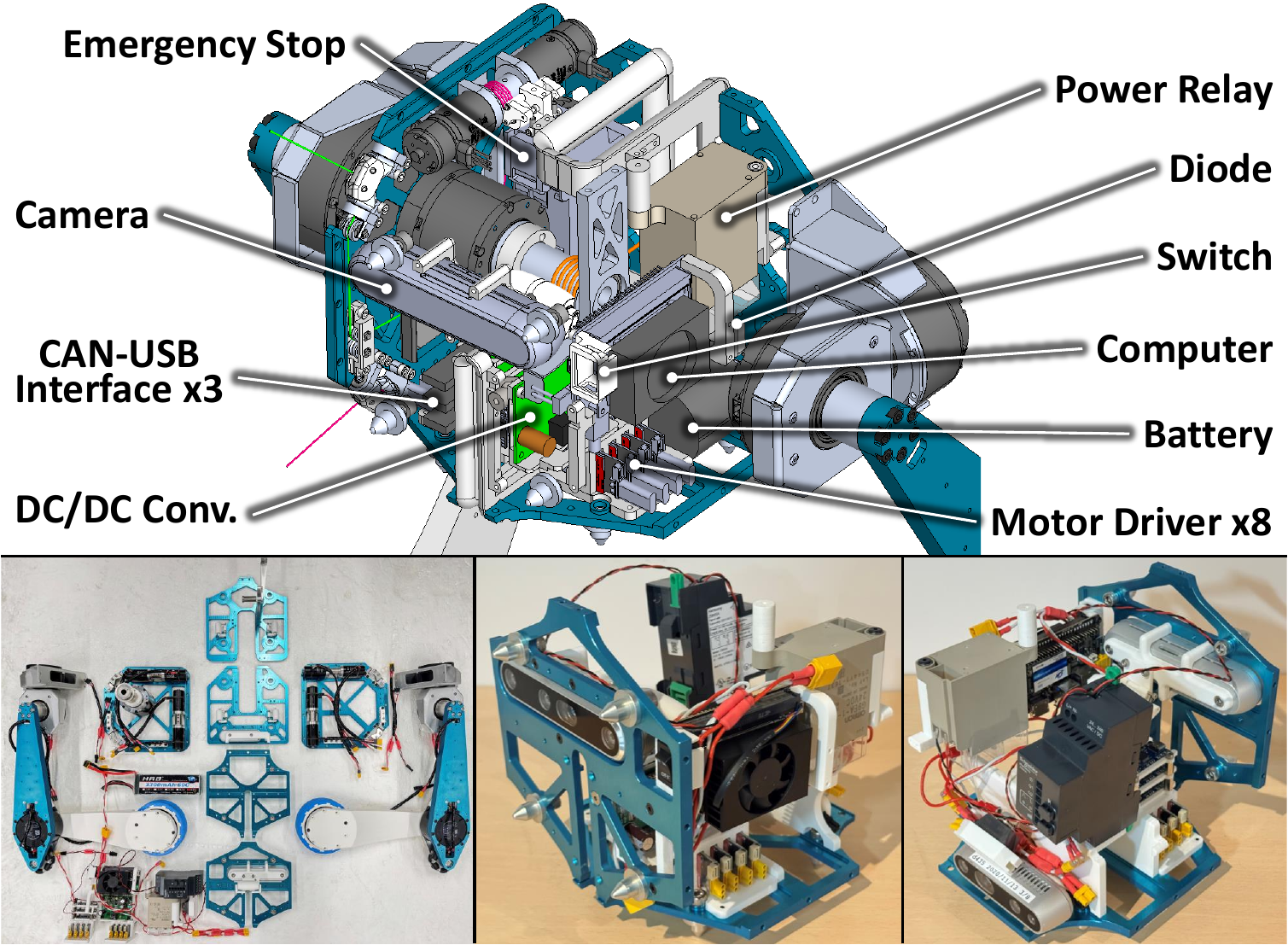}
      \vspace{-4.0ex}
      \caption{
        Layout of the electronic components.
        The components are housed within the main body, 
        arranged to prevent any interference with the wires.
      }
      \vspace{-2.0ex}
      \label{fig:actual_configuration}
    \end{center}
  \end{figure}

  % 電装図をバッテリーから説明していく
  WiXusの回路構成図を\figref{fig:circuit_configuration}に示す。
  24 VのバッテリがスイッチとDCDCコンバータを介してコンピュータ (Jetson Orin Nano) に接続される。
  コンピュータには2つのUSBハブを介して、2つのカメラ (Intel Realsense D455, D435i) と
  3つのCAN-USBインターフェースが接続される。本論文ではD435iは使用していない。
  コンピュータはCAN-USBインターフェースを介して各サーボモータ、モータドライバと通信を行う。

  また、同バッテリが電源リレーを介して各モータにも接続される。
  電源リレーは無線緊急停止装置により駆動され、実験時の異常動作を遠隔で停止できる。
  さらに、モータの逆起電力をバッテリへ回生するための経路をダイオードを用いて実現している。
  なお、関節駆動、車輪、道具合体用のモータはモータドライバと一体となったサーボモータである。

  \figref{fig:actual_configuration}上に示される通り、
  ロボット体内には上記の回路部品がワイヤと干渉することなく収められている。
  上記の回路構成を実機では\figref{fig:actual_configuration}下に示すとおり実装した。
  \figref{fig:actual_configuration}左下は実機ロボットを展開して並べたものであり、
  \figref{fig:actual_configuration}右下は回路部品の周辺のみを組み立てたものである。
}%

\subsection{Optional Sub-Module: Flying Anchor}
\switchlanguage%
{%
  \begin{figure}[t]
    \begin{center}
      \includegraphics[width=1.0\columnwidth]{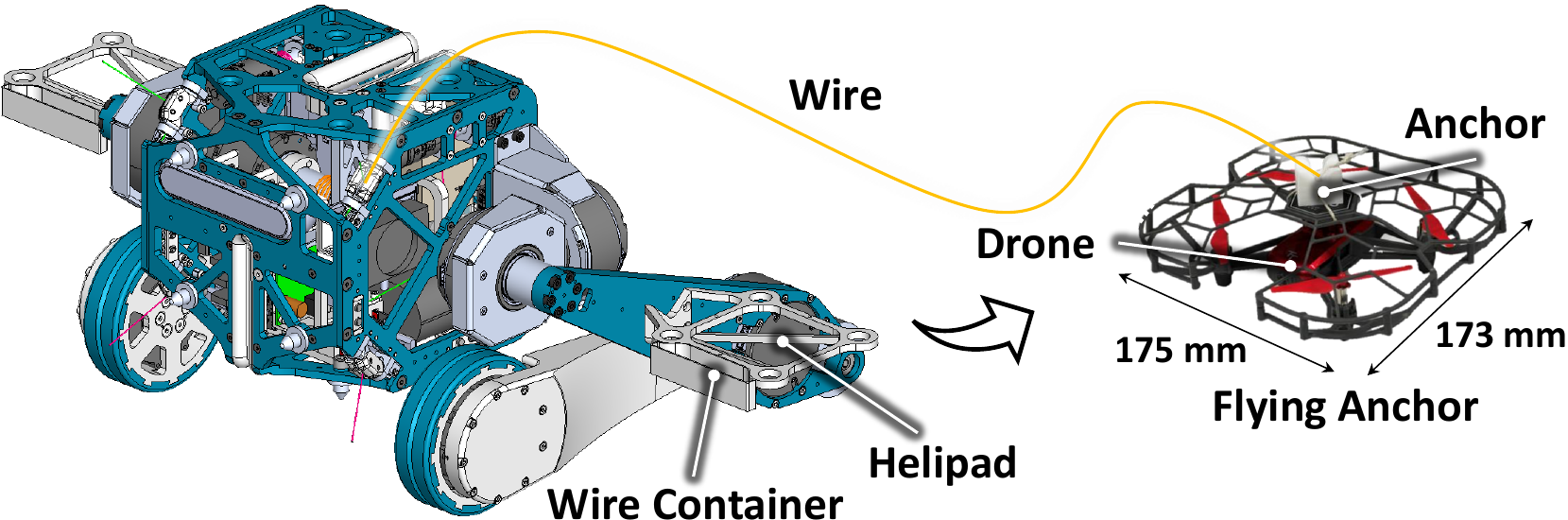}
      \vspace{-4.0ex}
      \caption{
        The flying anchor, a custom drone for autonomous wire anchoring to the environment.
        Attached to the tip of the wire, this module combines a drone with an anchor.
      }
      \vspace{-3.0ex}
      \label{fig:flying_anchor}
    \end{center}
  \end{figure}

  As a method for autonomously anchoring wires to the environment, the flying anchor \cite{inoue2024cubix} has been proposed.
  WiXus can also autonomously anchor its wires to the environment using the flying anchor.
  As shown in \figref{fig:flying_anchor}, the flying anchor consists of an anchor attached to the tip of a wire and a drone.
  After the flying anchor flies around the target structure to wrap the wire around it, 
  winding the wire secures the anchor in place.
  When using the flying anchor to anchor a wire to the environment, wire containers and helipads are mounted on both legs.
  The wire container stores the wire that has been previously unwound from the winch and bundled,
  and the helipad serves as a launch pad for the flying anchor.

  As in the method for controlling the flying anchor using an RGB-D camera \cite{inoue2025anchor}, 
  the flying anchor is autonomously controlled by WiXus itself while being observed by WiXus's camera (D455).
}%
{%
  \begin{figure}[t]
    \begin{center}
      \includegraphics[width=1.0\columnwidth]{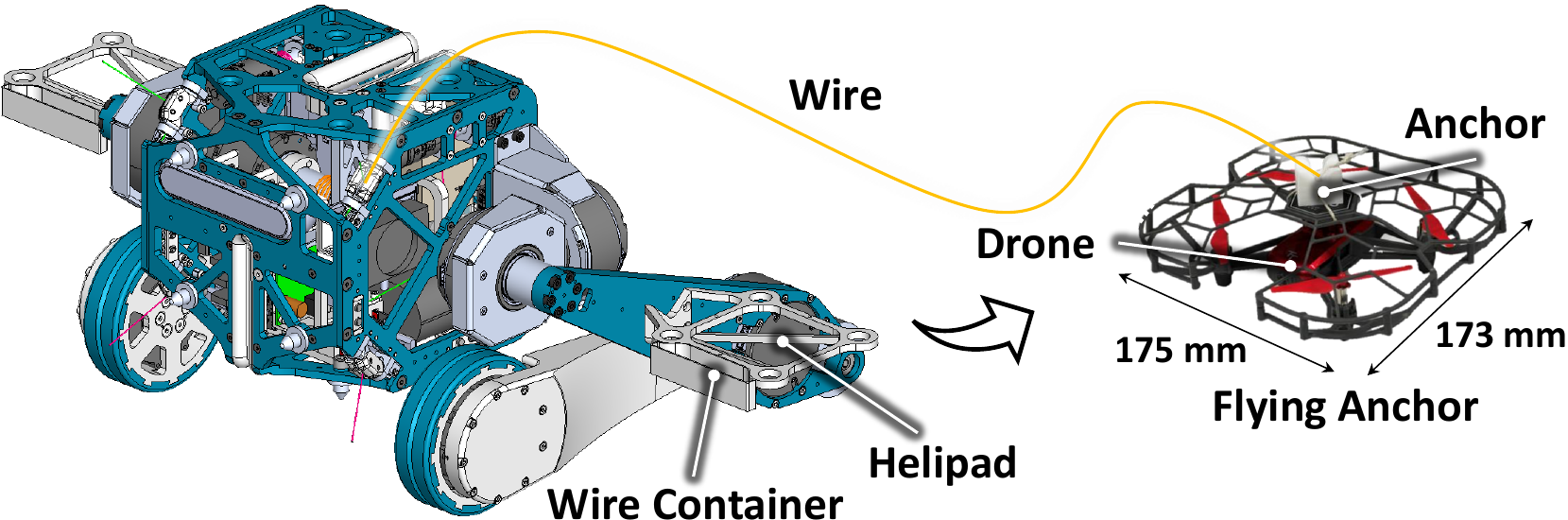}
      \vspace{-4.0ex}
      \caption{
        The flying anchor, a custom drone for autonomous wire anchoring to the environment.
        Attached to the tip of the wire, this module combines a drone with an anchor.
      }
      \vspace{-3.0ex}
      \label{fig:flying_anchor}
    \end{center}
  \end{figure}

  ロボットが自律的に環境にワイヤを接続する手段として飛行アンカー\cite{inoue2024cubix}が提案されてきた。
  WiXusについても飛行アンカーを用いてワイヤを自律的に環境へ接続できる。
  飛行アンカーは\figref{fig:flying_anchor}に示される通り、
  ワイヤの先端に取り付けたアンカーとドローンを組み合わせたものである。
  飛行アンカーがワイヤを接続対象に巻き付けるように飛行した後、
  ワイヤを巻き取るとアンカーがその巻き付けを固定する。

  環境へのワイヤ接続に飛行アンカーを用いる際には、両脚にワイヤコンテナとヘリパッドを装着する。
  ワイヤコンテナは予め巻き出して束ねたワイヤを収納する部分であり、
  ヘリパッドは飛行アンカーの発射台となる。

  RGB-Dカメラを用いて飛行アンカーを制御する手法\cite{inoue2025anchor}と同様、
  飛行アンカーはWiXusのカメラ (D455) で観測しながらWiXus自身で自律制御を行う。
}%

\section{Controller and System of WiXus} \label{sec:software}
\subsection{Control System Architecture}
\switchlanguage%
{%
  \begin{figure}[t]
    \begin{center}
      \includegraphics[width=1.0\columnwidth]{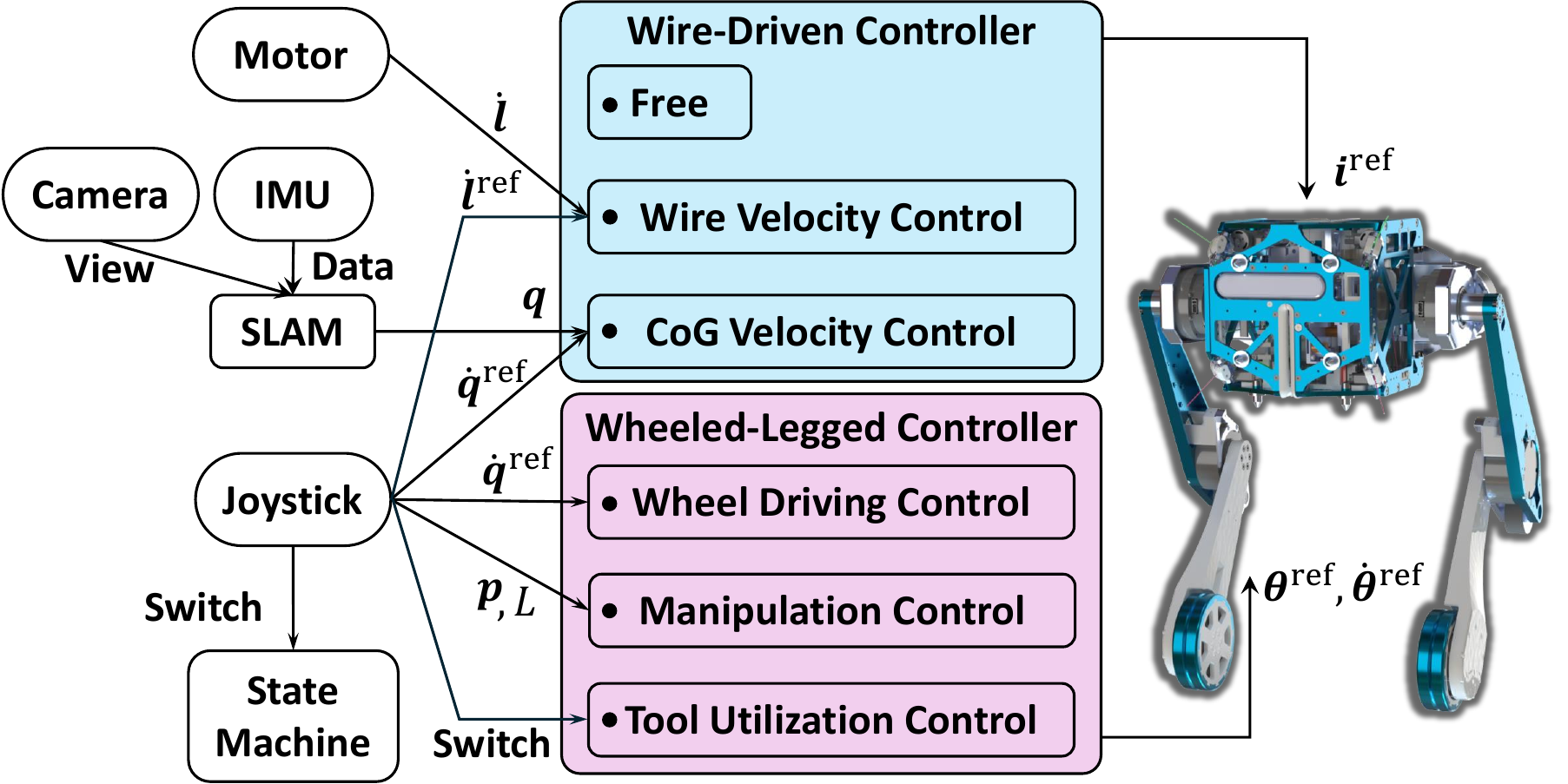}
      \vspace{-4.0ex}
      \caption{
        The software architecture of WiXus.
        The system features two main control modules that operate in parallel: 
        a wire-driven controller and a wheeled-legged controller.
      }
      \vspace{-5.0ex}
      \label{fig:software}
    \end{center}
  \end{figure}

  \begin{figure}[t]
    \begin{center}
      \includegraphics[width=1.0\columnwidth]{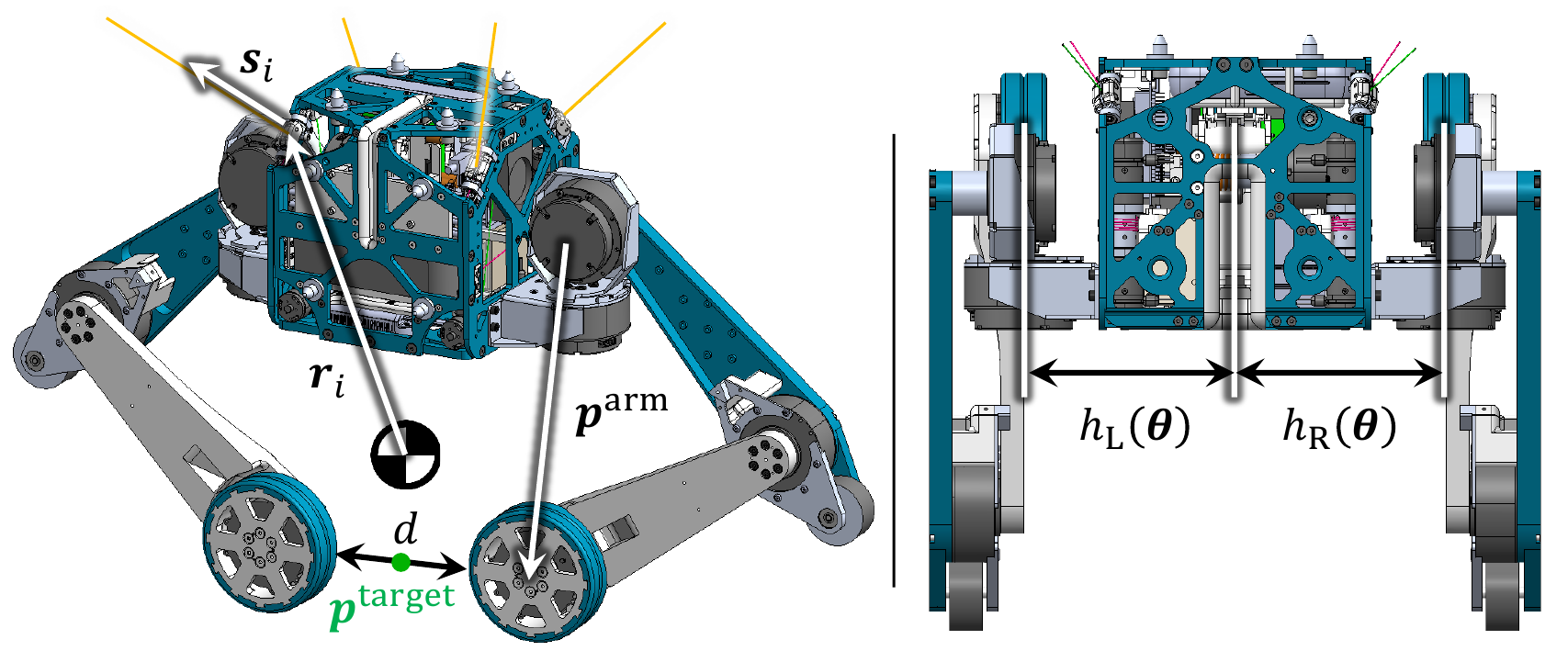}
      \vspace{-4.0ex}
      \caption{
        Definition of symbols used in the wire-driven, manipulation, and wheeled locomotion controllers.
      }
      \vspace{-3.0ex}
      \label{fig:control_character}
    \end{center}
  \end{figure}

  The system architecture of WiXus is shown in \figref{fig:software}.
  Two types of controllers---the wire-driven controller and the wheeled-legged controller---are executed in parallel, 
  constantly controlling all actuators.
  In addition, RTAB-Map \cite{rtabmap2019} for SLAM is running separately using the D455 camera.
  Furthermore, a state machine implemented with SMACH \cite{bohren2010smach} is used to manage the mode transitions 
  of the two controllers.
  Command inputs to each controller and mode switching are performed through joysticks 
  (Sony DualSense and 3Dconnexion SpaceNavigator).
}%
{%
  \begin{figure}[t]
    \begin{center}
      \includegraphics[width=1.0\columnwidth]{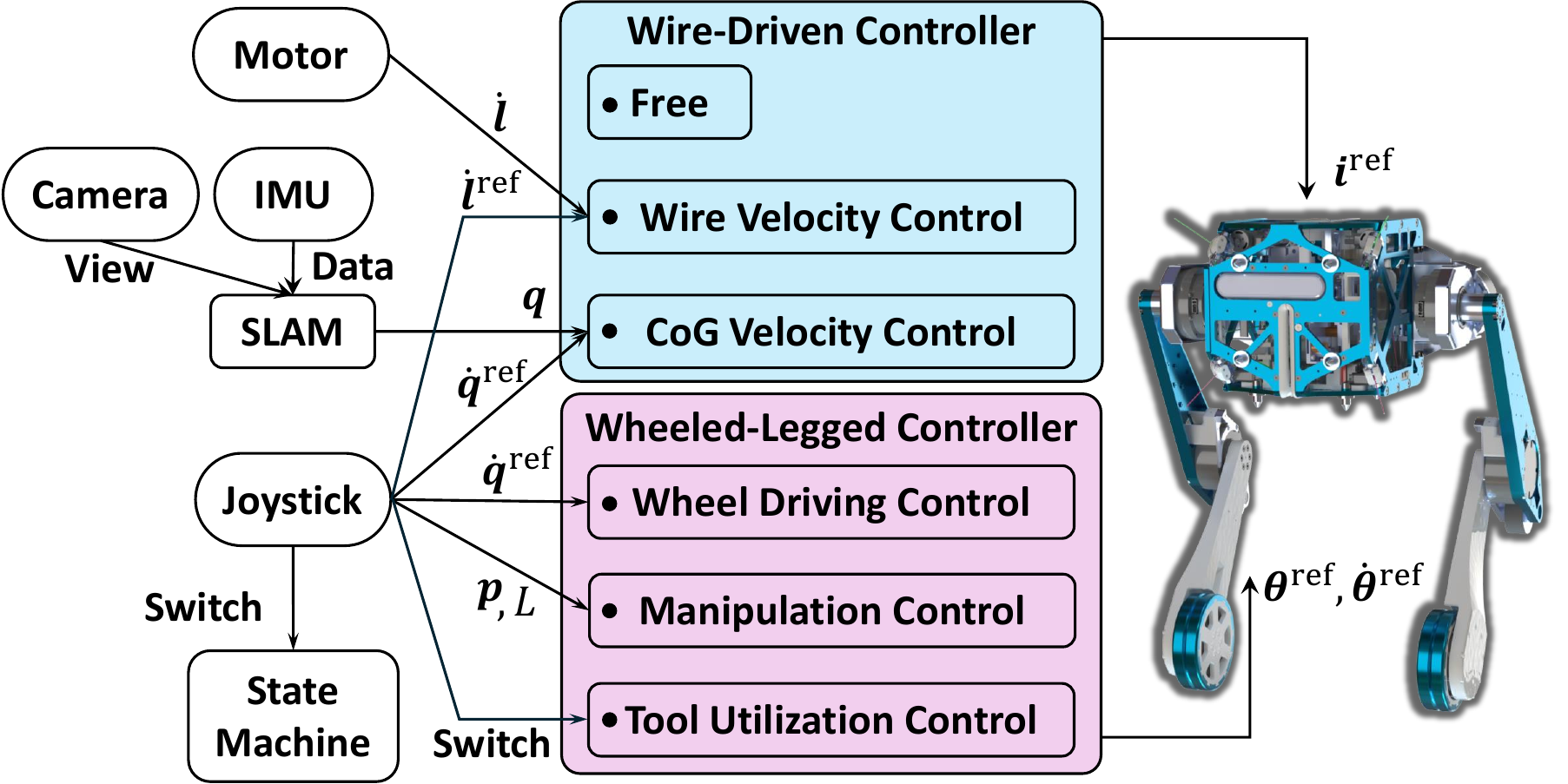}
      \vspace{-4.0ex}
      \caption{
        The software architecture of WiXus.
        The system features two main control modules that operate in parallel: 
        a wire-driven controller and a wheeled-legged controller.
      }
      \vspace{-2.0ex}
      \label{fig:software}
    \end{center}
  \end{figure}

  \begin{figure}[t]
    \begin{center}
      \includegraphics[width=1.0\columnwidth]{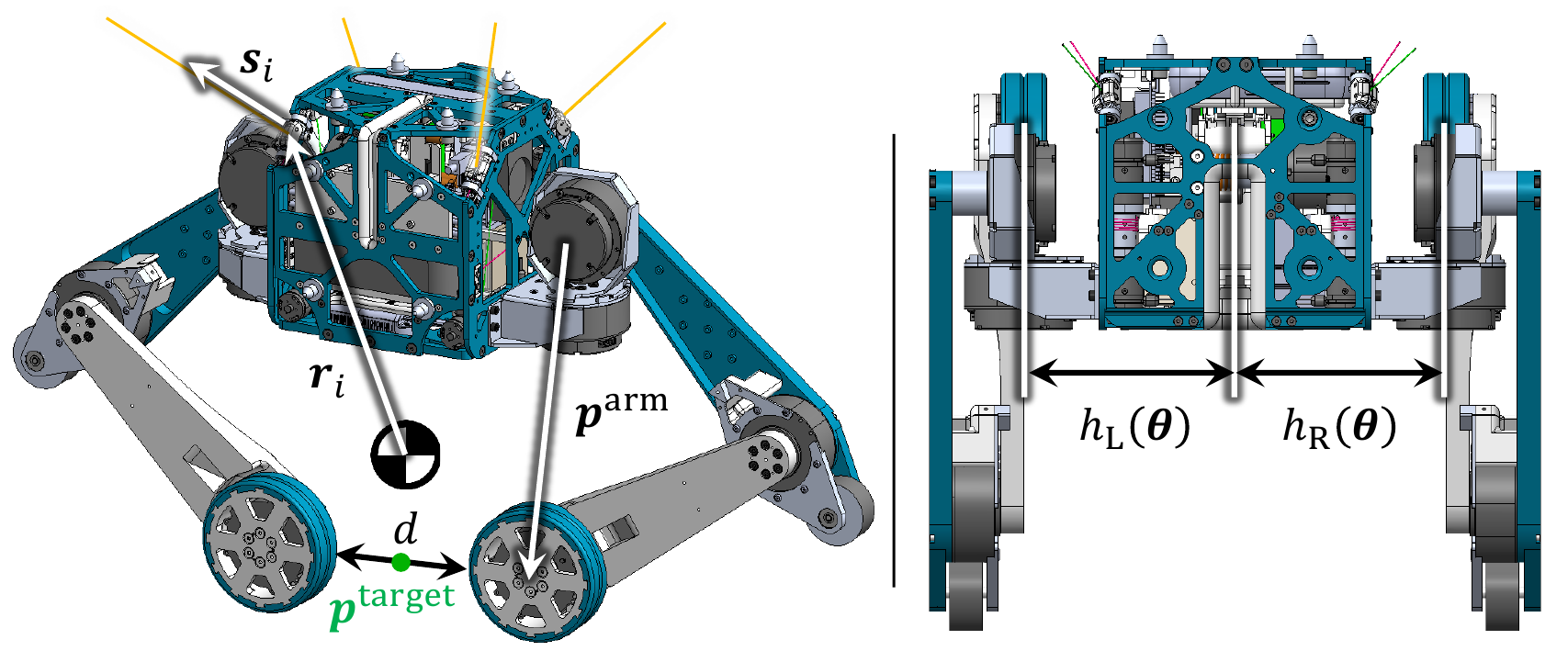}
      \vspace{-4.0ex}
      \caption{
        Definition of symbols used in the wire-driven, manipulation, and wheeled locomotion controllers.
      }
      \vspace{-3.0ex}
      \label{fig:control_character}
    \end{center}
  \end{figure}

  WiXusのシステム構成図を\figref{fig:software}に示す。
  ワイヤ駆動制御器と脚車輪制御の2種類の制御器がパラレルに実行されており、
  全アクチュエータを常に制御している。
  さらに、それらとは別に、D455を用いてRTAB-Map\cite{rtabmap2019}によるSLAMが実行されている。
  また、2種類の制御器それぞれのモード遷移を管理するためにSMACH\cite{bohren2010smach}によるステートマシンを用いている。
  各制御器への指令値入力や、モードの切り替えはJoystick (Sony DualSenseや3Dconnecxion SpaceNavigator) を介して行う。
}%

\subsection{Wire-Driven Control}
\switchlanguage%
{%
  The wire-driven controller has three modes: \textit{free}, \textit{wire velocity control}, 
  and \textit{center-of-gravity (CoG) velocity control}, and it controls the wires anchored to the environment.
  
  In the \textit{free} mode, each motor of the wire winding modules is always commanded with a current of 0 A.
  This mode is activated when only locomotion by the wheeled-legs is performed and wire-driven actuation is not used.
  
  In the \textit{wire velocity control} mode, the target wire velocity $\dot{\bm{l}}$ is given as input.
  The controller computes and outputs the target tensions $\bm{f}^\mathrm{ref}$ using P control 
  so that the actual wire velocities $\dot{\bm{l}}$ approach the targets while compensating for gravity.
  If the number of wires anchored to the environment is $m$, 
  the target tension $f_i^\mathrm{ref}$ of the $i$-th wire is calculated as:
  % \begin{equation}
  %   f_i^\mathrm{ref} = \cfrac{M\|\bm{g}\|}{m} + \mathrm{P}\left(\dot{l}_i \to \dot{l}_i^\mathrm{ref}\right)
  %   \label{eq:wire_vel}
  % \end{equation}
  \begin{equation}
    f_i^\mathrm{ref} = \cfrac{M\|\bm{g}\|}{m} + K_p\left(\dot{l}_i-\dot{l}_i^\mathrm{ref}\right)
    \label{eq:wire_vel}
  \end{equation}
  % where $M$ is the mass of the robot and $\bm{g}$ is the gravitational acceleration vector.
  where $M$ is the mass of the robot, $\bm{g}$ is the gravitational acceleration vector, and $K_p$ is the proportional gain.
  Because this control distributes the gravity compensation tension equally to each wire, 
  it does not require the robot's posture information, but it is only useful when the wires are nearly vertical.
  
  In the \textit{CoG velocity control} mode, the target CoG velocity $\dot{\bm{q}}^\mathrm{ref}$ is given as input.
  The controller computes and outputs the target tensions $\bm{f}^\mathrm{ref}$ 
  so that the robot follows the target while compensating for gravity, as:
  \begin{equation}
    % \bm{f}^\mathrm{ref} = \bm{f}^g + \mathrm{P}\left(\dot{\bm{l}}\to-\bm{W}^\top(\bm{q})\dot{\bm{q}}^\mathrm{ref}\right)
    \bm{f}^\mathrm{ref} = \bm{f}^g + K_p\left(\dot{\bm{l}}+\bm{W}^\top(\bm{q})\dot{\bm{q}}^\mathrm{ref}\right)
    \label{eq:cog_vel_control}
  \end{equation}
  Here, $\bm{W}(\bm{q})$ is a $6 \times m$ Jacobian matrix expressed using the position vector $\bm{r}_i$ 
  of the wire origin and the unit direction vector $\bm{s}_i$ (as shown in \figref{fig:control_character} left):
  \begin{equation}
    \bm{W}(\bm{q}) =
    \left[ \begin{array}{ccc}
      \bm{s}_{1} & \cdots & \bm{s}_{m} \\
      \bm{r}_{1}\times\bm{s}_{1} & \cdots & \bm{r}_{m}\times\bm{s}_{m}
    \end{array} \right]
    \label{eq:wire_matrix}
  \end{equation}
  The first term of \equref{eq:cog_vel_control} is the gravity compensation term $\bm{f}^g$, 
  computed by solving the following quadratic programming problem:
  \begin{equation}
  \begin{aligned}
    & \underset{\bm{f}^\mathrm{g}}{\text{min}}
    % & & \|\bm{f}^\mathrm{g}\|^2 + \left[{\bm{W}(\bm{q})}{\bm{f}^\mathrm{g}-M\bm{g}}\right]^\top \bm{\Lambda} \left[{\bm{W}(\bm{q})}{\bm{f}^\mathrm{g}-M\bm{g}}\right]\\
    & & \|{\bm{W}(\bm{q})}{\bm{f}^\mathrm{g}-M\bm{g}}\|^2\\
    & \text{s.t.}
    & & \bm{f}^\mathrm{min} \leq \bm{f}^\mathrm{g} \leq \bm{f}^\mathrm{max}
  \end{aligned}
  \label{eq:quad_prog}
  \end{equation}
  % where $\bm{\Lambda}$ is a weight matrix and $\bm{f}^\mathrm{min},\bm{f}^\mathrm{max}$ 
  where $\bm{f}^\mathrm{min},\bm{f}^\mathrm{max}$ are the minimum and maximum tension limits.
  The second term of \equref{eq:cog_vel_control} is the CoG velocity tracking term, 
  which converts the target CoG velocity $\dot{\bm{q}}^\mathrm{ref}$ to the target wire velocity 
  using the Jacobian $\bm{W}(\bm{q})$, 
  and applies P control so that the actual wire velocity $\dot{\bm{l}}$ follows it.
  All tensions and wire velocities are represented as $m$-dimensional vectors for the $m$ anchored wires.
  
  The motors (M2006) used in the anchoring wire winding modules are small and have gearheads with a reduction ratio of 36:1, 
  which causes friction effects during control.
  Therefore, the computed tensions are converted into current commands $\bm{i}^\mathrm{ref}$ 
  with compensation for both Coulomb friction and load-dependent friction as:
  \begin{equation}
    \bm{i}^\mathrm{ref} = r\bm{K_t}^{-1}\bm{f}^\mathrm{ref}+\bm{i_0}+\bm{i_L}\cfrac{\|\bm{f}^\mathrm{ref}\|}{M\|\bm{g}\|}
    \label{eq:friction_comp}
  \end{equation}
  where $r$ is the winch radius, $\bm{K_t}$ is a diagonal matrix of motor torque constants,
  $\bm{i_0}$ is the Coulomb friction compensation current, and $\bm{i_L}$ is the load-dependent friction coefficient.
  $\bm{i_0}$, $\bm{i_L}$, and $\bm{K_t}$ are identified from static current measurements under no-load and suspended conditions, 
  based on the balance of friction and gravity (details omitted due to space).

  % To obtain $\bm{i_0}$, $\bm{i_L}$, and $\bm{K_t}$, three types of currents are measured for each wire:
  % \begin{enumerate}
  %   \item $\bm{i_0}$: The current at which the wire starts to be wound in the no-load state.
  %   \item $\bm{i^\mathrm{up}}$: The current at which the robot starts to ascend when the current is increased 
  %     while it is suspended by a single wire.
  %   \item $\bm{i^\mathrm{down}}$: The current at which the robot starts to descend when the current is decreased 
  %     while it is suspended by a single wire.
  % \end{enumerate}
  % Assuming that friction acts against the motion, 
  % and letting $\bm{i^\mathrm{mg}}$ be the current to balance gravity, the following model is used:
  % \begin{equation}
  %   \begin{aligned}
  %     \bm{i^\mathrm{up}} &= \bm{i^\mathrm{mg}} + \bm{i_0} + \bm{i_L} \\
  %     \bm{i^\mathrm{down}} &= \bm{i^\mathrm{mg}} - \bm{i_0} - \bm{i_L}
  %   \end{aligned}
  % \end{equation}
  % Solving for $\bm{i^\mathrm{mg}}$ and $\bm{i_L}$ gives:
  % \begin{equation}
  %   \begin{aligned}
  %     \bm{i^\mathrm{mg}} &= \cfrac{\bm{i^\mathrm{up}}  + \bm{i^\mathrm{down}}}{2} \\
  %     \bm{i_L} &= \cfrac{\bm{i^\mathrm{up}}  - \bm{i^\mathrm{down}}}{2} - \bm{i_0}
  %   \end{aligned}
  % \end{equation}
  % Then $\bm{K_t}$ is calculated as:
  % \begin{equation}
  %   \bm{K_t} = rM\|\bm{g}\|\mathrm{diag}\left(\bm{i^\mathrm{mg}}\right)^{-1}
  % \end{equation}
  
  Each anchoring wire winding module has two motors per winch, 
  which are controlled as one motor by applying the same magnitude currents with opposite signs.
}%
{%
  ワイヤ駆動制御はFreeとワイヤ速度制御、重心速度制御の3モードであり、
  環境に接続するワイヤに関する制御を行う。

  Freeモードでは常にワイヤ巻取りモジュールの各モータに0 Aが司令される。
  これは、脚車輪のみの移動を行う場合など、ワイヤ駆動は行わない状態で起動される。

  ワイヤ速度制御モードでは、目標ワイヤ速度$\dot{\bm{l}}$を入力に持ち、
  重力補償をしながらワイヤ速度$\bm{l}$を目標へ近づける目標張力$\bm{f}^\mathrm{ref}$をP制御で計算し出力する。
  ここで、環境に接続したワイヤ本数を$m$とすると、環境に接続した$i$番目のワイヤの目標張力$f_i^\mathrm{ref}$は以下の式で書ける。
  \begin{equation}
    f_i^\mathrm{ref} = \cfrac{M\|\bm{g}\|}{m} + \mathrm{P}\left(\dot{l}_i \to \dot{l}_i^\mathrm{ref}\right)
    % \bm{f}^\mathrm{ref} = \cfrac{M\|\bm{g}\|}{m} + \mathrm{P}\left(\dot{\bm{l}} \to \dot{\bm{l}}^\mathrm{ref}\right)
    \label{eq:wire_vel}
  \end{equation}
  ただし、$M$はロボットの質量であり、$\bm{g}$は重力加速度ベクトルである。
  この制御では重力補償張力を各ワイヤへ均等に配分するため、ロボットの姿勢情報を必要としない代わりに、
  ワイヤの向きが鉛直上向きに近い場合にのみにしか有用でないことに注意する。

  重心速度制御モードでは、重心の目標速度$\dot{\bm{q}}^\mathrm{ref}$を入力に持ち、
  重力補償をしながらそれに追従するワイヤ目標張力$\bm{f}^\mathrm{ref}$を以下の式で計算し出力する。
  \begin{equation}
    \bm{f}^\mathrm{ref} = \bm{f}^g + \mathrm{P}\left(\dot{\bm{l}}\to-\bm{W}^\top(\bm{q})\dot{\bm{q}}^\mathrm{ref}\right)
    \label{eq:cog_vel_control}
  \end{equation}
  ただし、$\bm{W}(\bm{q})$はサイズが$6 \times m$のヤコビアンであり、
  \figref{fig:control_character}左に示したワイヤの出発点と向きに関する位置ベクトル
  $\bm{r}_i$と$\bm{s}_i$を用いて以下のように表される。
  \begin{equation}
    \begin{split}
      \bm{W}(\bm{q}) =
      \left[ \begin{array}{ccc}
        \bm{s}_{1} & \cdots & \bm{s}_{m} \\
        \bm{r}_{1}\times\bm{s}_{1} & \cdots & \bm{r}_{m}\times\bm{s}_{m}
      \end{array} \right]
    \end{split}
    \label{eq:wire_matrix}
  \end{equation}
  \equref{eq:cog_vel_control}の第1項は重力補償項であり、次の二次計画問題を解くことで計算する。
  \begin{equation}
    \begin{aligned}
      & \underset{\bm{f}^\mathrm{g}}{\text{min}}
      & & \|\bm{f}^\mathrm{g}\|^2 + \left[{\bm{W}(\bm{q})}{\bm{f}^\mathrm{g}-M\bm{g}}\right]^\top \bm{\Lambda} \left[{\bm{W}(\bm{q})}{\bm{f}^\mathrm{g}-M\bm{g}}\right]\\
      & \text{s.t.}
      & & \bm{f}^\mathrm{min} \leq \bm{f}^\mathrm{g} \leq \bm{f}^\mathrm{max} \\
    \end{aligned}
    \label{eq:quad_prog}
  \end{equation}
  ここで、$\bm{\Lambda}$は重み行列であり、$\bm{f}^\mathrm{min},\bm{f}^\mathrm{max}$はワイヤ張力の最小、最大値ベクトルである。
  \\\equref{eq:cog_vel_control}の第2項は
  目標重心速度追従項であり、ヤコビアン$\bm{W}(\bm{q})$を用いて目標重心速度$\dot{\bm{q}}$を
  目標ワイヤ速度に変換し、ワイヤ速度$\dot{\bm{l}}$がそれに近づくようにP制御を行って計算する。
  なお、上記の各張力やワイヤ速度は環境に接続した$m$本のワイヤについてまとめた$m$次元ベクトルである。

  環境接続用の巻取りモジュールに採用したモータであるM2006は、
  小型モータのために減速比36:1のギアヘッドが搭載されており制御時に摩擦の影響を受ける。
  そこで、各モードで計算した張力はワイヤ巻取りモジュールの定摩擦と荷重比例摩擦を補償しながら、
  以下の式で司令電流$\bm{i}^\mathrm{ref}$に変換する。
  下記の電流$\bm{i}$についても$m$本のワイヤについてまとめた$m$次元ベクトルである。
  \begin{equation}
    \bm{i}^\mathrm{ref} = r\bm{K_t}^{-1}\bm{f}^\mathrm{ref}+\bm{i_0}+\bm{i_L}\cfrac{\|\bm{f}^\mathrm{ref}\|}{M\|\bm{g}\|}
    \label{eq:friction_comp}
  \end{equation}
  \equref{eq:friction_comp}の第1項はウィンチ半径$r$と各モータのトルク定数を対角成分に並べた対角行列$\bm{K_t}$を用いた
  、摩擦のないモータにおける電流と力の変換項である。
  \equref{eq:friction_comp}の第2項は負荷時摩擦を補正する項$\bm{i_0}$である。
  \equref{eq:friction_comp}の第3項は荷重比例摩擦を補償する項であり、$\bm{i_L}$はその係数である。

  ここで、$\bm{i_0}$、$\bm{i_L}$、$\bm{K_t}$を入手する手法について説明する。
  まず、実機の各ワイヤについて下記3種類の電流を計測する。
  \begin{enumerate}
    \item $\bm{i_0}$: 無負荷状態で、電流を増やした時にワイヤが巻き取られ始める電流。
    \item $\bm{i^\mathrm{up}}$: 1本のワイヤで自身を懸垂した状態で、電流を増やした時に上昇し始める電流。
    \item $\bm{i^\mathrm{down}}$: 1本のワイヤで自身を懸垂した状態で、電流を減らした時に下降し始める電流。
  \end{enumerate}
  摩擦力は動作を妨げる方向に作用することに注意して、
  自重との釣り合いに必要な張力を発生させる電流$\bm{i^\mathrm{mg}}$を用いた以下のモデルを仮定する。
  \begin{equation}
    \begin{aligned}
      \bm{i^\mathrm{up}} &= \bm{i^\mathrm{mg}} + \bm{i_0} + \bm{i_L} \\
      \bm{i^\mathrm{down}} &= \bm{i^\mathrm{mg}} - \bm{i_0} - \bm{i_L}
    \end{aligned}
  \end{equation}
  これを$\bm{i^\mathrm{mg}}$と$\bm{i_L}$に付いて解くと、以下のようになる。
  \begin{equation}
    \begin{aligned}
      \bm{i^\mathrm{mg}} &= \cfrac{\bm{i^\mathrm{up}}  + \bm{i^\mathrm{down}}}{2} \\
      \bm{i_L} &= \cfrac{\bm{i^\mathrm{up}}  - \bm{i^\mathrm{down}}}{2} - \bm{i_0}
    \end{aligned}
  \end{equation}
  $\bm{K_t}$は上記$\bm{i^\mathrm{mg}}$を用いて以下のように計算できる。
  \begin{equation}
    \bm{K_t} = rM\|\bm{g}\|\mathrm{diag}\left(\bm{i^\mathrm{mg}}\right)^{-1}
  \end{equation}

  環境接続用のワイヤ巻取りモジュールは、1つのウィンチにつき2つのモータが取り付けられているが、
  逆符号で同値の電流を同時に流すことで、1つのモータとして扱う。
}%

\subsection{Wheeled-Legged Control}
\switchlanguage%
{%
	The wheeled-legged controller has three modes: 
  \textit{wheel driving control}, \textit{manipulation control}, and \textit{tool utilization control}.
	
	In the \textit{wheel driving control} mode, 
  when the robot takes a vehicle form with its wheels and support wheels contacting the ground, 
  the controller outputs the target wheel angular velocities 
  $\dot{\theta}_\mathrm{L\,wheel}^\mathrm{ref},\,\dot{\theta}_\mathrm{R\,wheel}^\mathrm{ref}$ 
  that track the target translational velocity $\dot{q}_x^\mathrm{ref}$ and 
  the target yaw angular velocity $\dot{q}_\mathrm{yaw}^\mathrm{ref}$, as:
	\begin{equation}
	  \begin{aligned}
      \dot{\theta}_\mathrm{L\,wheel}^\mathrm{ref} &= \cfrac{\dot{q}_x^\mathrm{ref} - \dot{q}_\mathrm{yaw}^\mathrm{ref}h_\mathrm{L}(\bm{\theta})}{R} \\
      \dot{\theta}_\mathrm{R\,wheel}^\mathrm{ref} &= \cfrac{\dot{q}_x^\mathrm{ref} + \dot{q}_\mathrm{yaw}^\mathrm{ref}h_\mathrm{R}(\bm{\theta})}{R}
	  \end{aligned}
	\end{equation}
	where $R$ is the wheel radius and $h_\mathrm{L,R}(\bm{\theta})$ is the distance from the robot center to each wheel, 
  as shown in \figref{fig:control_character} (right).
	The target joint angles $\bm{\theta}^\mathrm{ref}$ of the other joints are fixed in the vehicle posture, 
  except for the Hip-P joint, which can be operated to change the pitch angle of the Base-Link.
	
	In the \textit{manipulation control} mode, as shown in \figref{fig:control_character} (left), 
  both wheeled-legs grasp and manipulate the target object.
	The inputs are the target position $\bm{p}^\mathrm{target}$ between the wheels 
  and the width $d$ (shown in \figref{fig:control_character} left).
	Only the Hip-P joint angle $\theta_\text{Hip-P}$ and the Knee-P joint angle $\theta_\text{Knee-P}$ are controlled, 
  and the two leg tips always move on the same plane.
	In this configuration, each wheeled-leg behaves as a typical 2-DOF arm, 
  and their target joint angles can be computed analytically by solving the inverse kinematics as:
	\begin{equation}
	  \begin{aligned}
	    \theta_\text{Knee-P}^\mathrm{ref} &= \arccos\cfrac{\|\bm{p}^\mathrm{arm}\|^2 - L_\mathrm{thigh}^2 - L_\mathrm{calf}^2}{2L_\mathrm{thigh}L_\mathrm{calf}} \\
	    \theta_\text{Hip-P}^\mathrm{ref} &= \arctan\cfrac{\bm{p}^\mathrm{arm}_y}{\bm{p}^\mathrm{arm}_x} - \arctan\cfrac{L_\mathrm{calf}\sin\theta_\text{Knee-P}^\mathrm{ref}}{L_\mathrm{thigh} + L_\mathrm{calf}\cos\theta_\text{Knee-P}^\mathrm{ref}}
	  \end{aligned}
	\end{equation}
	where $\bm{p}^\mathrm{arm}$ is the target tip position vector 
  geometrically calculated from $\bm{p}^\mathrm{target}$, $d$, and $R$, 
  and $L_\mathrm{thigh}$ and $L_\mathrm{calf}$ are the link lengths shown in \figref{fig:control_character} (left).
	In addition, the wheel angular velocities can also be controlled to lift the object by rotating the wheels
  when grasping it with the wheels.
	
	In the \textit{tool utilization control} mode, 
  the controller outputs the target joint angles $\theta^\mathrm{ref}$ of the wheeled-legs for using the tool.
	For example, in the tool utilization experiment described in \subsecref{subsec:exp_tool}, loppers are used.
	Using the loppers requires opening and closing motions.
	The joint angles when opening and closing the loppers with the wheeled-legs are recorded in advance, 
  and the controller switches between these joint angles to achieve the tool utilization.
}%
{%
  脚車輪制御は車輪駆動制御、物体操作制御、道具利用制御の3モードである。

  車輪駆動制御モードでは、車輪と補助輪を地面に接する車両系になった際の、
  目標並進速度$\dot{q}_x^\mathrm{ref}$と目標角速度$\dot{q}_\mathrm{yaw}^\mathrm{ref}$に追従する
  目標輪速$\dot{\theta}_\mathrm{L\,wheel}^\mathrm{ref},\,\dot{\theta}_\mathrm{R\,wheel}^\mathrm{ref}$を以下の式で出力する。
  \begin{equation}
    \begin{aligned}
      \dot{\theta}_\mathrm{L\,wheel}^\mathrm{ref} &= \cfrac{\dot{q}_x^\mathrm{ref} - \dot{q}_\mathrm{yaw}^\mathrm{ref}h_L(\bm{\theta})}{R} \\
      \dot{\theta}_\mathrm{R\,wheel}^\mathrm{ref} &= \cfrac{\dot{q}_x^\mathrm{ref} + \dot{q}_\mathrm{yaw}^\mathrm{ref}h_R(\bm{\theta})}{R}
    \end{aligned}
  \end{equation}
  ただし、$R$は車輪半径、$h_{L,R}(\bm{\theta})$は\figref{fig:control_character}右に示されるロボット中心から車輪までの距離である。
  他の関節における目標関節角度$\bm{\theta}^\mathrm{ref}$は車両形態をなす姿勢で固定であるが、
  Hip-P関節のみはbase linkのPitch角度を変更するために操作可能とした。

  物体操作制御モードでは、\figref{fig:control_character}左に示される姿勢で、
  両脚車輪で対象物体を把持、操作する。
  入力は\figref{fig:control_character}左に示される両車輪の間の位置$\bm{p}^\mathrm{target}$と幅$d$である。
  Hip-P関節角度$\theta_\text{Knee-P}$とKnee-P関節角度$\theta_\text{Knee-P}$のみを制御対象とし、両脚先は常に同一平面上を移動する。
  この時、脚車輪1つあたりの系は一般的な2自由度アームと等しくなるので、
  それらの目標関節角度は以下のように解析的に逆運動学を解くことで決定できる。
  \begin{equation}
    \begin{aligned}
      \theta_\text{Knee-P}^\mathrm{ref} &= \arccos\cfrac{\|\bm{p}^\mathrm{arm}\|^2 - L_\mathrm{thigh}^2 - L_\mathrm{calf}^2}{2L_\mathrm{thigh}L_\mathrm{calf}} \\
      \theta_\text{Hip-P}^\mathrm{ref} &= \arctan\cfrac{\bm{p}^\mathrm{arm}_y}{\bm{p}^\mathrm{arm}_x} - \arctan\cfrac{L_\mathrm{calf}\sin\theta_\text{Knee-P}^\mathrm{ref}}{L_\mathrm{thigh} + L_\mathrm{calf}\cos\theta_\text{Knee-P}^\mathrm{ref}}
    \end{aligned}
  \end{equation}
  ただし、$\bm{p}^\mathrm{arm}$は$\bm{p}^\mathrm{target},\,d,\,R$から幾何的に計算可能な脚先の目標位置ベクトル、
  $L_\mathrm{thigh},\,L_\mathrm{calf}$は各リンクの長さであり、それぞれ\figref{fig:control_character}左で示した。
  なお、物体を把持する際に車輪を回転させることで物体を巻き上げられるため、輪速も操作可能とした。

  道具利用制御モードでは、使用する道具を利用するための脚車輪の目標関節角度$\theta^\mathrm{ref}$を出力する。
  例えば、本論文の\subsecref{subsec:exp_tool}で扱う道具利用実験では、刈込鋏を利用する。
  刈込鋏の利用には、それを閉じる動作と開く動作を行う必要がある。
  刈込鋏を脚車輪を用いて開閉した際の関節角度を予め記録し、
  それらを交互に遷移する目標関節角度$\theta^\mathrm{ref}$を計算することで刈込鋏を利用する。
}%

\section{Experiments} \label{sec:experiments}
\subsection{Mobility Capabilities}
\switchlanguage%
{%
  To demonstrate the mobility capabilities of WiXus, we conducted the following two experiments.
}%
{%
  WiXusの移動能力を示すために以下の2種類の実験を行う。
}%

\subsubsection{Planar Mobility and Mapping}
\switchlanguage%
{%
  In this experiment, WiXus creates a map while moving on a flat surface using its wheels.
	% The experimental environment and the mapping process are shown in \figref{fig:expA-1},
	% and the time series data of the velocities $q_x,\,q_y,\,q_\mathrm{yaw},\,q_x^\mathrm{ref},\,q_y^\mathrm{ref},\,q_\mathrm{yaw}^\mathrm{ref}$
	% and the wheel velocities $\dot{\theta}_\mathrm{L\,wheel},\,\dot{\theta}_\mathrm{R\,wheel}$ are shown in \figref{fig:expA-1_data}.
	As shown in \figref{fig:expA-1}, 
  WiXus builds a map of the surrounding environment and estimates its own position and velocity 
  using RTAB-Map \cite{rtabmap2019} with the D455 camera while moving on wheels.
  Here, the wire-driven controller is in the \textit{free} mode, and the wheeled-legged controller 
  is in the \textit{wheel driving control} mode.
	In addition, as shown in \figref{fig:expA-1}\ctext{3}, 
  the pitch angle of the Base-Link is changed, 
  which also changes the pitch direction of the camera's field of view used for SLAM.
	
	Looking at \figref{fig:expA-1_data}, the translational velocity $q_x$ is achieved as commanded through 
  the wheel velocities $\dot{\theta}_\mathrm{L\,wheel},\,\dot{\theta}_\mathrm{R\,wheel}$.
	Although there is some error in the rotational velocity $q_\mathrm{yaw}$, 
  considering that $q_y$ (which should be 0 m/s) is not zero, it is likely due to estimation errors in the SLAM system.
	In this study, the self-localization uses only the RGB-D images and IMU sensor data from the D455, 
  but incorporating wheel velocity or wire velocity measurements could lead to more stable estimation.
	
	For a robot driven by anchoring wires to the environment, 
  observing and recording the environment is an important preparatory task,
	and this experiment shows that WiXus can perform it by itself.
	In the experiments in \subsecref{subsec:exp_rescue} and \subsecref{subsec:exp_tool},
	the environments to which the wires are anchored are assumed to be known in advance,
	and this prior information is obtained by WiXus itself through such observation and recording.

  \begin{figure}[t]
    \begin{center}
      \includegraphics[width=1.0\columnwidth]{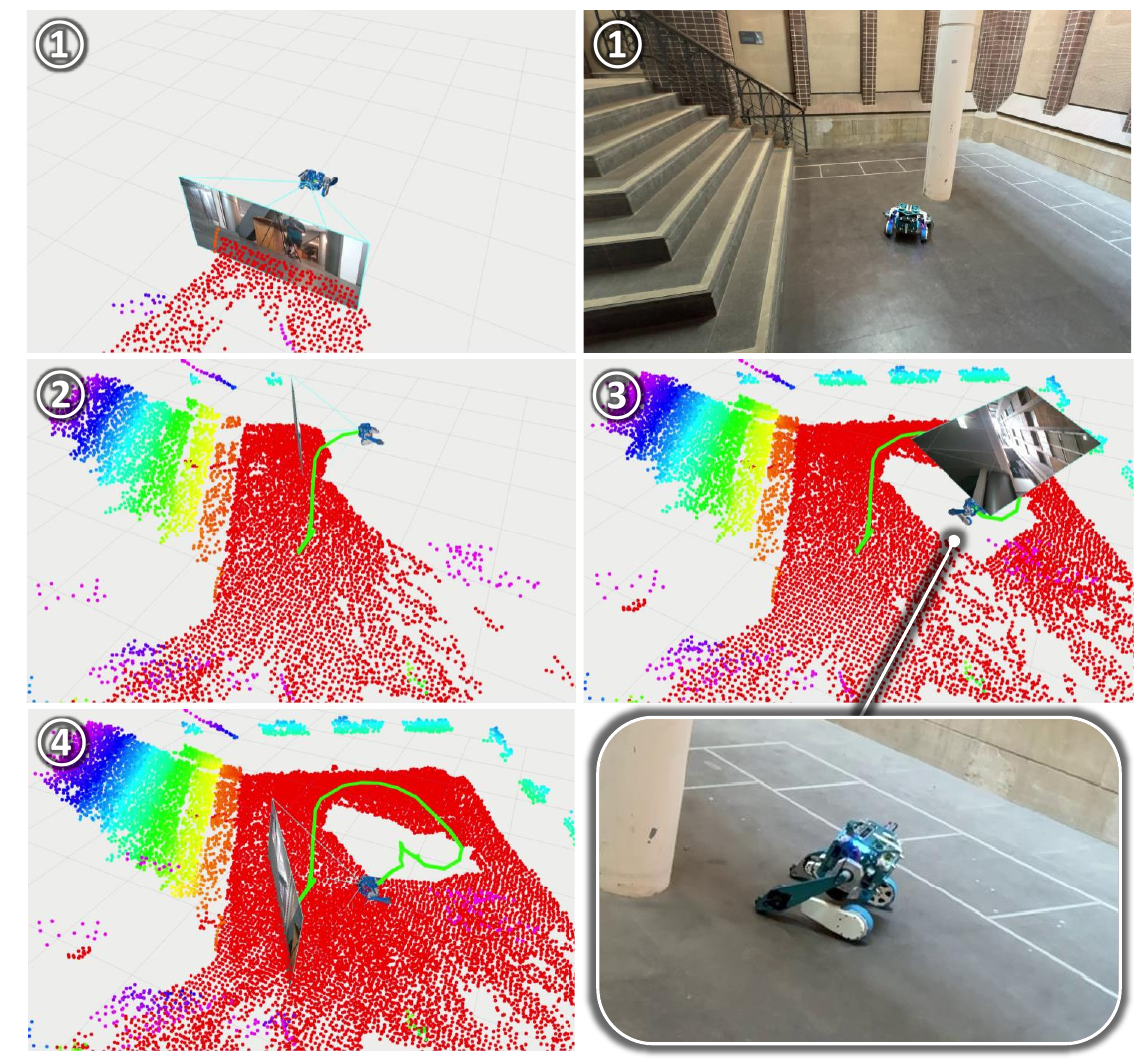}
      \vspace{-5.0ex}
      \caption{
        Scenes from the planar mobility and mapping experiment. 
        WiXus can generate a map while performing 
        wheeled locomotion and controlling the pitch angle of its main body.
      }
      \vspace{-2.0ex}
      \label{fig:expA-1}
    \end{center}
  \end{figure}

  \begin{figure}[t]
    \begin{center}
      \includegraphics[width=1.0\columnwidth]{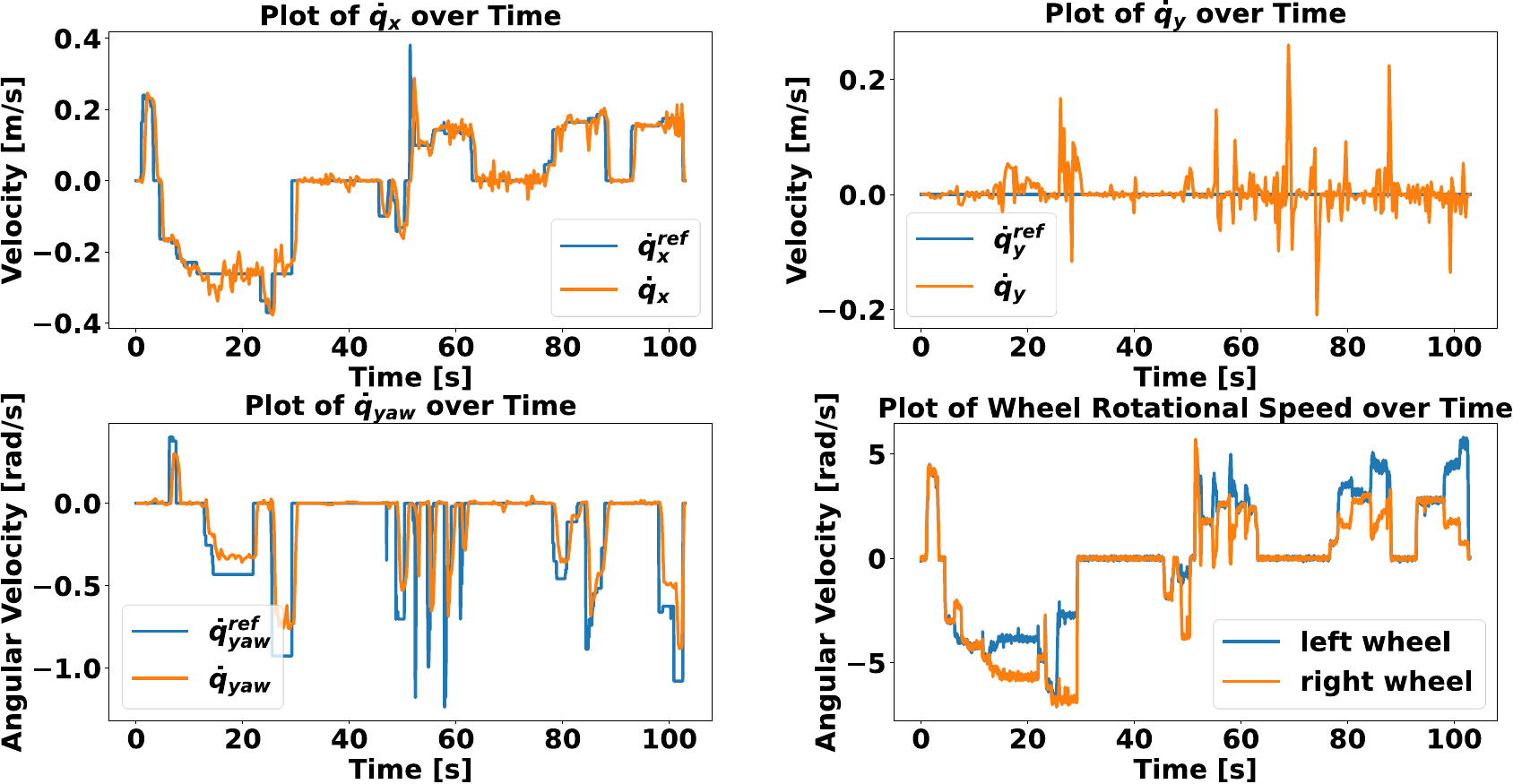}
      \vspace{-4.0ex}
      \caption{
         Time-series data of the velocity and actual wheel speeds during mapping. 
         The wheel speeds are calculated to track the commanded velocity.
      }
      \vspace{-4.0ex}
      \label{fig:expA-1_data}
    \end{center}
  \end{figure}

}%
{%
  \begin{figure}[t]
    \begin{center}
      \includegraphics[width=1.0\columnwidth]{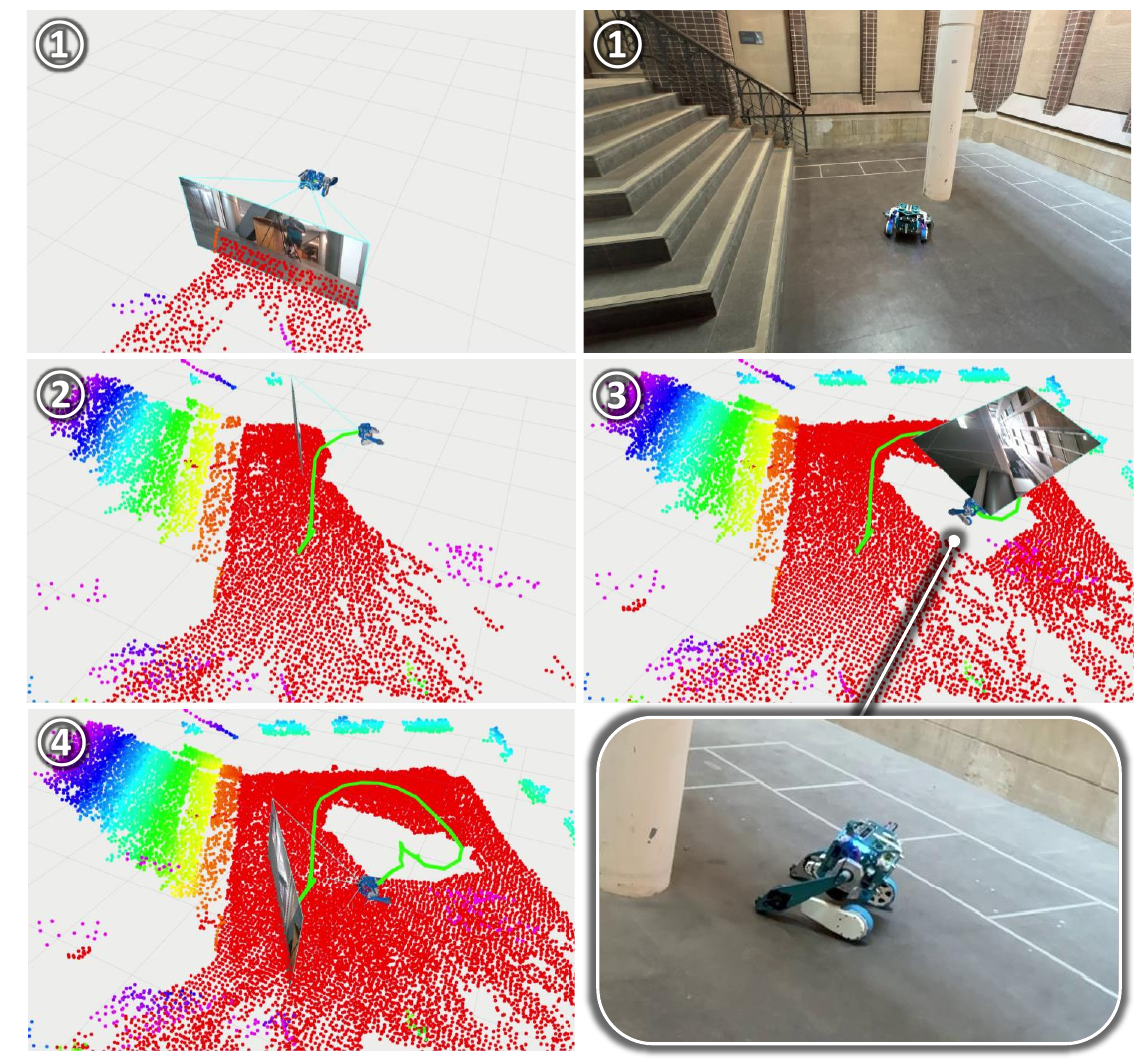}
      \vspace{-5.0ex}
      \caption{
        Scenes from the planar mobility and mapping experiment. 
        WiXus can generate a map while performing 
        wheeled locomotion and controlling the pitch angle of its main body.
      }
      \vspace{-2.0ex}
      \label{fig:expA-1}
    \end{center}
  \end{figure}

  \begin{figure}[t]
    \begin{center}
      \includegraphics[width=1.0\columnwidth]{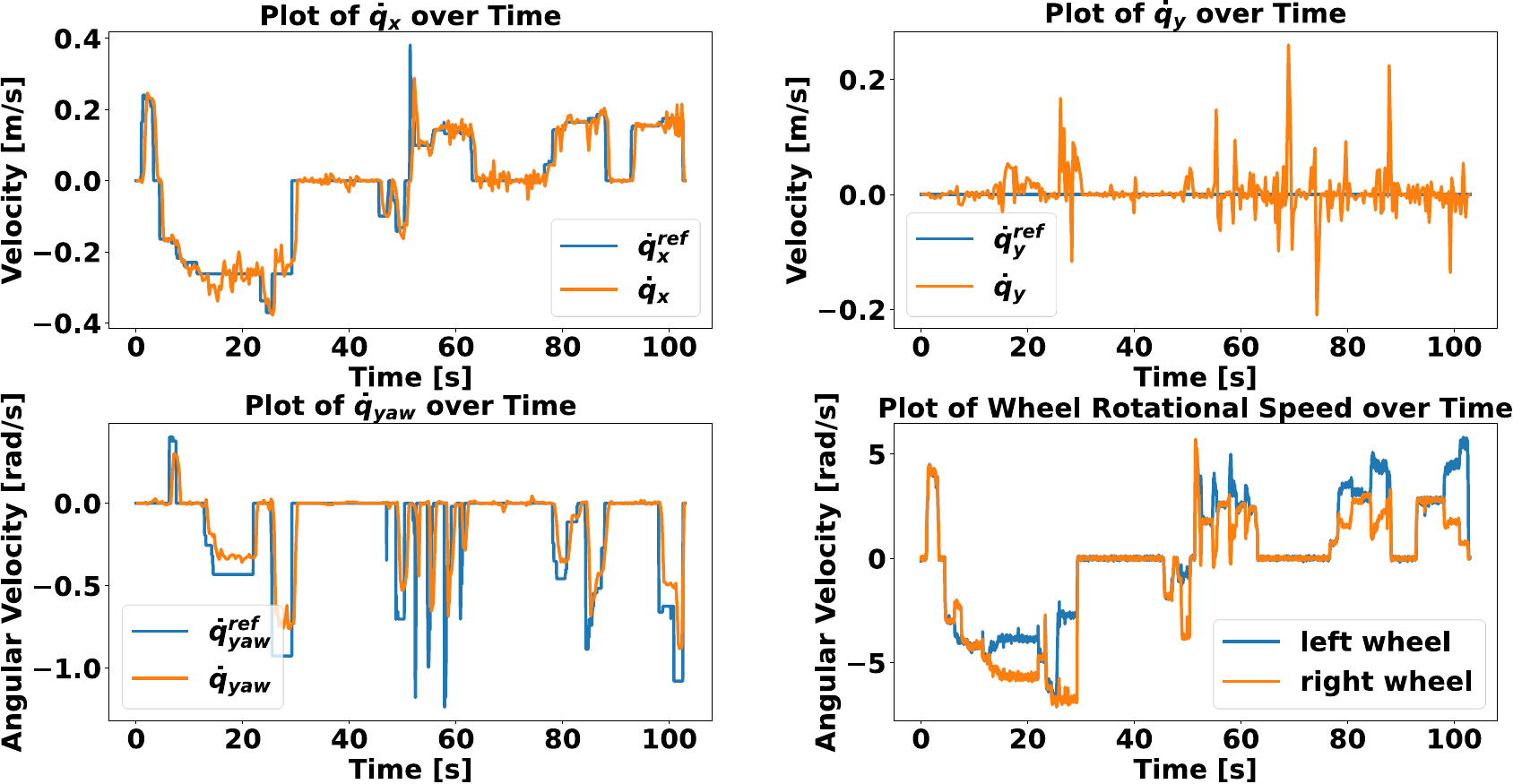}
      \vspace{-4.0ex}
      \caption{
         Time-series data of the velocity and actual wheel speeds during mapping. 
         The wheel speeds are calculated to track the commanded velocity.
      }
      \vspace{-4.0ex}
      \label{fig:expA-1_data}
    \end{center}
  \end{figure}

  この実験では、平面を車輪で移動しながら地図を作成する。
  実験環境と地図を作成する様子を\figref{fig:expA-1}に，
  その際の速度$q_x,\,q_y,\,q_\mathrm{yaw},\,q_x^\mathrm{ref},\,q_y^\mathrm{ref},\,q_\mathrm{yaw}^\mathrm{ref}$と
  輪速$\dot{\theta}_\mathrm{L\,wheel},\,\dot{\theta}_\mathrm{R\,wheel}$の時系列データを\figref{fig:expA-1_data}に示す。

  \figref{fig:expA-1}に示される通り、WiXusは車輪で移動することで、D455を用いたRTAB-Map\cite{rtabmap2019}により
  ロボット周囲の環境の地図を作成しつつ、自己位置推定を行うことができている。
  ここでは、ワイヤ駆動制御はFreeモードであり、脚車輪制御は車輪駆動制御モードである。
  また、\figref{fig:expA-1}\ctext{3}ではbase linkのPitch角度を変更して、
  SLAMに用いているカメラの視野をPitch方向にも変更できていることがわかる。

  \figref{fig:expA-1_data}を見ると、輪速$\dot{\theta}_\mathrm{L\,wheel},\,\dot{\theta}_\mathrm{R\,wheel}$により
  並進速度$q_x$が目標通りに実現されていることがわかる。
  回転速度$q_\mathrm{yaw}$には誤差があるが、
  0 m/sであるはずの$q_y$がそうでないことから考えて、
  SLAMによる速度推定の誤差である可能性が考えられる。
  本論文における自己位置推定ではD455のRGB-D画像とIMUセンサの情報のみを入力としたが、
  輪速やワイヤ速度等を与えることで、より安定した推定が行えると考えられる。

  環境にワイヤを接続して駆動するロボットにおいて、
  その環境を観測し記録することは重要な事前準備であるが、
  本実験はWiXus自身がそれを実行可能であることを示している。
  なお、\subsecref{subsec:exp_rescue}や\subsecref{subsec:exp_tool}の実験において、
  ワイヤを接続する環境を既知として実験を開始するが、
  それは予めWiXusによって観測し記録したデータを用いている。
}%

\subsubsection{Three-Dimensional Mobility: Cliff Climbing}
\switchlanguage%
{%
  \begin{figure}[t]
    \begin{center}
      \includegraphics[width=1.0\columnwidth]{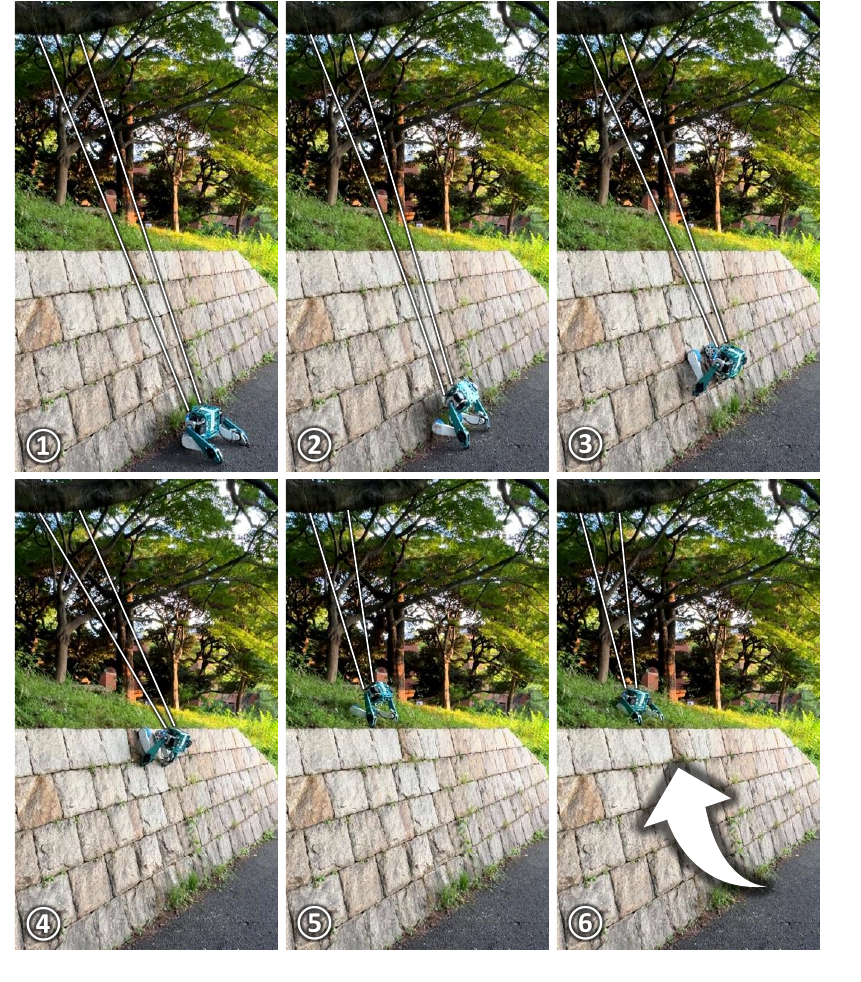}
      \vspace{-6.0ex}
      \caption{
        Scenes from the cliff climbing experiment. 
        WiXus successfully climbs the cliff by coordinating its two wheeled-legs with two wires anchored to the environment.
      }
      \vspace{-2.0ex}
      \label{fig:expA-2}
    \end{center}
  \end{figure}

  \begin{figure}[t]
    \begin{center}
      \includegraphics[width=1.0\columnwidth]{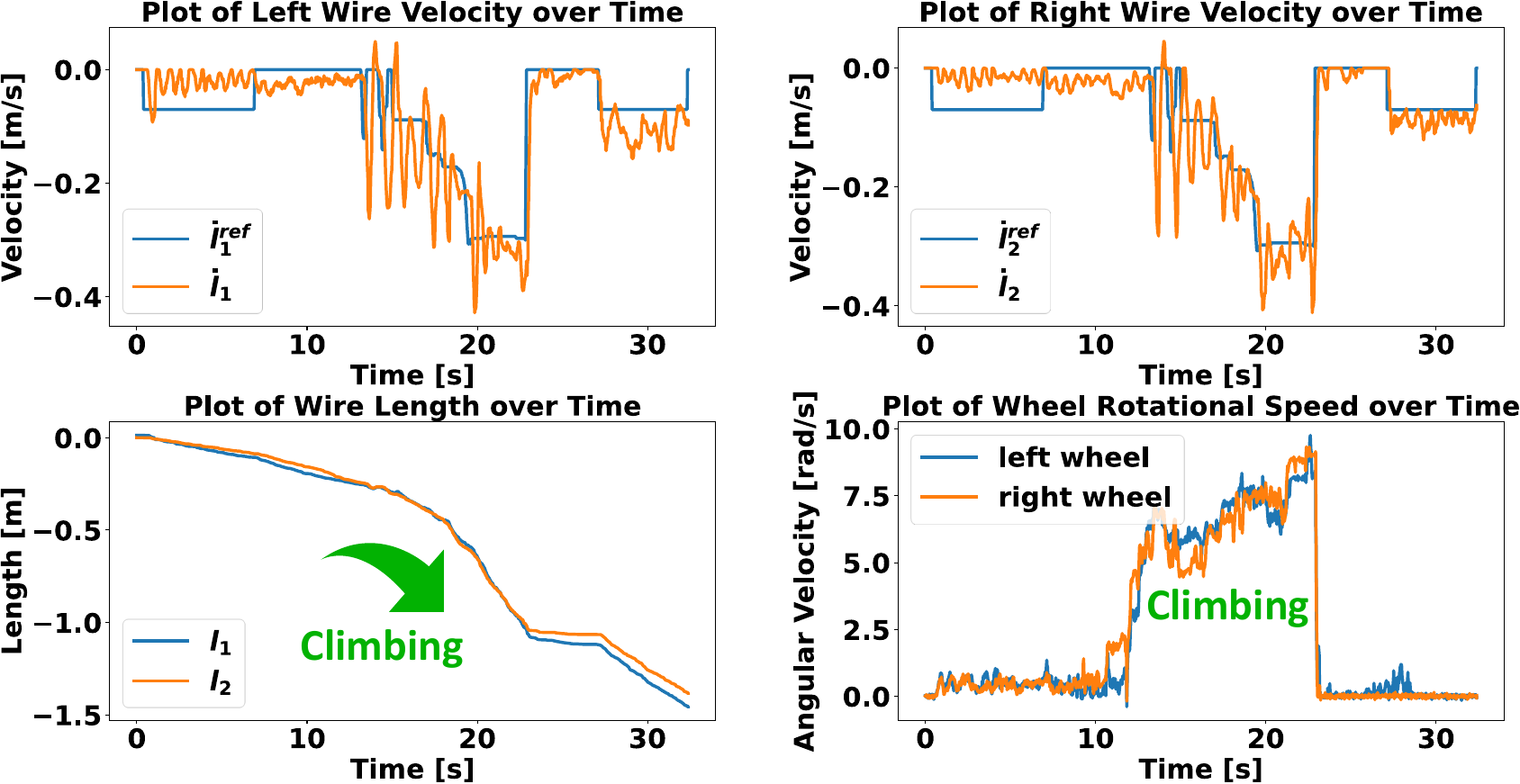}
      \vspace{-4.0ex}
      \caption{
         Time-series data of wire velocity, wire length, and wheel speeds. 
         The data shows that the wire winding and wheel drive systems are actuated simultaneously.
      }
      \vspace{-5.0ex}
      \label{fig:expA-2_data}
    \end{center}
  \end{figure}

  In this experiment, WiXus climbs a cliff by combining wire-driven actuation and wheeled-legged locomotion.
  % The cliff climbing process is shown in \figref{fig:expA-2},
  % and the time series data of the wire velocities $\dot{l}_1,\,\dot{l}_2,\,\dot{l}_1^\mathrm{ref},\,\dot{l}_2^\mathrm{ref}$,
  % the wire lengths $l_1,\,l_2$, and the wheel velocities $\dot{\theta}_\mathrm{L\,wheel},\,\dot{\theta}_\mathrm{R\,wheel}$
  % are shown in \figref{fig:expA-2_data}.
  As shown in \figref{fig:expA-2},
  WiXus anchors two wires to a tree branch located above the cliff 
  and winds them up to climb while driving its wheels along the cliff surface.
  Here, the wire-driven controller is in the \textit{wire velocity control} mode, 
  and the wheeled-legged controller is in the \textit{wheel driving control} mode.

  Looking at \figref{fig:expA-2_data},
  the decrease in the wire lengths $l_1,\,l_2$ together 
  with the wheel velocities $\dot{\theta}_\mathrm{L\,wheel},\,\dot{\theta}_\mathrm{R\,wheel}$
  indicates that WiXus climbs the cliff by coordinating wire-driven actuation and wheeled-legged locomotion.
  Focusing on the wire velocities $\dot{l}_1,\,\dot{l}_2$, they track the target values with oscillations.
  This is considered to be caused by the insufficient control frequency of the velocity P control.
  The M2006 motors used for the wire-driven actuation are designed to receive current commands as control inputs.
  Therefore, the P control is executed on WiXus's onboard computer, resulting in a lower control frequency.
  Since the wire lengths do not oscillate, this does not affect the behavior in this experiment,
  but for more precise tasks, high-frequency feedback control on the motor driver board with servo motors would be desirable.
}%
{%
  \begin{figure}[t]
    \begin{center}
      \includegraphics[width=1.0\columnwidth]{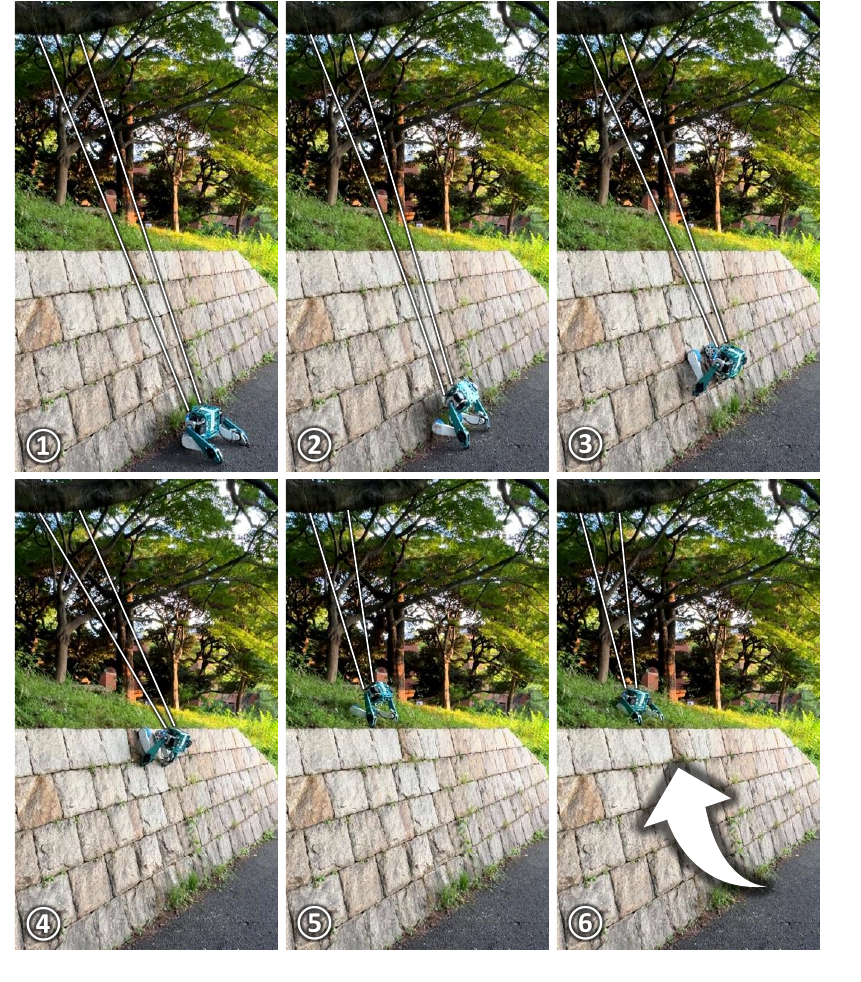}
      \vspace{-4.0ex}
      \caption{
        Scenes from the cliff climbing experiment. 
        WiXus successfully climbs the cliff by coordinating its two wheeled-legs with two wires anchored to the environment.
      }
      \vspace{-4.0ex}
      \label{fig:expA-2}
    \end{center}
  \end{figure}

  \begin{figure}[t]
    \begin{center}
      \includegraphics[width=1.0\columnwidth]{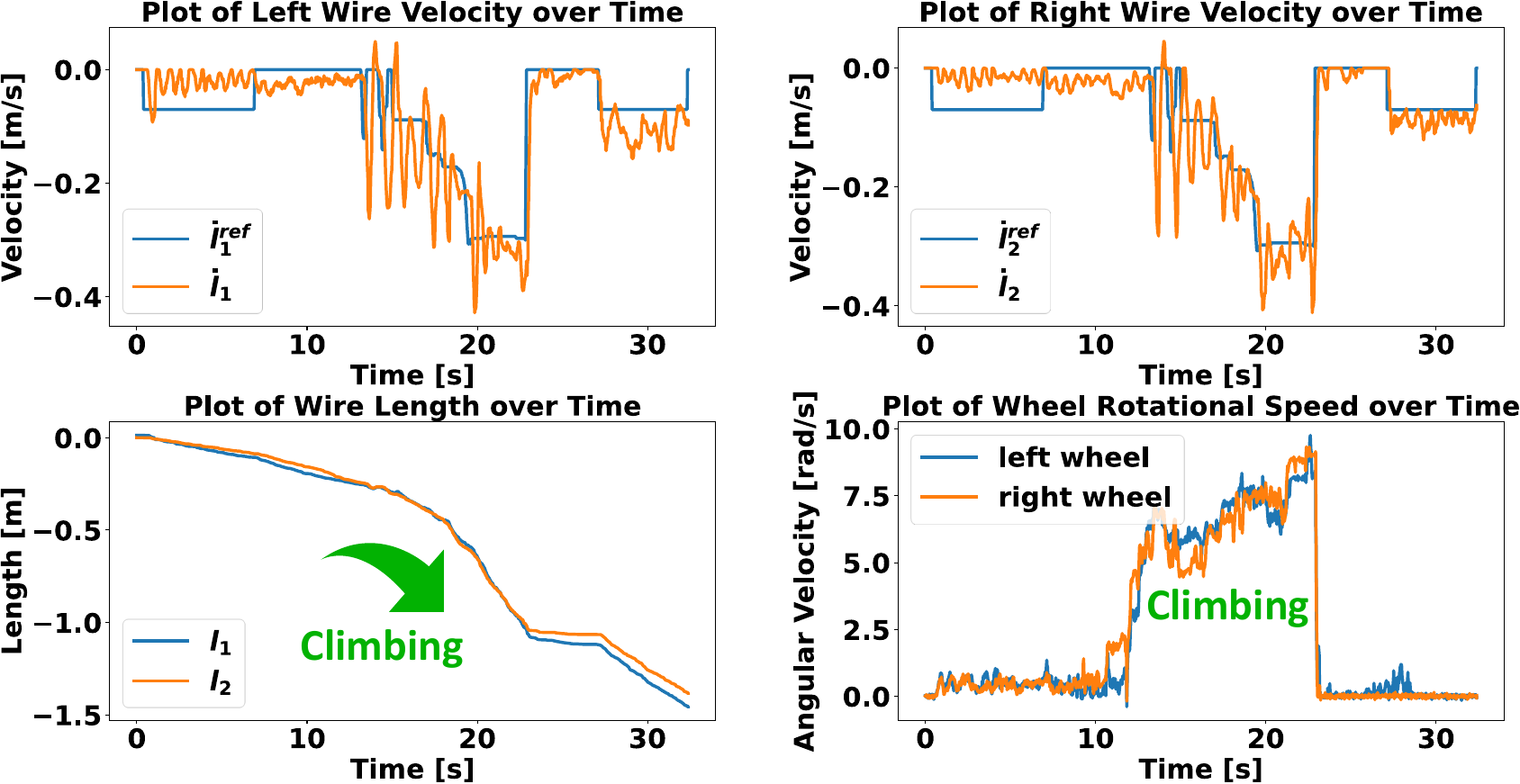}
      \vspace{-4.0ex}
      \caption{
         Time-series data of wire velocity, wire length, and wheel speeds. 
         The data shows that the wire winding and wheel drive systems are actuated simultaneously.
      }
      \vspace{-3.0ex}
      \label{fig:expA-2_data}
    \end{center}
  \end{figure}

  この実験では、ワイヤ駆動と脚車輪を組み合わせた動作による崖上りを行う。
  WiXusが崖上りを行う様子を\figref{fig:expA-2}に，
  その際のワイヤ速度$\dot{l}_1,\,\dot{l}_2,\,\dot{l}_1^\mathrm{ref},\,\dot{l}_2^\mathrm{ref}$と
  ワイヤ長さ$l_1,\,l_2$、輪速$\dot{\theta}_\mathrm{L\,wheel},\,\dot{\theta}_\mathrm{R\,wheel}$の
  時系列データを\figref{fig:expA-2_data}に示す。

  \figref{fig:expA-2}に示される通り、
  崖の上にある木の枝に2本のワイヤを接続し、
  それらを巻取ることで上昇しながら、崖を車輪で走行することによって崖上りを実現している。
  ここでは、ワイヤ駆動制御はワイヤ速度制御モード、脚車輪制御は車輪駆動制御モードである。

  \figref{fig:expA-2_data}を見ると、
  ワイヤ長さ$l_1,\,l_2$の減少と輪速$\dot{\theta}_\mathrm{L\,wheel},\,\dot{\theta}_\mathrm{R\,wheel}$から、
  ワイヤ駆動と脚車輪を連携させて崖上りを実現したことがわかる。
  ここで、ワイヤ速度$\dot{l}_1,\,\dot{l}_2$に注目すると、
  目標値に対して振動を伴って追従していることがわかる。
  これは、速度P制御の制御周波数が十分でないことに起因すると考えられる。
  ワイヤ駆動に用いたモータであるM2006は、制御入力として電流値を受け取る仕様である。
  そのため、P制御はWiXusのコンピュータ内で実行され、制御周波数が低くなる。
  ワイヤ長さが振動していないことから、本実験の動作に影響していないことがわかるが、
  より精密なタスクのためには、サーボモータを用いた基板内での高周波数なフィードバック制御が望まれる。
}%

\subsection{Object Manipulation: A Rescue Task} \label{subsec:exp_rescue}
\switchlanguage%
{%
  \begin{figure}[t]
    \begin{center}
      \includegraphics[width=1.0\columnwidth]{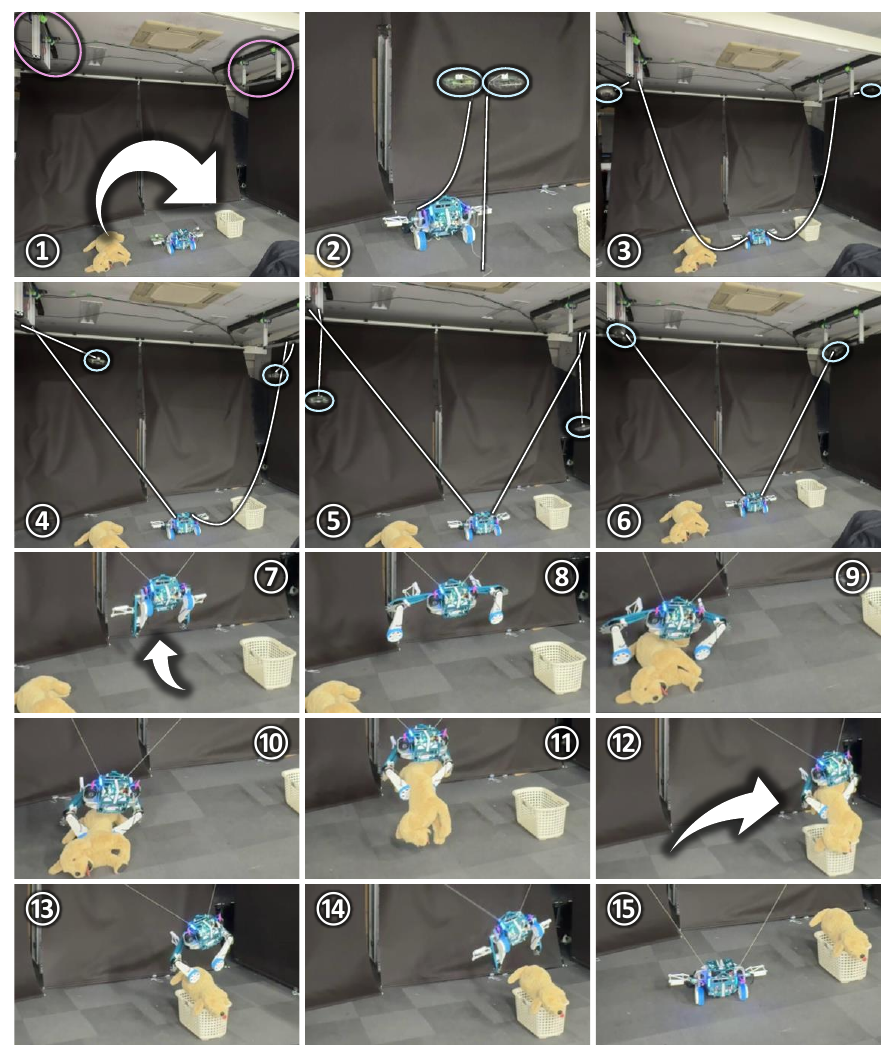}
      \vspace{-4.0ex}
      \caption{
        Scenes from the rescue task experiment. 
        Two Flying Anchors autonomously anchor two wires to the environment, 
        enabling WiXus to move the space using the wires. 
        Subsequently, the robot manipulates and transfers a dog using its wheeled-legs.
      }
      \vspace{-2.0ex}
      \label{fig:expB}
    \end{center}
  \end{figure}

  \begin{figure}[t]
    \begin{center}
      \includegraphics[width=1.0\columnwidth]{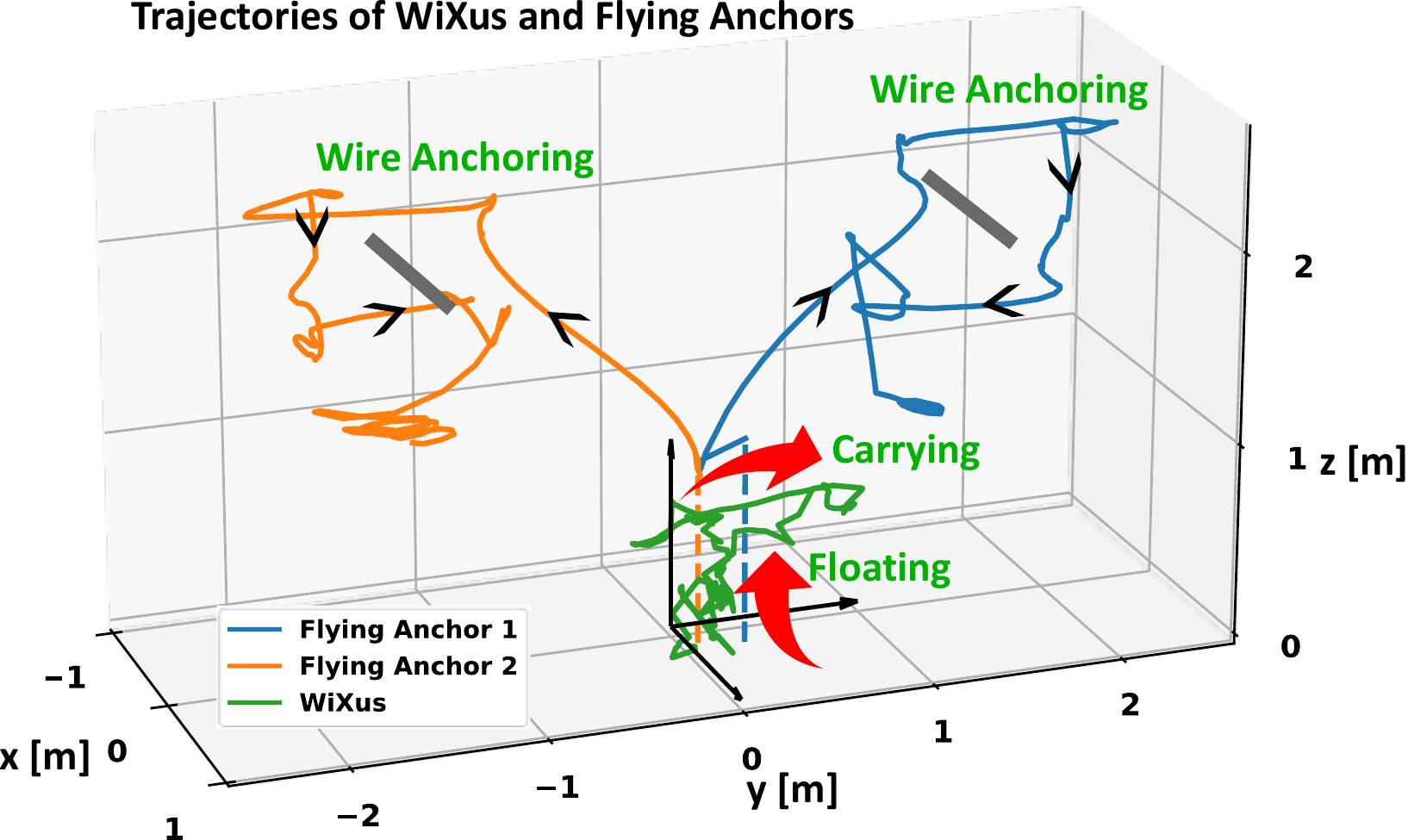}
      \vspace{-4.0ex}
      \caption{
        The trajectories of the two Flying Anchors and WiXus. 
        This figure shows the paths of the Flying Anchors as they anchor the wires to the environment, 
        and the resulting trajectory of WiXus as it moves through the space.
      }
      \vspace{-4.0ex}
      \label{fig:expB_data}
    \end{center}
  \end{figure}

  In this experiment, WiXus performs a rescue-like task to demonstrate its object manipulation capability,
  lifting and transporting a dog (a stuffed animal) to a safe location.
  In this task, two flying anchors are used to autonomously anchor two wires to the environment,
  also demonstrating that WiXus can autonomously anchor wires to the environment.
  % The execution of the rescue task is shown in \figref{fig:expB},
  % and the trajectories of WiXus and the flying anchors are shown in \figref{fig:expB_data}.

  As shown in \figref{fig:expB}\ctext{1}, the goal of this task is to carry the dog into the basket.
  A ceiling-mounted frame marked with red circles is available for wire anchoring.
  At first, the wire-driven controller is in the \textit{free} mode 
  and the wheeled-legged controller is in the \textit{wheel driving control} mode.
  In \figref{fig:expB}\ctext{2}--\ctext{6}, 
  the flying anchors fly around using the RGB-D camera-based control method \cite{inoue2025anchor}
  and autonomously anchor the wires to the environment.
  In \figref{fig:expB}\ctext{7}, WiXus suspends itself by wire-driven actuation,
  and in \figref{fig:expB}\ctext{8}, it changes its joint angles to take a posture suitable for object manipulation.
  Here, the wire-driven controller is switched to the \textit{CoG velocity control} mode,
  and the wheeled-legged controller is switched to the \textit{manipulation control} mode.
  In \figref{fig:expB}\ctext{9}--\ctext{15}, WiXus grasps the dog with its wheeled-legs
  and transports it to the basket by moving through space while suspended by the wires.

  As shown in \figref{fig:expB_data},
  the two flying anchors anchor the wires to the environment by flying around the frame,
  and WiXus moves through space while suspended.

  This experiment demonstrates the unique capability of combining wire-driven actuation and wheeled-legged locomotion,
  where the wheeled-legs become arms when WiXus is suspended by the wires, enabling object manipulation.
  Furthermore, the entire sequence from wire anchoring to object manipulation operates seamlessly,
  showing the high task execution capability of WiXus.
}%
{%
  \begin{figure}[t]
    \begin{center}
      \includegraphics[width=1.0\columnwidth]{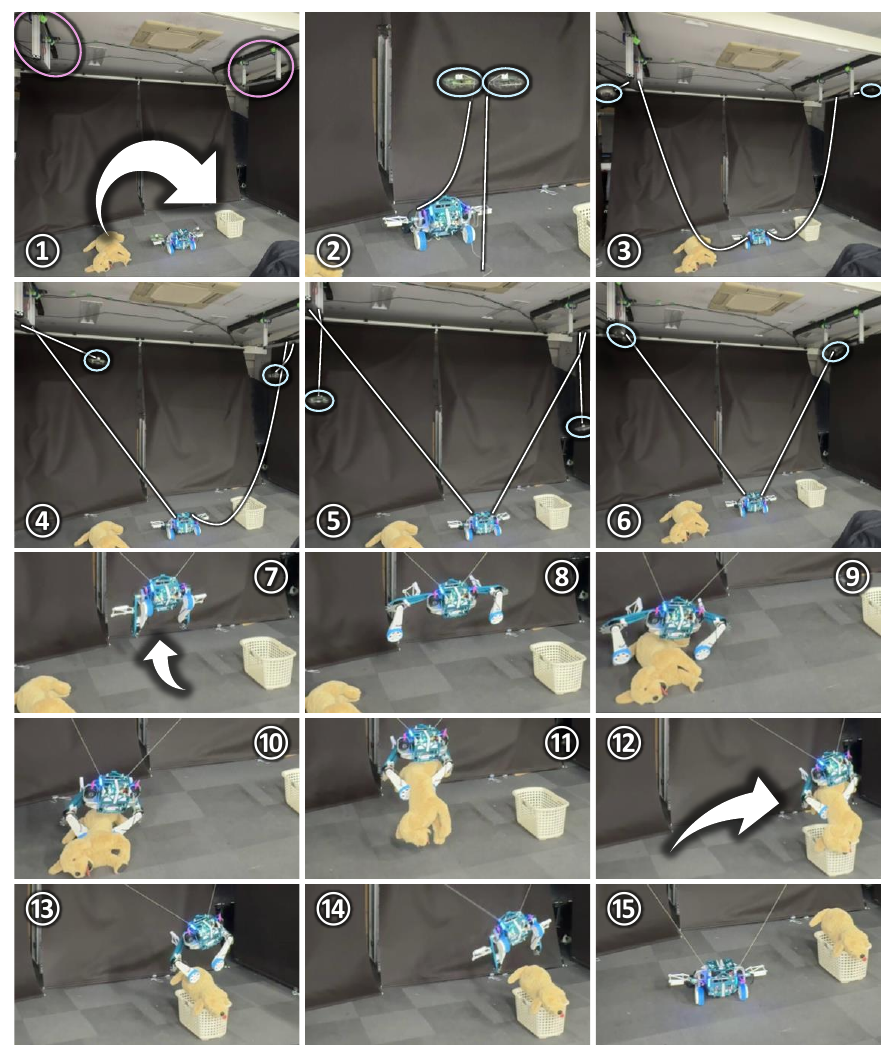}
      \vspace{-4.0ex}
      \caption{
        Scenes from the rescue task experiment. 
        Two Flying Anchors autonomously anchor two wires to the environment, 
        enabling WiXus to move the space using the wires. 
        Subsequently, the robot manipulates and transfers a dog using its wheeled-legs.
      }
      \vspace{-3.0ex}
      \label{fig:expB}
    \end{center}
  \end{figure}

  \begin{figure}[t]
    \begin{center}
      \includegraphics[width=1.0\columnwidth]{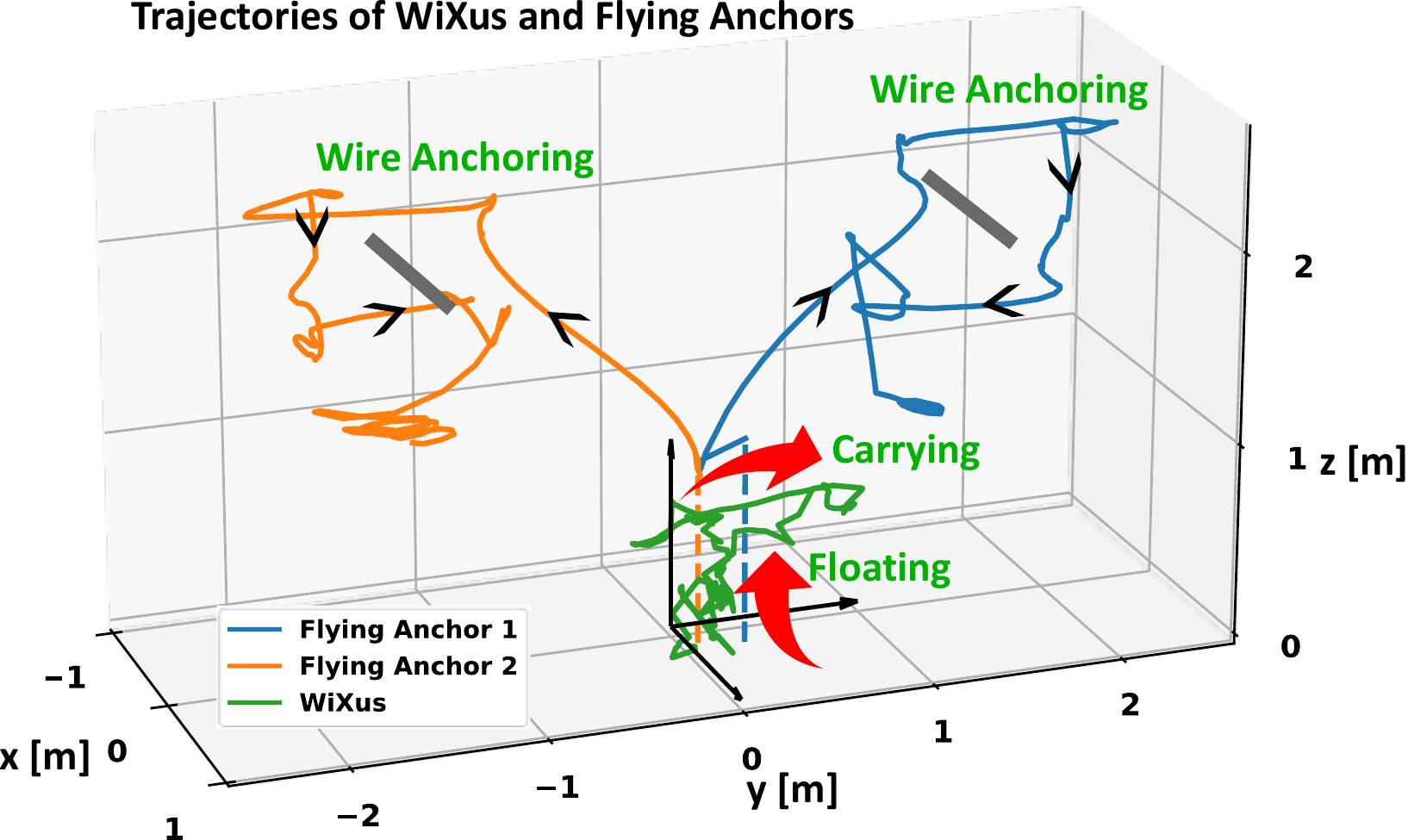}
      \vspace{-4.0ex}
      \caption{
        The trajectories of the two Flying Anchors and WiXus. 
        This figure shows the paths of the Flying Anchors as they anchor the wires to the environment, 
        and the resulting trajectory of WiXus as it moves through the space.
      }
      \vspace{-3.0ex}
      \label{fig:expB_data}
    \end{center}
  \end{figure}

  この実験では、WiXusの物体操作能力を示すため、
  犬 (ぬいぐるみ) を安全な場所へ持ち上げて移動させる、救助に見立てたタスクを行う。
  このタスクでは、２本ワイヤを2台の飛行アンカーによって環境へ接続することで、
  WiXusが自律的にワイヤを環境に接続できることも示す。
  救助タスク実行の様子を\figref{fig:expB}に、
  その際のWiXusと飛行アンカーの軌跡を\figref{fig:expB_data}に示す。

  \figref{fig:expB}\ctext{1}に示される通り、このタスクは犬をカゴの中へ運ぶことを目標とする。
  その際、赤丸で示した天井に固定されたフレームをワイヤ接続に利用できるものとする。
  はじめは、ワイヤ駆動制御はFreeモード、脚車輪制御は車輪駆動制御モードである。
  \figref{fig:expB}\ctext{2}--\ctext{6}で、飛行アンカーがRGB-Dカメラを用いた制御手法\cite{inoue2025anchor}により飛行し、
  自律的にワイヤを環境に結びつけた。
  \figref{fig:expB}\ctext{7}でワイヤ駆動により浮遊し、
  \figref{fig:expB}\ctext{8}で関節角度を変更することで物体操作ができる姿勢となった。
  ここで、ワイヤ駆動制御は重心速度制御モード、脚車輪制御は物体操作制御モードに切り替えられている。
  \figref{fig:expB}\ctext{9}--\ctext{15}で、犬を脚車輪で把持しながら、ワイヤ駆動で空間を移動することで、
  犬をカゴへ運ぶことを実現した。

  \figref{fig:expB_data}を見ると、
  ２台の飛行アンカーがフレームの周りを飛行することでワイヤを環境に接続し、
  WiXusが空間を移動していることがわかる。

  本実験は、ワイヤ駆動で浮遊した脚車輪が腕となって物体操作を可能とするという、
  ワイヤ駆動と脚車輪を組み合わせて発揮される特別な性質を示している。
  さらに、環境へのワイヤ接続から物体操作までが途切れなく一連として動作しており、
  WiXusのタスク実行能力の高さがわかる。
}%

\subsection{Tool Utilization: Harvesting with Loppers} \label{subsec:exp_tool}
\switchlanguage%
{%
  \begin{figure}[t]
    \begin{center}
      \includegraphics[width=1.0\columnwidth]{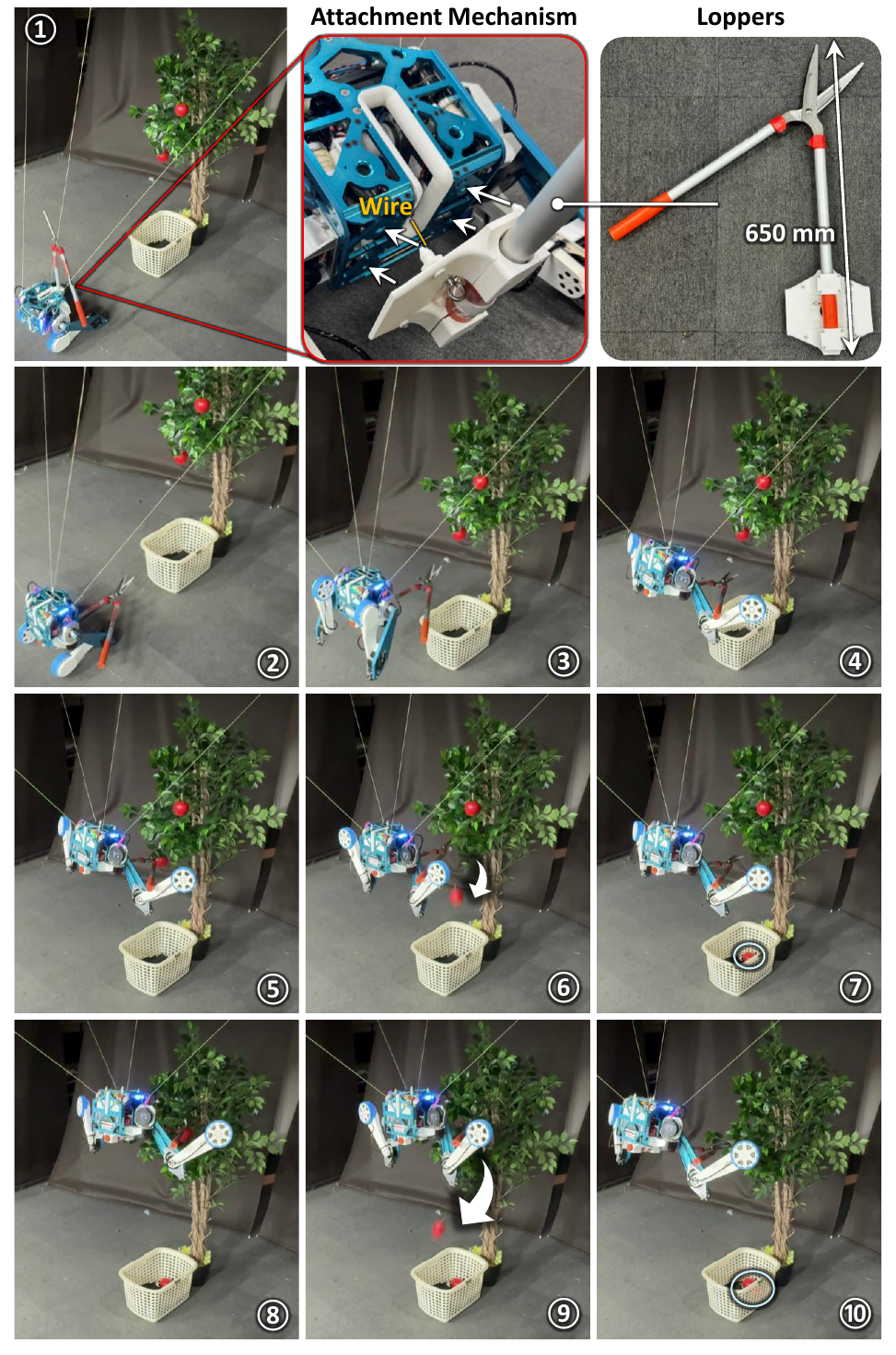}
      \vspace{-4.0ex}
      \caption{
        The tool attachment mechanism for the loppers and scenes from the apple harvesting task experiment. 
        WiXus moves to the target apple using its wire-drive, 
        and then uses its wheeled-legs to actuate the loppers and perform the harvest.
      }
      \vspace{-5.0ex}
      \label{fig:expC}
    \end{center}
  \end{figure}

  \begin{figure}[t]
    \begin{center}
      \includegraphics[width=1.0\columnwidth]{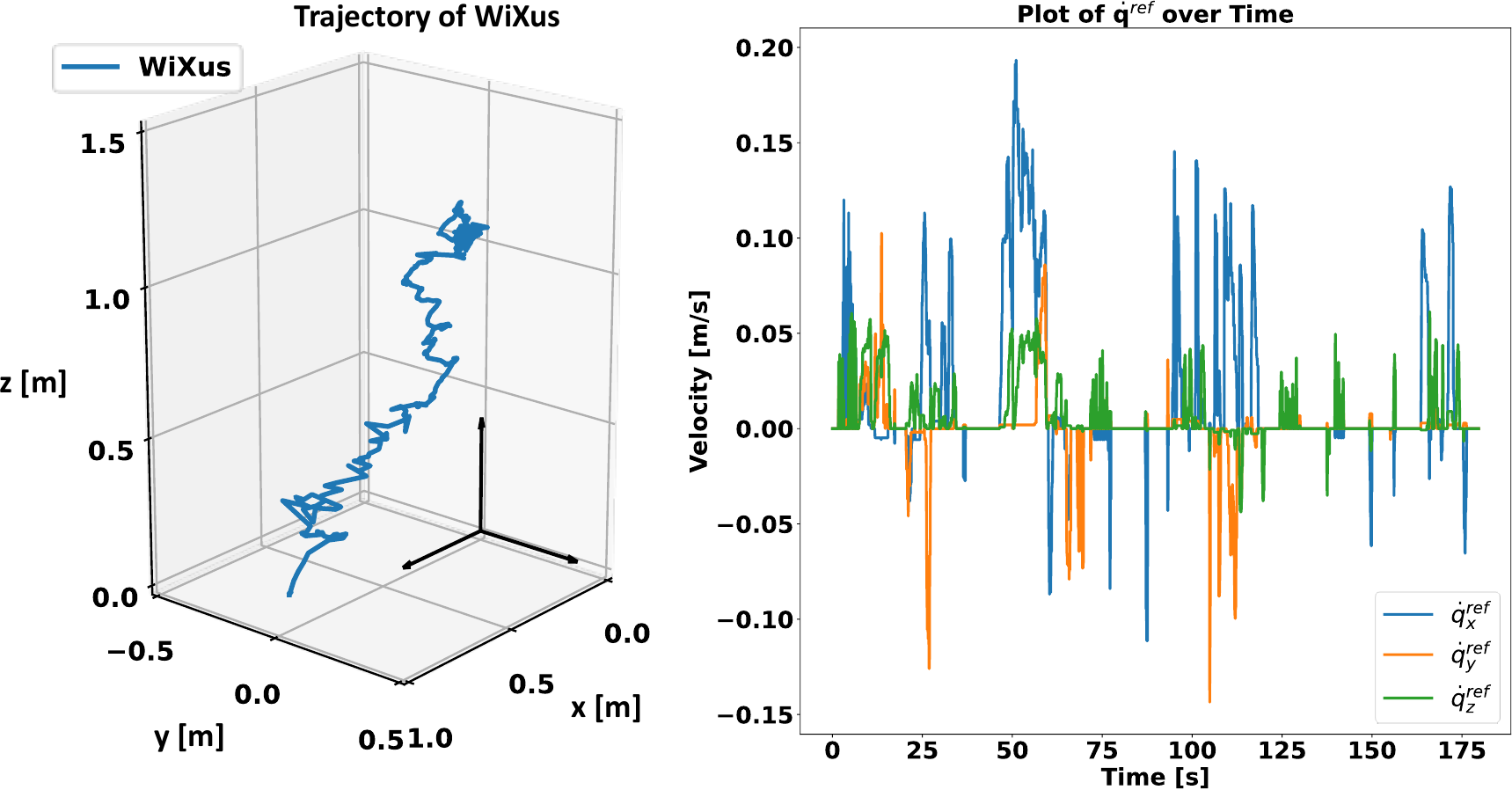}
      \vspace{-4.0ex}
      \caption{
        Trajectory of WiXus and the corresponding time-series data of the commanded velocity. 
        The results show WiXus following fine manual control commands to move through space.
      }
      \vspace{-4.0ex}
      \label{fig:expC_data}
    \end{center}
  \end{figure}

  In this experiment, WiXus performs a harvesting-like task using a tool to demonstrate its ability to execute tasks with tools.
  It uses loppers to cut apples from a tree.
  % The tool used and the harvesting process are shown in \figref{fig:expC},
  % and the trajectory of WiXus, the target CoG velocities $q_x,\,q_y,\,q_z,\,q_x^\mathrm{ref},\,q_y^\mathrm{ref},\,q_z^\mathrm{ref}$,
  % and the time series data of the wire lengths $l_1,\,l_2,\,l_3,\,l_4$ are shown in \figref{fig:expC_data}.

  As shown in \figref{fig:expC}\ctext{1}, the experiment is conducted with four wires anchored to the environment.
  WiXus attaches a 650 mm long pair of loppers to its body using the tool attachment wire.
  The faces of the Base-Link have uneven surfaces, 
  which fit into the matching shape of the fixture attached to the loppers to hold them in place.
  The loppers are secured to the body by keeping tension on the tool attachment wire.
  In \figref{fig:expC}\ctext{2}--\ctext{4}, WiXus suspends itself with the wire-driven actuation
  and changes the joint angles of its wheeled-legs to take a posture for operating the loppers.
  Here, the wire-driven controller is in the \textit{CoG velocity control} mode, 
  and the wheeled-legged controller is in the \textit{tool utilization control} mode.
  In \figref{fig:expC}\ctext{5}--\ctext{7}, WiXus moves through space using the wire-driven actuation to reach an apple,
  and closes and opens the loppers with its wheeled-legs to harvest an apple.
  In \figref{fig:expC}\ctext{8}--\ctext{10}, it similarly harvests a second apple.

  In this experiment, an operator controls the target CoG velocity of WiXus while observing the apples and the loppers.
  As seen in \figref{fig:expC_data},
  the target CoG velocities $q_x^\mathrm{ref},\,q_y^\mathrm{ref},\,q_z^\mathrm{ref}$ are not smooth,
  showing that WiXus is finely maneuvered.

  The loppers used in this experiment require not only controlling their orientation
  but also switching their state between closing and opening.
  WiXus attaches the loppers to its body using the tool attachment wire,
  controls its whole-body posture including the loppers using the anchored wires,
  and controls the state of the loppers (closing/opening) using its wheeled-legs.
  This experiment demonstrates a new approach in which a wheeled-legged robot performs tool utilization
  that requires tool operation while suspended by wire-driven actuation.
}%
{%
  \begin{figure}[t]
    \begin{center}
      \includegraphics[width=1.0\columnwidth]{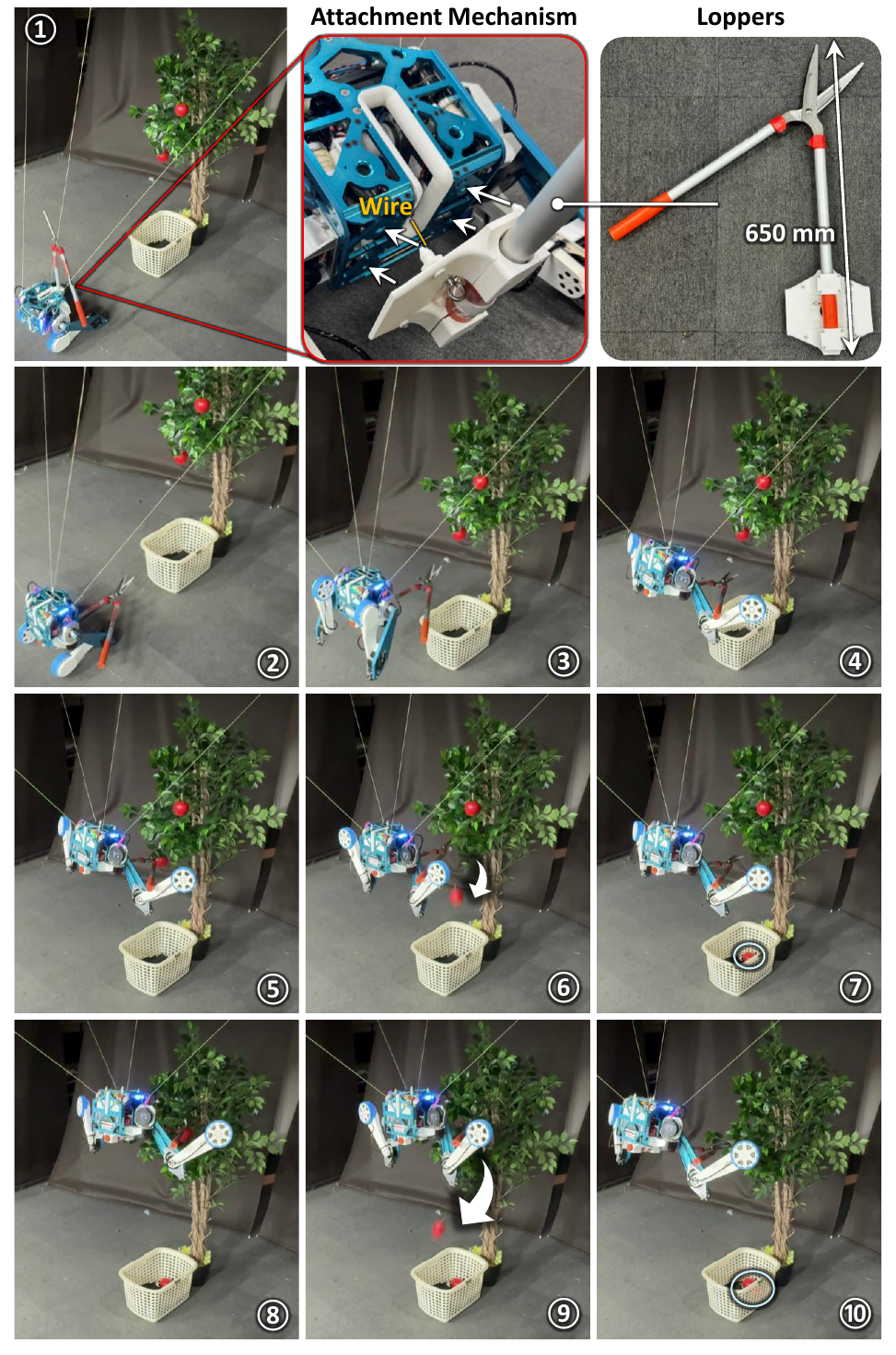}
      \vspace{-4.0ex}
      \caption{
        The tool integration mechanism for the loppers and scenes from the apple harvesting task experiment. 
        WiXus moves to the target apple using its wire-drive, 
        and then uses its wheeled-legs to actuate the loppers and perform the harvest.
      }
      \vspace{-3.0ex}
      \label{fig:expC}
    \end{center}
  \end{figure}

  \begin{figure}[t]
    \begin{center}
      \includegraphics[width=1.0\columnwidth]{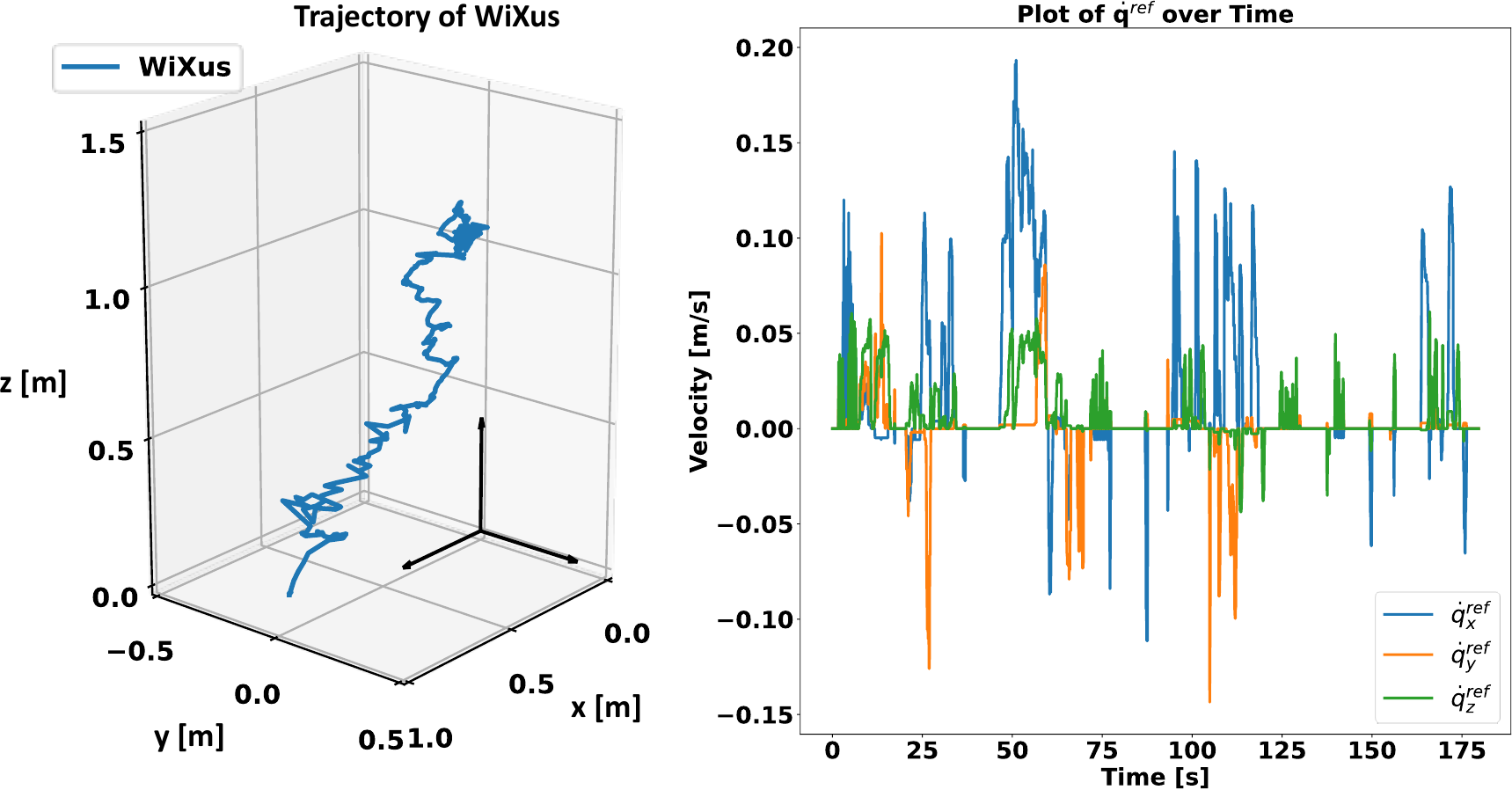}
      \vspace{-4.0ex}
      \caption{
        Trajectory of WiXus and the corresponding time-series data of the commanded velocity. 
        The results show WiXus following fine manual control commands to move through space.
      }
      \vspace{-3.0ex}
      \label{fig:expC_data}
    \end{center}
  \end{figure}

  この実験では、WiXusが道具を利用してタスクを実行できることを示すため、
  刈込鋏でリンゴ (模型) を木 (模型) から切り離す、リンゴの収穫に見立てたタスクを行う。
  利用する道具とリンゴ収穫の様子を\figref{fig:expC}に，
  その際のWiXusの軌跡と目標重心速度$q_x,\,q_y,\,q_z,\,q_x^\mathrm{ref},\,q_y^\mathrm{ref},\,q_z^\mathrm{ref}$、
  ワイヤ長さ$l_1,\,l_2,\,l_3,\,l_4$の時系列データ\figref{fig:expC_data}に示す．

  \figref{fig:expC}\ctext{1}に示される通り、４本のワイヤを環境に接続した状態で実験を行う。
  WiXusには全長650 mmの刈込鋏を道具合体用のワイヤで合体させる。
  WiXusのbase linkの各面は凹凸形状を有しており、刈込鋏を固定する治具の凹凸とはめ合うことで道具を固定する。
  合体用のワイヤに常に張力をかけることで道具を身体に合体させる。
  \figref{fig:expC}\ctext{2}--\ctext{4}で、ワイヤ駆動により浮遊し、
  脚車輪の関節角度を変更することで刈込鋏を操作できる姿勢となった。
  ここでは、ワイヤ駆動制御は重心速度制御モード、脚車輪制御は道具利用制御モードである。
  \figref{fig:expC}\ctext{5}--\ctext{7}で、リンゴが収穫できる位置までワイヤ駆動で空間を移動し、
  脚車輪を動かすことで刈込鋏を閉じて開くことで、リンゴを収穫した。
  \figref{fig:expC}\ctext{8}--\ctext{10}も同様に2つ目のリンゴを収穫した。

  本実験では操縦者がリンゴと刈込鋏を見ながらWiXusの目標重心速度を操作したが、
  \figref{fig:expC_data}から、
  なめらかでない目標重心速度$q_x^\mathrm{ref},\,q_y^\mathrm{ref},\,q_z^\mathrm{ref}$で
  WiXusが細かく操縦されたことがわかる。

  本実験で利用した道具である刈込鋏は、刈込鋏の姿勢を操作することに加えて、
  刈込鋏そのものを閉じる、開くという2状態に操作する必要があった。
  WiXusはワイヤにより刈込鋏を身体に合体させ、
  環境に接続したワイヤを用いたワイヤ駆動で刈込鋏を含んだ自身の姿勢を制御し、
  脚車輪によって刈込鋏の状態を制御した。
  ワイヤ駆動によって脚車輪ロボットが操作を要する道具を利用するという新たな展開が示された。
}%

% \section{Discussion} \label{sec:discussion}
% \switchlanguage%
% {%
%   hoge
% }%
% {%
%   hoge
% }%

\section{Conclusion} \label{sec:conclusion}
\switchlanguage%
{%
  In this study, we developed WiXus, a novel robot that integrates a wheeled-legged system and a wire-driven system,
  and conducted experiments to demonstrate its capabilities:
  a mapping experiment and a cliff climbing experiment for mobility capabilities,
  a rescue task experiment for object manipulation capability,
  and a harvesting experiment with loppers for tool utilization capability.
  WiXus is a two-wheeled-legged robot equipped with four wire winding modules for anchoring to the environment
  and one wire winding module for tool attachment.
  By using wire-driven actuation with wires anchored to the environment,
  WiXus frees its wheeled-legs from the roles of locomotion,
  and the experimental results confirm that this enhances the task execution capability of the wheeled-legs.
  Therefore, this study demonstrates that the approach of environmental anchoring with wire-driven actuation
  can expand the operational domain of wheeled-legged robots.

  While the demonstrations involve partial operator input, 
  they represent an initial step to validate the feasibility of the proposed concept, 
  which can be extended to autonomous task execution by integrating environmental perception and motion planning. 
  % In this paper, the operator provided target CoG velocities and wire velocities for the tasks.
  As a future direction, WiXus could plan its own motions to accomplish tasks based on the given environment and tools.
  If WiXus can autonomously perform motion planning with its body that is capable of both locomotion and manipulation,
  it is expected to accomplish more complex and diverse tasks.
}%
{%
  本研究では、
  脚車輪ロボットとワイヤ駆動ロボットの二つを融合させた新しいロボットであるWiXusを開発し、
  移動能力を示す実験として地図作成実験と崖上り実験、
  物体操作能力を示す実験として救助タスク実験、
  道具が利用可能であることを示す実験としてリンゴ収穫実験を行った。
  WiXusは環境接続用のワイヤ巻取りモジュールを4つ、道具合体用のワイヤ巻取りモジュールを1つ
  体内に搭載した2脚車輪ロボットである。
  環境に接続したワイヤを用いたワイヤ駆動で脚車輪を移動と支持の役割から解放し、
  脚車輪ロボットのタスク実行能力が引き上げられたことが各実験より確認できた。
  よって、ワイヤ駆動による環境利用というアプローチが、脚車輪ロボットの活動領域を拡張することを示した。

  本論文では実行するタスクに対して操縦者が目標の重心速度やワイヤ速度などを入力したが、
  今後の展望として、WiXus自身が与えられた環境、道具に基づいてタスクを完遂するための運動計画を行うことが考えられる。
  移動と物体操作が共に実現可能な身体で、自律的に運動計画を行うことができれば、
  より複雑な多種のタスクを実行できるようになると考えられる。
}%

{
  %\footnotesize
  %\small
  %\bibliographystyle{junsrt}
  \bibliographystyle{IEEEtran}
  \bibliography{bib}

@string{IROS2025 = "Proceedings of the 2025 IEEE/RSJ International Conference on Intelligent Robots and Systems"}

@string{IROS2024 = "Proceedings of the 2024 IEEE/RSJ International Conference on Intelligent Robots and Systems"}

@string{IROS2021 = "Proceedings of the 2021 IEEE/RSJ International Conference on Intelligent Robots and Systems"}

@string{IROS2019 = "Proceedings of the 2019 IEEE/RSJ International Conference on Intelligent Robots and Systems"}

@string{IROS2018 = "Proceedings of the 2018 IEEE/RSJ International Conference on Intelligent Robots and Systems"}

@string{ICRA2019 = "Proceedings of the 2019 IEEE International Conference on Robotics and Automation"}

@string{ROBIO2019 = "Proceedings of the 2019 IEEE International Conference on Robotics and Biomimetics"}

@INPROCEEDINGS{zhang2019system,
  author={Zhang, Chao and Liu, Tangyou and Song, Shuang and Meng, Max Q.-H.},
  booktitle=ROBIO2019, 
  title={{System Design and Balance Control of a Bipedal Leg-wheeled Robot}}, 
  year={2019},
  volume={},
  number={},
  pages={1869-1874},
  doi={10.1109/ROBIO49542.2019.8961814},
}

@INPROCEEDINGS{liu2019dynamic,
  author={Liu, Tangyou and Zhang, Chao and Song, Shuang and Meng, Max Q.-H.},
  booktitle=ROBIO2019, 
  title={{Dynamic Height Balance Control for Bipedal Wheeled Robot Based on ROS-Gazebo}}, 
  year={2019},
  volume={},
  number={},
  pages={1875-1880},
  doi={10.1109/ROBIO49542.2019.8961739},
}

@INPROCEEDINGS{bjelonic2021wholebody,
  author={Bjelonic, Marko and Grandia, Ruben and Harley, Oliver and Galliard, Cla and Zimmermann, Samuel and Hutter, Marco},
  booktitle=IROS2021, 
  title={{Whole-Body MPC and Online Gait Sequence Generation for Wheeled-Legged Robots}}, 
  year={2021},
  volume={},
  number={},
  pages={8388-8395},
  doi={10.1109/IROS51168.2021.9636371},
}

@article{joonho2024learning,
	author = {Joonho Lee  and Marko Bjelonic  and Alexander Reske  and Lorenz Wellhausen  and Takahiro Miki  and Marco Hutter },
	title = {{Learning robust autonomous navigation and locomotion for wheeled-legged robots}},
	journal = {Science Robotics},
	volume = {9},
	number = {89},
	pages = {eadi9641},
	year = {2024},
	doi = {10.1126/scirobotics.adi9641},
}

@INPROCEEDINGS{yang2023design,
  author={Yang, Zeyi and Bian, Zekun and Zhang, Wei},
  booktitle={2023 International Conference on Communications, Computing and Artificial Intelligence (CCCAI)}, 
  title={Design and Control of Multi-mode Wheeled-Bipedal Robot with Parallel Mechanism}, 
  year={2023},
  volume={},
  number={},
  pages={69-74},
  doi={10.1109/CCCAI59026.2023.00021},
}

@INPROCEEDINGS{klemm2019ascento,
  author={Klemm, Victor and Morra, Alessandro and Salzmann, Ciro and Tschopp, Florian and Bodie, Karen and Gulich, Lionel and Küng, Nicola and Mannhart, Dominik and Pfister, Corentin and Vierneisel, Marcus and Weber, Florian and Deuber, Robin and Siegwart, Roland},
  booktitle=ICRA2019,
  title={{Ascento: A Two-Wheeled Jumping Robot}}, 
  year={2019},
  volume={},
  number={},
  pages={7515-7521},
  doi={10.1109/ICRA.2019.8793792},
}

@misc{handle2019,
  title        = {{Handle (Boston Dynamics)}},
  howpublished = {\url{https://robotsguide.com/robots/handle}},
}

@misc{tita2024,
  title        = {{TITA (Direct Drive Technology)}},
  howpublished = {\url{https://shop.directdrive.com/products/tita}},
}

@misc{diabolo2024,
  title        = {{DIABOLO (Direct Drive Technology)}},
  howpublished = {\url{https://shop.directdrive.com/products/diablo-world-s-first-direct-drive-self-balancing-wheeled-leg-robot}},
}

@misc{go2w2025,
  title        = {{Go2-W (Unitree Robotics)}},
  howpublished = {\url{https://www.unitree.com/go2-w}},
}

@misc{b2w2025,
  title        = {{B2-W (Unitree Robotics)}},
  howpublished = {\url{https://www.unitree.com/b2-w}},
}

@INPROCEEDINGS{li2018wlr,
  author={Li, Xu and Zhou, Haitao and Feng, Haibo and Zhang, Songyuan and Fu, Yili},
  booktitle=IROS2018, 
  title={{Design and Experiments of a Novel Hydraulic Wheel-Legged Robot (WLR)}}, 
  year={2018},
  volume={},
  number={},
  pages={3292-3297},
  doi={10.1109/IROS.2018.8594484},
}

@INPROCEEDINGS{li2019wlr2,
  author={Li, Xu and Zhou, Haitao and Zhang, Songyuan and Feng, Haibo and Fu, Yili},
  booktitle=IROS2019, 
  title={{WLR-II, a Hose-less Hydraulic Wheel-legged Robot}},
  year={2019},
  volume={},
  number={},
  pages={4339-4346},
  doi={10.1109/IROS40897.2019.8967935},
}

@article{cone1985skycam,
  title={Skycam-an aerial robotic camera system},
  author={Cone, Lawrence L},
  journal={Byte},
  volume={10},
  number={10},
  pages={122},
  year={1985},
  publisher={BYTE PUBL INC 70 MAIN ST, PETERBOROUGH, NH 03458}
}

@INPROCEEDINGS{8967836,
  author={Bury, Diane and Izard, Jean-Baptiste and Gouttefarde, Marc and Lamiraux, Florent},
  booktitle=IROS2019,
  title={{Continuous Collision Detection for a Robotic Arm Mounted on a Cable-Driven Parallel Robot}}, 
  year={2019},
  volume={},
  number={},
  pages={8097-8102},
  doi={10.1109/IROS40897.2019.8967836}
}

@INPROCEEDINGS{8794265,
  author={Miki, Takahiro and Khrapchenkov, Petr and Hori, Koichi},
  booktitle=ICRA2019,
  title="{UAV/UGV Autonomous Cooperation: UAV assists UGV to climb a cliff by attaching a tether}", 
  year={2019},
  volume={},
  number={},
  pages={8041-8047},
  doi={10.1109/ICRA.2019.8794265}
}

@inproceedings{inoue2024cubix,
  title={{CubiX: Portable Wire-Driven Parallel Robot Connecting to and Utilizing the Environment}},
  author={Shintaro Inoue and Kento Kawaharazuka and Temma Suzuki and Sota Yuzaki and Kei Okada and Masayuki Inaba},
  booktitle=IROS2024,
  year={2024},
  pages={1296-1301},
  doi={10.1109/IROS58592.2024.10802299}
}

@article{inoue2024overcoming,
  author = {Inoue, Shintaro and Kawaharazuka, Kento and Suzuki, Temma and Yuzaki, Sota and Okada, Kei and Inaba, Masayuki},
  title = {{Overcoming Physical Limitations Utilizing the Surrounding Environment with a Wire-Driven Multipurpose Robot}},
  journal = {Advanced Robotics Research},
  volume = {1},
  number = {1},
  pages = {202400021},
  doi = {https://doi.org/10.1002/adrr.202400021},
}

@inproceedings{inoue2025anchor,
  author={Inoue, Shintaro and Kawaharazuka, Kento and Yoneda, Keita and Yuzaki, Sota and Sahara, Yuta and Suzuki, Temma and Okada, Kei},
  booktitle=IROS2025,
  title={{An RGB-D Camera-Based Multi-Small Flying Anchors Control for Wire-Driven Robots Connecting to the Environment}}, 
  year={2025},
  volume={},
  number={},
  pages={20442-20447},
  doi={10.1109/IROS60139.2025.11247141},
}

@ARTICLE{bohren2010smach,
  author={Bohren, Jonathan and Cousins, Steve},
  journal={IEEE Robotics \& Automation Magazine}, 
  title={{The SMACH High-Level Executive [ROS News]}}, 
  year={2010},
  volume={17},
  number={4},
  pages={18-20},
  doi={10.1109/MRA.2010.938836}
}

@article{rtabmap2019,
  author = {Labb{\'e}, Mathieu and Michaud, Fran{\c{c}}ois},
  title = {{RTAB-Map as an open-source lidar and visual simultaneous localization and mapping library for large-scale and long-term online operation}},
  journal = {Journal of Field Robotics},
  volume = {36},
  number = {2},
  pages = {416-446},
  doi = {https://doi.org/10.1002/rob.21831},
  year = {2019}
}
}

\end{document}